\documentclass[10pt]{article}
\usepackage[preprint]{rlc}

\usepackage{amsmath}
\usepackage{amsthm}
\usepackage{amssymb}
\usepackage{booktabs}
\usepackage{multirow}
\usepackage{makecell}
\usepackage{algorithm}
\usepackage{algpseudocode}
\usepackage{graphicx}
\usepackage{subcaption}
\usepackage{nicefrac}
\usepackage{microtype}
\usepackage{enumitem}

\newcommand{\rev}[1]{{#1}}

\newtheorem{definition}{Definition}
\newtheorem{theorem}{Theorem}[section]
\newtheorem{proposition}[theorem]{Proposition}

\title{MemQ: Integrating Q-Learning into Self-Evolving Memory Agents over Provenance DAGs}

\author{Junwei Liao$^{1,2\rev{,*}}$ \quad Haoting Shi$^{1\rev{,*}}$ \quad Ruiwen Zhou$^{3}$ \quad Jiaqian Wang$^{4}$ \quad Shengtao Zhang$^{1}$ \\
\textbf{Wei Zhang$^{1}$ \quad Ying Wen$^{1,2}$ \quad Zhiyu Li$^{6}$ \quad Feiyu Xiong$^{6}$ \quad Bo Tang$^{5,6}$} \\
\textbf{Weinan Zhang$^{1,2}$ \quad Muning Wen$^{1}$} \\
$^{1}$Shanghai Jiao Tong University \quad $^{2}$Shanghai Innovation Institute \\
$^{3}$National University of Singapore \quad $^{4}$Xidian University \\
$^{5}$University of Science and Technology of China \quad $^{6}$MemTensor (Shanghai) Technology Co., Ltd. \\
\texttt{\{jwliao.ai, wnzhang, muningwen\}@sjtu.edu.cn \quad tangb@memtensor.cn}}

\begin{document}

\maketitle
\begingroup
\renewcommand{\thefootnote}{\rev{*}}
\footnotetext[1]{\rev{Equal contribution.}}
\endgroup

\begin{abstract}
Episodic memory allows LLM agents to accumulate and retrieve experience, but current methods treat each memory independently, i.e., evaluating retrieval quality in isolation without accounting for the dependency chains through which memories enable the creation of future memories. We introduce \textbf{MemQ}, which applies TD($\lambda$) eligibility traces to memory Q-values, propagating credit backward through a \emph{provenance DAG} that records which memories were retrieved when each new memory was created. Credit weight decays as $(\gamma\lambda)^d$ with DAG depth $d$, replacing temporal distance with structural proximity. \rev{We formalize this setting as an \emph{Exogenous-Context MDP}, separating externally supplied tasks from the evolving memory store, and represent retrieval-set values by a first-order mean of per-memory values plus an explicit approximation residual.} \rev{Across five interaction, code-generation, and agentic benchmarks, MemQ improves average held-out, final runtime, and cumulative success rates over the baseline with the highest average for each metric by 2.53, 4.23, and 2.13 percentage points, respectively. {Ablations on LiveCodeBench, BFCL, and DSBench support the benefit of bootstrapping and multi-depth credit propagation. Analyses of Q-values and runtime learning show that learned memory values are associated with subsequent success and that MemQ retains more of its historical task coverage.} {We also explore the limits of MemQ's benefits using multiple-choice reasoning tasks, where limited interaction offers fewer opportunities to reuse procedural experience.}} Code is available at \url{https://github.com/jwliao-ai/MemQ}.
\end{abstract}

\section{Introduction}
\label{sec:intro}

\rev{Large language models (LLMs) deployed as agents must adapt to novel tasks and changing environments, but updating their weights can be costly. External \emph{episodic memory} offers a complementary route: agents accumulate successes, failures, and discovered strategies \citep{park2023generative, packer2023memgpt, zhong2023memorybank, sumers2023cognitive}, then retrieve relevant experience to guide future behavior \citep{shinn2023reflexion, zhao2024expel, wang2023voyager}. Early systems score memories with fixed heuristics, typically embedding similarity, without learning memory values from task outcomes.}

\rev{Recent RL-based methods instead learn which memories are worth retrieving. Parametric approaches train retrieval policies \citep{yan2025memoryr1, zhang2025memact} and optimize broader memory operations \citep{ma2026finemem, shen2026membuilder, zhang2026retroagent}, whereas non-parametric approaches attach value estimates directly to memory entries \citep{pritzel2017neural}. Most closely related, MemRL \citep{zhang2026memrl} associates each vector-store entry with a Q-value and updates it by single-step exponential moving average (EMA), equivalent to a contextual-bandit update with $\gamma=0$. This makes retrieval quality learnable without changing the LLM's weights, but the feedback remains local to the current retrieval event.}

\rev{The unresolved gap is \emph{multi-step credit assignment across memory-creation events}. Retrieving memories for one task produces a new memory, which can later be retrieved to produce another, yielding exposure chains such as $m_a \to m_b \to m_c \to r$. A single-step update credits the memories retrieved for the current task, but gives earlier ancestors no signal from that outcome unless they are retrieved as well. Classical eligibility traces address delayed credit by propagating TD errors over multiple steps \citep{sutton1988learning, singh1996reinforcement, sutton2018reinforcement}. Off-policy multi-step variants refine how such credit is truncated or corrected \citep{watkins1989learning, peng1996incremental, munos2016safe}, while true-online traces, generalized advantage estimation, and distributed V-trace instantiate related bias--variance tradeoffs \citep{vanseijen2014true, schulman2016gae, espeholt2018impala}. Here, the relevant distance is structural depth in a memory-construction graph whose edges record retrieval exposure.}

\rev{We introduce \textbf{MemQ} to address this structural credit bottleneck. MemQ records retrieved parents in a \emph{provenance DAG} and propagates TD errors along every backward provenance path, weighting each path of length $d$ by $(\gamma\lambda)^d$. This replaces temporal step count in TD($\lambda$) with structural distance. An \emph{Exogenous-Context MDP} formalizes the externally supplied task stream, while a first-order per-memory value projection makes combinatorial retrieval tractable and retains an explicit approximation residual. Experiments on eight benchmarks evaluate held-out performance and runtime learning, separating interaction, code-generation, and agentic tasks from multiple-choice reasoning. We then isolate the provenance update through ablation studies, examine the learned values and learning trajectories, and measure the associated computational costs.}

\begin{figure}[t]
\centering
\includegraphics[width=\linewidth]{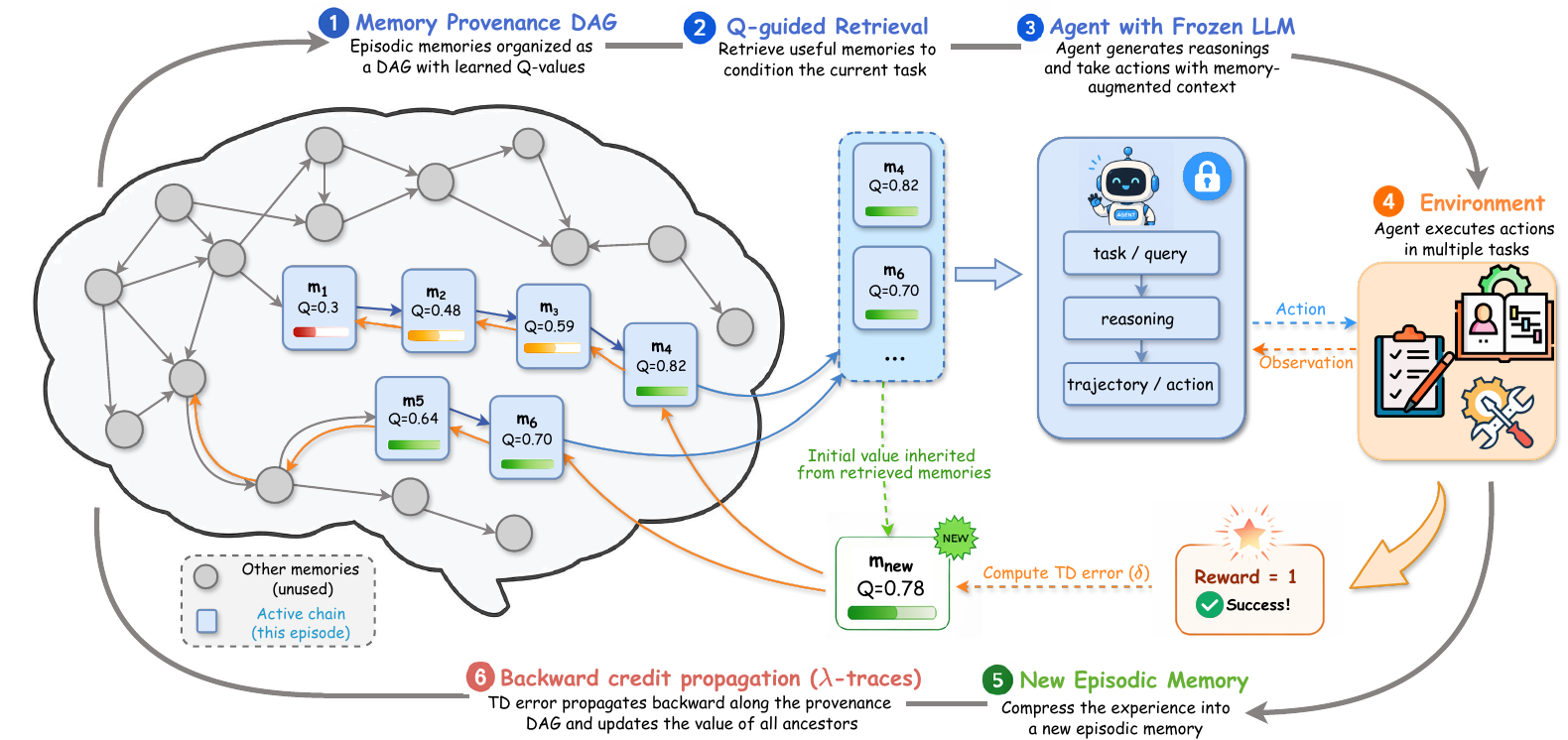}
\caption{\rev{Conceptual overview of MemQ. Retrieved memories support a new memory, and later outcomes send structurally decayed feedback to exposed ancestors in the resulting provenance DAG.}}
\label{fig:high_level}
\end{figure}

Our contributions are as follows:
\begin{enumerate}[leftmargin=*]
  \item {We identify the \emph{multi-step credit assignment problem} in episodic memory and formalize the setting as an \emph{Exogenous-Context MDP}, whose factored transition decouples exogenous task dynamics from endogenous memory evolution, motivating value decomposition over individual memories.}
  \item {We develop \textbf{MemQ}, a provenance-based credit propagation mechanism that applies TD($\lambda$) eligibility traces through the memory construction DAG, replacing temporal distance with structural depth, the first provenance-based credit assignment method for episodic memory valuation.}
  \item {MemQ achieves the highest mean held-out, final runtime, and cumulative success rates on all eight benchmarks. The largest held-out gains over the best baseline for each benchmark occur on DSBench and AutomationBench. Ablations on LiveCodeBench, BFCL, and DSBench support the benefit of propagating credit beyond directly retrieved memories. We further examine MemQ's learning dynamics, the evolution of memory Q-values, and its limitations.}
\end{enumerate}

\section{Related Works}
\label{sec:related_works}

\subsection{Self-Evolving Memory Agents}
\label{sec:related_memory_agents}

Early memory-augmented agents rely on fixed heuristic retrieval---embedding similarity or hand-crafted scores \citep{park2023generative, packer2023memgpt, shinn2023reflexion, zhao2024expel, wang2023voyager, zhong2023memorybank, kynoch2023recallm, sumers2023cognitive}---with no learning signal from task outcomes. More recent work makes memory \emph{self-evolving} along two paradigms. \emph{Parametric approaches} \citep{yan2025memoryr1, zhang2025memact, ma2026finemem, shen2026membuilder, zhang2026retroagent, zhou2025mem1, yue2026memt} learn network parameters for memory operations, requiring gradient-based optimization. \emph{Non-parametric approaches} avoid weight updates, attaching value estimates or update rules directly to memory entries: REMEMBERER \citep{zhang2023rememberer} equips a frozen LLM with experience memory updated via RL without weight modification; MemRL \citep{zhang2026memrl} further attaches Q-values with single-step EMA ($\gamma = 0$); Memento \citep{zhou2025memento} learns a case-selection policy with memory rewriting; other systems employ rule-based curation \citep{mem0}, cognitive self-organization \citep{nan2025nemori}, meta-evolution \citep{zhang2025memevolve}, learnable skills \citep{zhang2025memskill}, online experience weighting \citep{zhang2026liveevo}, Hebbian graphs \citep{zhu2026helamem}, utility-based pruning \citep{cao2025reme}, or procedural memory distillation \citep{fang2026mempexploringagentprocedural}. {Among the closest methods considered here, learning or selection remains local to a memory operation or current retrieval event. MemQ instead studies how delayed feedback can be propagated across recorded memory-creation exposures while retaining the non-parametric, frozen-LLM setting.}

\subsection{Reinforcement Learning for Memory}
\label{sec:related_rl}

RL for memory spans from episodic control methods \citep{blundell2016model, pritzel2017neural, lin2018episodic} and differentiable memory architectures \citep{graves2014neural, guu2020realm, schaul2016prioritized} to recent LLM-agent-specific methods. The parametric methods in \S\ref{sec:related_memory_agents} train neural policies for memory operations. On the non-parametric side, MemRL uses TD(0) with $\gamma = 0$, and Memento~2 \citep{wang2025memento2} optimizes retrieval via supervised learning within the Reflected MDP formalism---absorbing the frozen LLM into environment dynamics so that the retrieval policy becomes the sole decision variable and the memory store a valid MDP state. Other work applies RL to memory-augmented retrieval \citep{yuan2025memsearcher, ouyang2025reasoningbank, wei2025evomemory}, {focusing on what to store and when to retrieve.}
{Classical trace theory \citep{sutton1988learning, singh1996reinforcement, sutton2018reinforcement, peng1996incremental, vanseijen2014true, schulman2016gae} propagates delayed credit over temporal steps. MemQ transfers this update principle to a structural domain: traces follow recorded provenance exposures and use DAG depth as the decay distance.}

\section{Problem Formulation}
\label{sec:problem_formulation}

We consider a frozen LLM agent tackling tasks from an unknown distribution. Unable to learn via gradient updates, it relies on a continually growing episodic memory store. {The core challenge is assigning credit across multi-step memory chains: because early retrievals can indirectly enable future success, a principled solution must account for the downstream impact of every retrieval on future rewards.}

\paragraph{Exogenous-Context MDP.}
A key structural feature of this setting is that the state factors into two components with fundamentally different dynamics: the \emph{exogenous} task stream, which the agent cannot influence, and the \emph{endogenous} memory store, which evolves as a direct consequence of the agent's retrieval actions. We make this factorization explicit by defining the \emph{Exogenous-Context MDP} (EC-MDP).

\begin{definition}[Exogenous-Context MDP]
\label{def:ec_mdp}
The EC-MDP is a tuple $\langle \mathcal{S}, \mathcal{M}, \mathcal{A}, P_\mathrm{exo}, P_\mathrm{endo}, R, \gamma \rangle$ where $\mathcal{S}$ is the \textbf{exogenous state space} (the set of all tasks) with evolution governed by $P_\mathrm{exo}(s_{t+1}) = \rho(s_{t+1})$, independent of the agent's actions or memory; $\mathcal{M} \subseteq 2^{\mathcal{M}_\infty}$\footnote{Intuitively, since the memory bank is a collection of past interaction experiences, any valid memory state $\mathcal{M}$ is a finite subset of the universe of all theoretically possible experiences $\mathcal{M}_\infty$.} is the \textbf{endogenous state space} (the memory store), whose evolution is fully determined by the current task, memory, and retrieval action; $\mathcal{A}(\mathcal{M}) = \{A \subseteq \mathcal{M} : |A| \leq k\}$ is the \textbf{action space} of retrieval subsets of size at most $k$; $P_\mathrm{endo}(\mathcal{M}_{t+1} \mid s_t, \mathcal{M}_t, A_t)$ is the \textbf{endogenous transition kernel}, absorbing the frozen LLM:
\[
P_\mathrm{endo}(\mathcal{M}_{t+1} \mid s_t, \mathcal{M}_t, A_t) = \sum_{\tau} \pi_\mathrm{LLM}(\tau \mid s_t, A_t) \cdot \mathbf{1}\!\bigl[\mathcal{M}_{t+1} = \mathcal{M}_t \cup \{\mathrm{Build}(s_t, \tau)\}\bigr];
\]
$R(s_t, A_t) = \mathbb{E}_{\tau \sim \pi_\mathrm{LLM}(\cdot \mid s_t, A_t)}[r(\tau)]$ is the \textbf{reward}; and $\gamma \in [0,1]$ is the discount factor. A retrieval policy $\pi_\mathrm{ret}(A \mid s, \mathcal{M})$ serves as the agent's \emph{sole optimized policy}: while $\pi_\mathrm{LLM}$ remains frozen, $\pi_\mathrm{ret}$ maps the current task and memory store to a distribution over retrieval actions.
\end{definition}

{This setting models benchmark instances, independent user queries, and externally scheduled tasks, whose distribution is independent of the current retrieval action.}

\begin{figure}[t]
  \centering
  \includegraphics[width=\linewidth]{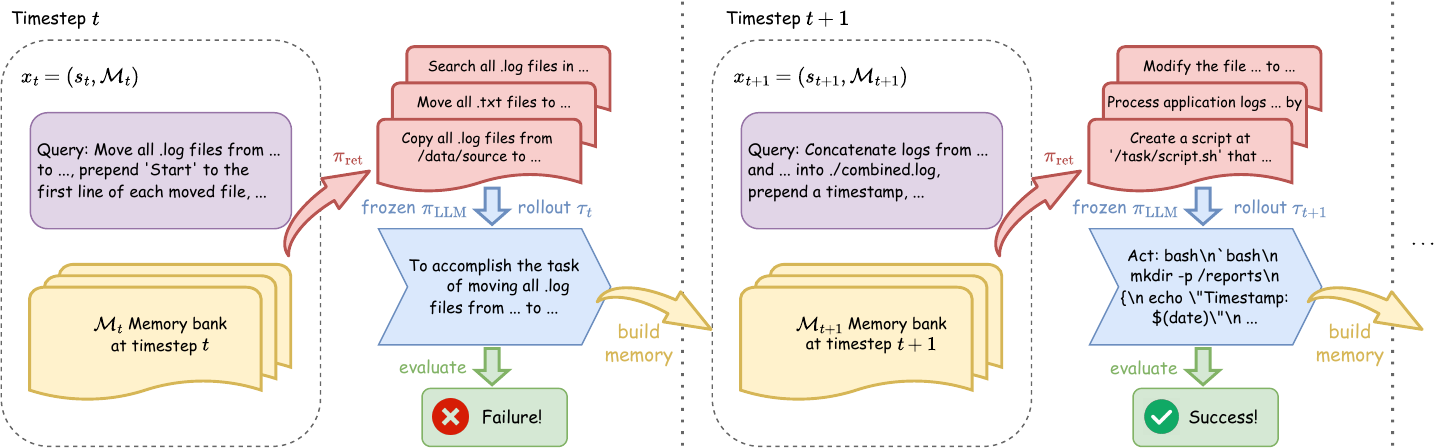}
  \caption{The EC-MDP. The state factors into an exogenous task stream $s_t \sim \rho$ and an endogenous memory store $\mathcal{M}_t$. The retrieval policy $\pi_\mathrm{ret}$ selects memories $m_t$, the frozen agent $\pi_\mathrm{LLM}$ produces a trajectory $\tau_t$ and receives reward $r_t$, and the memory is built and stored.}
  \label{fig:reflected_mdp}
\end{figure}

{The EC-MDP joint transition kernel factors as:}
\begin{equation}
\label{eq:factored_transition}
P(s_{t+1}, \mathcal{M}_{t+1} \mid s_t, \mathcal{M}_t, A_t) = P_\mathrm{exo}(s_{t+1}) \cdot P_\mathrm{endo}(\mathcal{M}_{t+1} \mid s_t, \mathcal{M}_t, A_t).
\end{equation}
{Relative to the Reflected-MDP perspective of \citet{wang2025memento2}, this factorization separates the externally generated task from the memory state changed by retrieval. Monotonic memory growth ($\mathcal{M}_{t+1} \supseteq \mathcal{M}_t$), together with the current task and memory store in the state, yields a Markov process under the frozen-LLM transition. The stochastic reward $r_t=r(\tau_t)$ and newly constructed memory $m_\mathrm{new}=\mathrm{Build}(s_t,\tau_t)$ then depend on the current task and retrieved set:}
\begin{equation}
\label{eq:conditional_independence}
P(r_t, m_\mathrm{new} \mid s_t, \mathcal{M}_t, A_t) = P(r_t, m_\mathrm{new} \mid s_t, A_t).
\end{equation}

\paragraph{Value functions and learning objective.}
The \emph{state-value} captures the long-term value of the memory store via the expected cumulative discounted reward:
\[
V^{\pi_\mathrm{ret}}(\mathcal{M}_t) = \mathbb{E}_{\substack{s_k \sim \rho, \, A_k \sim \pi_\mathrm{ret}, \, \tau_k \sim \pi_\mathrm{LLM}}}\!\biggl[\sum_{k=0}^{\infty} \gamma^k r(\tau_{t+k}) \,\bigg|\, \mathcal{M}_t\biggr].
\]
The \emph{action-value} for retrieving set $A$ is $Q(s, A;\mathcal{M}) = \mathbb{E}_{\tau}\!\bigl[r(\tau) + \gamma\, V^{\pi_\mathrm{ret}}(\mathcal{M}') \,\big|\, s, A, \mathcal{M}\bigr]$, where $\mathcal{M}' = \mathcal{M} \cup \{m_\mathrm{new}(\tau)\}$. The learning objective is thus to maximize the initial state value: $$\max_{\pi_\mathrm{ret}} V^{\pi_\mathrm{ret}}(\mathcal{M}_0).$$
{Directly representing the action value of every retrieval subset is intractable. We therefore use a parameter-free first-order projection onto per-memory values, while retaining an explicit residual for interactions within the retrieved set:}
\begingroup
\begin{equation}
\label{eq:value_decomposition}
Q^\pi(s,A;\mathcal{M})
= \frac{1}{|A|}\sum_{m\in A}q^\pi(m)
+ \xi^\pi(s,A;\mathcal{M}),
\quad A\sim\pi_\mathrm{ret}(\cdot\mid s,\mathcal{M}).
\end{equation}
\endgroup
{The residual $\xi^\pi$ captures synergy, redundancy, and distraction among co-retrieved memories. MemQ uses the first-order approximation $Q^\pi(s,A;\mathcal M)\approx |A|^{-1}\sum_{m\in A}q^\pi(m)$ and learns $Q(m)$ as an estimate of the per-memory term $q^\pi(m)$. Appendix~\ref{app:set_value_approximation} gives the approximation-error and one-step selection-regret bounds.}

\section{Method}
\label{sec:method}

\begin{figure}[t]
\centering
\includegraphics[width=\linewidth]{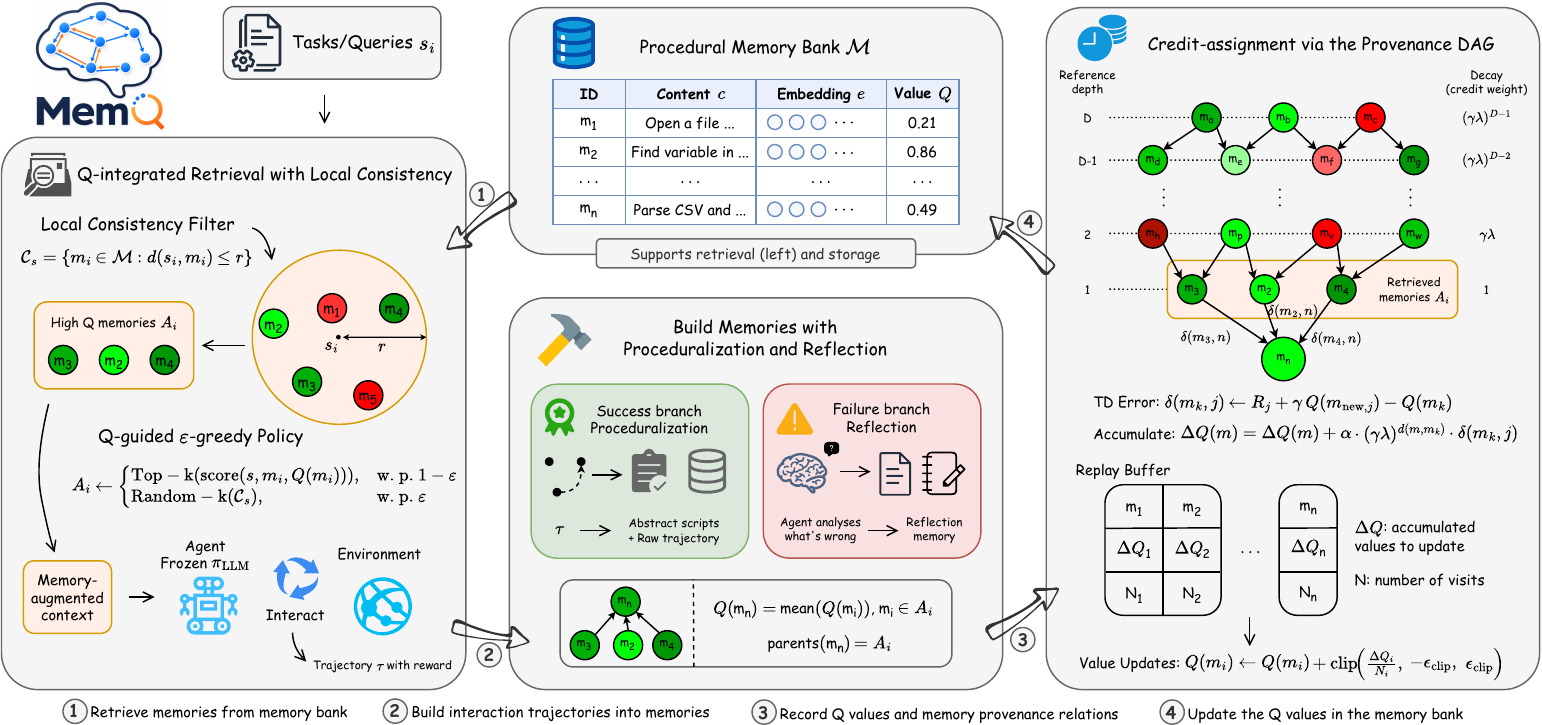}
\caption{MemQ Framework Overview. The continuous learning loop features three stages: Retrieve: Following locality filtering, an $\epsilon$-greedy policy selects memories via learned $Q$-values to contextualize the LLM. Build: Task trajectories are distilled into new memories (via proceduralization or reflection) and integrated into the provenance DAG, linked to their parents. Update: TD errors backpropagate through the DAG to assign multi-step credit and update the $Q$-values of all contributing ancestors. }
\label{fig:framework}
\end{figure}

Section~\ref{sec:problem_formulation} formulated retrieval as an EC-MDP whose action-value decomposes over individual memories (Eq.~\ref{eq:value_decomposition}). MemQ is a method for learning per-memory Q-values in this setting. Its key idea is that when a retrieval set $A$ is used for a task and produces a new memory $m_\mathrm{new}$, we record $\mathrm{parents}(m_\mathrm{new}) = A$, inducing a \emph{provenance DAG}, i.e., a directed acyclic graph over memory entries whose edges indicate ``was retrieved for a task that produced the next memory.'' This DAG provides the structural substrate for multi-step credit assignment: instead of updating only the directly retrieved memories, MemQ propagates TD($\lambda$) eligibility traces backward through the DAG, crediting ancestor memories proportionally to their structural depth from the outcome. We describe MemQ in three parts: the Q-augmented memory store, a two-phase retrieval policy, and credit propagation via the provenance DAG.

\begin{algorithm}[t]
\caption{{MemQ with batch-parallel memory construction and every-path provenance credit propagation}}
\label{alg:memrl2}
\begingroup
\begin{algorithmic}[1]
\Require Task set $\mathcal{D}$, memory store $\mathcal{M}$, frozen agent $\pi_\mathrm{LLM}$
\Require Hyperparameters: $\alpha, \gamma, \lambda, \varepsilon, k, \theta_\mathrm{sim}, w_s, w_q, \epsilon_\mathrm{clip}$, empty-parent value $Q_0$
\State Initialize DAG $\gets \emptyset$; \;$\Delta Q_i \gets 0$; \;$N_i \gets 0$ for all $m_i$
\For{each epoch $\ell = 1, 2, \ldots$}
    \State Shuffle $\mathcal{D}$ into mini-batches $\mathcal{B}_1, \ldots, \mathcal{B}_L$ of size Batch size
    \State $\mathcal{T} \gets \emptyset$ \Comment{transition buffer}
    \For{each mini-batch $\mathcal{B}_b = \{s_1, \ldots, s_B\}$}
        \State \textcolor{blue}{// Trajectory sampling}
        \For{each task $s_j \in \mathcal{B}_b$} \Comment{in parallel}
            \State $\mathcal{C}_j \gets \{m \in \mathcal{M} : \mathrm{sim}(\phi(s_j), \mathbf{e}_m) \geq \theta_\mathrm{sim}\}$ \Comment{locality filter}
            \State $\mathrm{score}(m) \gets w_s \cdot \mathrm{sim}(s_j, m) + w_q \cdot Q(m)$ for all $m \in \mathcal{C}_j$ \Comment{Q-guided scoring}
            \State $A_j \gets \varepsilon\text{-greedy top-}k$ from $\mathcal{C}_j$
            \For{each step $t = 1, 2, \ldots$} \Comment{agent-environment interaction}
                \State $a_t \gets \pi_\mathrm{LLM}(s_j, A_j, h_{<t})$; \;$o_t \gets \mathrm{Environment}(a_t)$
            \EndFor
            \State $R_j \gets \mathrm{Environment}.\mathrm{reward}()$
        \EndFor
        \For{each completed record $(s_j,\tau_j,A_j,R_j)$} \Comment{batch-parallel build}
            \State $c(m_{\mathrm{new},j}) \gets \mathrm{Build}(s_j, \tau_j)$
            \State $Q(m_{\mathrm{new},j}) \gets |A_j|^{-1}\sum_{m_i\in A_j}Q(m_i)$ if $|A_j|>0$, else $Q_0$
            \State $\mathcal{M} \gets \mathcal{M} \cup \{m_{\mathrm{new},j}\}$; \;$\mathcal{T} \gets \mathcal{T} \cup \{(A_j,m_{\mathrm{new},j},R_j)\}$
        \EndFor
    \EndFor
    \State Record $\mathrm{parents}(m_{\mathrm{new},j}) \gets A_j$ for each transition in $\mathcal{T}$ \Comment{construct DAG}
    \State \textcolor{blue}{// Value update using MemQ}
    \State Compute $\delta(m_0, j) \gets R_j + \gamma\, Q(m_{\mathrm{new},j}) - Q(m_0)$ for each $m_0 \in A_j$ \Comment{Eq.~\ref{eq:td_error}}
    \State Enumerate every backward path $p$ from each root $m_0$, up to depth $D$: \Comment{path-wise DAG traversal}
    \For{each path $p$ ending at ancestor $m$}
        \State $\Delta Q(m) \gets \Delta Q(m) + \alpha \cdot (\gamma\lambda)^{|p|} \cdot \delta(m_0, j)$; \;$N_m \gets N_m + 1$ \Comment{one contribution per path}
    \EndFor
    \State $Q(m_i) \gets Q(m_i) + \mathrm{clip}\!\left(\frac{\Delta Q_i}{N_i},\, -\epsilon_\mathrm{clip},\, \epsilon_\mathrm{clip}\right)$ for all $m_i$; \;$\Delta Q_i \gets 0$; \;$N_i \gets 0$ \Comment{flush}
\EndFor
\end{algorithmic}
\endgroup
\end{algorithm}

\subsection{Procedural Construction of Q-Augmented Memory}
\label{sec:memory_construction}
Let $\mathcal{M} = \{m_1, m_2, \ldots, m_N\}$ denote the episodic memory store, where $N$ grows as the agent accumulates experience. Each memory entry is a tuple
\[
m_i = (c_i, \mathbf{e}_i, Q_i),
\]
where $c_i \in \Sigma^*$ is the memory content (natural language text encoding a past experience), $\mathbf{e}_i = \phi(c_i) \in \mathbb{R}^d$ is its retrieval embedding computed by a frozen encoder $\phi$, and $Q_i \in \mathbb{R}$ is the \emph{Q-value}, the agent's current estimate of $m_i$'s value when retrieved for a relevant task.

{After each interaction, the system constructs a memory from the sampled trajectory. The builder applies \emph{proceduralization}: an LLM distills $\tau$ into an abstract 3--5 step script and stores both the script and raw trajectory. Successful trajectories are stored directly; after failure, an LLM produces a reflection that is stored separately. To mitigate cold start, the new memory inherits the uniform mean Q-value of its retrieved parents. This estimate is parameter-free, available directly from the recorded parent set, and consistent with the first-order term in Eq.~\ref{eq:value_decomposition}:}
\[
Q(m_{\mathrm{new}}) = \frac{1}{|A_j|} \sum_{m_i \in A_j} Q(m_i),
\]
where $A_j$ is the retrieved set for task $j$. When no memories are retrieved ($A_j = \emptyset$), we set $Q(m_{\mathrm{new}})=Q_0$.

\subsection{Q-integrated Retrieval Policy within Local Consistency}
\label{sec:retrieval_policy}
The retrieval policy decomposes the action selection $A \subseteq \mathcal{M}$ into two phases addressing orthogonal concerns: \emph{locality}, ensuring that retrieved memories lie within a neighborhood where the LLM can reliably generalize, and \emph{Q-guided selection}, selecting among those local candidates the memories whose RL-learned Q-values indicate high long-run value.

\paragraph{Locality filtering.} Capable LLMs can act near-optimally by interpolating from similar past experiences, eliminating the need for exact memory matches. This relies on the \emph{LLM local consistency} property \citep{wang2025memento2}: if a retrieved memory lies within distance $r$ (the \emph{radius of competence}) of the current state $s$, the LLM's optimality gap $\varepsilon_{\mathrm{LLM}}(r)$ remains small. Locality filtering operationalizes this by identifying the local neighborhood of $s$. While any state-space metric suffices, we instantiate it using cosine similarity in embedding space: $\mathrm{sim}(s, m_i) = \phi(s)^\top \mathbf{e}_i / (\|\phi(s)\| \cdot \|\mathbf{e}_i\|)$. This forms the candidate set $\mathcal{C}_s = \{m_i \in \mathcal{M} : \mathrm{sim}(s, m_i) \geq \theta_\mathrm{sim}\}$, where the threshold $\theta_\mathrm{sim}$ represents the radius of competence in similarity space. If no memories meet this threshold, the task is considered novel, and the agent defaults to a zero-shot fallback.

\paragraph{Q-guided selection.} Once locality filtering ensures all candidates fall within the LLM's radius of competence, the second step leverages reinforcement learning to discriminate among them. We combine raw similarity scores in $[0, 1]$ with Q-values normalized to $[0, 1]$. Each candidate is then scored by a composite function:
\[
\mathrm{score}(s, m_i) = w_s \cdot \mathrm{sim}(s, m_i) + w_q \cdot Q(m_i),
\]
where $w_s$ and $w_q$ control the trade-off between relevance and learned value. The Q-values $Q(m_i)$ are learned via the TD updates developed below, capturing long-run value beyond surface-level similarity. Selection follows an $\varepsilon$-greedy policy:
\[
A = \begin{cases}
\mathrm{Top}\text{-}k \;\text{by}\; \mathrm{score}(s, m_i) & \text{with probability } 1 - \varepsilon, \\
\text{Uniform sample of } k \text{ from } \mathcal{C}_s & \text{with probability } \varepsilon,
\end{cases}
\]
ensuring continued exploration of currently undervalued memories.

\subsection{Provenance-Based Credit Assignment}
\label{sec:credit_propagation}

In classical TD($\lambda$) \citep{sutton2018reinforcement, sutton1988learning}, the \emph{$\lambda$-return} $G_t^\lambda = (1{-}\lambda)\sum_{n=1}^{\infty} \lambda^{n-1} G_t^{(n)}$ interpolates between the one-step bootstrap $G_t^{(1)} = r_t + \gamma\,Q(s_{t+1}, a_{t+1})$ and the Monte Carlo return $G_t^{(\infty)} = \sum_{k=0}^{\infty} \gamma^k r_{t+k}$. The $\lambda$-return advantage is standardly expressed as a discounted sum of future TD errors:
\begin{equation}
\label{eq:telescoping}
G_t^\lambda - Q(s_t, a_t) = \sum_{k=0}^{T-t-1} (\gamma\lambda)^k \,\delta_{t+k},
\end{equation}
{where $\delta_t = r_t + \gamma\,Q(s_{t+1}, a_{t+1}) - Q(s_t, a_t)$. MemQ applies the geometric weighting in this identity to backward provenance paths, replacing elapsed time with DAG depth. The resulting structural update is analyzed directly: Appendix~\ref{app:convergence} proves convergence and fixed-point preservation in a fixed-DAG evaluation setting.}

\paragraph{Per-memory TD error.}
For task $s_j$ with retrieved set $A_j$, outcome $R_j$, and constructed memory $m_{\mathrm{new},j}$, the per-memory TD error is
\begin{equation}
\label{eq:td_error}
\delta(m_i, j) = R_j + \gamma\, Q(m_{\mathrm{new},j}) - Q(m_i), \quad \forall\, m_i \in A_j.
\end{equation}
All $m_i \in A_j$ share the same reward and bootstrap target but differ in their current Q-value estimates. Setting $\gamma = 0$ discards the bootstrap term $\gamma Q(m_\mathrm{new})$ and reduces the update to single-step exponential moving average, crediting each memory only with the immediate reward.

\paragraph{Credit propagation via the provenance DAG.}
{Single-step updates reach only directly retrieved memories. MemQ instead traverses the exposure-based provenance DAG, which records $\mathrm{parents}(m_{\mathrm{new}})=A$ for each constructed memory. For task $s_j$, it computes $\delta(m_0,j)$ for every $m_0\in A_j$ (Eq.~\ref{eq:td_error}) and enumerates every backward provenance path from each root to propagate that observed error to reachable ancestors:}
\begingroup
\begin{equation}
\label{eq:general_update}
\Delta Q(m) \mathrel{+}=
\alpha \sum_{m_0 \in A_j}
\sum_{p\in\mathcal{P}_D(m_0\rightsquigarrow m)}
(\gamma\lambda)^{|p|}\,\delta(m_0,j),
\end{equation}
\endgroup
{Here $\mathcal{P}_D(m_0\rightsquigarrow m)$ is the set of all distinct backward provenance paths from root $m_0$ to ancestor $m$ with length at most $D$; it includes the length-zero path when $m=m_0$. Every path contributes separately according to its own length $|p|$. The counter $N_m$ is incremented once per path contribution. The factor $(\gamma\lambda)^{|p|}$ transfers the temporal decay in Eq.~\ref{eq:telescoping} to structural distance. Enumeration is bounded by depth $D$ and prunes a path when $(\gamma\lambda)^{|p|}<10^{-12}$. Updates are accumulated within each epoch with count $N_i$ and then flushed:}
\[
Q(m_i) \leftarrow Q(m_i) + \mathrm{clip}\!\Bigl(\frac{\Delta Q(m_i)}{N_i},\; -\epsilon_\mathrm{clip},\; \epsilon_\mathrm{clip}\Bigr).
\]

\section{Experiments}
\label{sec:experiments}

{We evaluate MemQ on eight benchmarks and organize the analysis around four research questions: (RQ1) Does MemQ improve held-out performance and runtime learning on interaction, code-generation, and agentic tasks? (RQ2) How does provenance credit assignment affect performance? (RQ3) How does MemQ use memory during runtime learning? (RQ4) When are MemQ's gains limited?}

\subsection{Experimental Setup}
\label{sec:experimental_setup}

\paragraph{Baselines.}
We compare against six baselines. \textbf{No Memory}: a frozen LLM without retrieved context. \textbf{RAG}: retrieves the top-$k$ memories via cosine similarity without quality re-ranking. \textbf{Self-RAG}~\citep{asai2024selfrag}: generates queries and filters unhelpful retrieved memories via self-evaluation. \textbf{Mem0}~\citep{mem0}: manages episodic memory lifecycles through extraction, updates, and deletion. \textbf{MemP}~\citep{fang2026mempexploringagentprocedural}: distills past trajectories into step-by-step instructions and script-like abstractions as procedural memory, with dynamic update, correction, and deprecation rules. \textbf{MemRL}~\citep{zhang2026memrl}: a closely related approach that updates per-memory utility values via single-step EMA without eligibility traces.

\paragraph{{Benchmarks and backbones.}}
{We split each benchmark into memory-accumulation and held-out sets. LifeLongAgentBench (LLAB)~\citep{zheng2026lifelongagentbench} tests multi-step OS-level planning; BFCL v3~\citep{patil2025bfcl} evaluates multi-turn function calling; LiveCodeBench v6~\citep{jain2025livecodebench} covers code generation; MMMU Pro~\citep{yue-etal-2025-mmmu} tests multimodal reasoning; ERQA~\citep{geminiroboticsteam2025geminiroboticsbringingai} evaluates physically grounded reasoning; and GPQA~\citep{rein2024gpqa} contains expert-level science questions. DSBench~\citep{jing2024dsbench} and AutomationBench~\citep{automationbench2026} add tool-mediated data-science and workflow-automation tasks. All methods use the same frozen backbone and data split within each benchmark. We use GPT-4o-mini for LLAB, Qwen3.5-35B-A3B with FP8 quantization~\citep{qwen3.5} for BFCL, DeepSeek-v4-flash-0731 (284B)~\citep{deepseekv4flash} for DSBench and AutomationBench, and Gemma-4-E4B-it~\citep{gemma_4_e4b_it} for the remaining four benchmarks. We analyze LLAB, LiveCodeBench, BFCL, DSBench, and AutomationBench as interaction, code-generation, and agentic tasks, and MMMU Pro, ERQA, and GPQA as multiple-choice reasoning tasks. We select backbones to meet each benchmark's modality requirements and explore how MemQ's gains vary with model scale.}

\subsection{\texorpdfstring{{RQ1: Does MemQ improve held-out performance and runtime learning on interaction, code-generation, and agentic tasks?}}{RQ1: Does MemQ improve held-out performance and runtime learning on interaction, code-generation, and agentic tasks?}}
\label{sec:results}

{We first evaluate MemQ on LLAB, LiveCodeBench, BFCL, DSBench, and AutomationBench, covering environment interaction, code generation, and agentic tasks. DSBench and AutomationBench extend the evaluation to tool-mediated data science and workflow automation. MemQ attains average held-out, final runtime, and cumulative SRs of 56.17\%, 71.50\%, and 78.88\%, respectively (Tables~\ref{tab:eval_results} and~\ref{tab:runtime_results}). The held-out average exceeds MemP, the strongest baseline on this metric, by 2.53~pp. MemRL is the strongest baseline for final runtime SR and CSR; the corresponding gains are 4.23~pp and 2.13~pp.}

\paragraph{{Held-out performance.}}
{MemQ's consistent held-out advantage indicates that its benefits extend to new tasks. Its two largest margins over each task's strongest baseline occur on DSBench and AutomationBench, both using the 284B DeepSeek backbone. These larger gains suggest that model capability and learned memory selection can be complementary in agentic tasks. A plausible explanation is that more capable models make better use of retrieved procedures: they can adapt past experience to new requirements and translate it into effective tool-use decisions, increasing the value of selecting useful memories.}

\begin{table}[t]
\centering
\caption{{Held-out success rate (Eval SR, \%) on interaction, code-generation, and agentic tasks. The best-epoch memory bank is selected on the memory-accumulation split before held-out evaluation. Entries are mean $\pm$ sample SD. Average across the five benchmarks is reported with approximate 95\% CIs below. Best in \textbf{bold}; second-best \underline{underlined}.}}
\label{tab:eval_results}
\begingroup
\setlength{\tabcolsep}{5pt}
\resizebox{\linewidth}{!}{%
\begin{tabular}{l | c | ccccccc}
\toprule
\textbf{Benchmark} & \textbf{Model} & \textbf{No Mem.} & \textbf{RAG} & \textbf{Self-RAG} & \textbf{Mem0} & \textbf{MemP} & \textbf{MemRL} & \emph{\textbf{MemQ}} \\
\midrule
\makecell[l]{LLAB\\[-3pt]{\scriptsize OS Interaction}} & \texttt{4o-mini} & 66.89$_{\pm 1.26}$ & 68.89$_{\pm 0.38}$ & 70.45$_{\pm 0.39}$ & 70.00$_{\pm 0.67}$ & 71.11$_{\pm 0.38}$ & \underline{74.44$_{\pm 1.02}$} & \textbf{74.67$_{\pm 0.67}$} \\[3pt]
\midrule
\makecell[l]{LiveCodeBench\\[-3pt]{\scriptsize Coding}} & \makecell[c]{\texttt{Gemma-}\\[-3pt]\texttt{4-E4B-it}} & 44.76$_{\pm 2.29}$ & \underline{50.48$_{\pm 1.65}$} & 49.52$_{\pm 1.65}$ & 47.62$_{\pm 1.65}$ & 49.52$_{\pm 1.65}$ & 45.71$_{\pm 2.86}$ & \textbf{51.43$_{\pm 2.86}$} \\[3pt]
\midrule
\makecell[l]{BFCL\\[-3pt]{\scriptsize Function Call}} & \makecell[c]{\texttt{Qwen3.5-}\\[-3pt]\texttt{35B-A3B}} & 56.71$_{\pm 1.71}$ & 55.45$_{\pm 2.62}$ & 60.07$_{\pm 2.49}$ & 54.79$_{\pm 1.68}$ & \underline{61.39$_{\pm 0.99}$} & 60.07$_{\pm 2.29}$ & \textbf{62.38$_{\pm 1.71}$} \\[3pt]
\midrule
\makecell[l]{DSBench\\[-3pt]{\scriptsize Data Science}} & \multirow{2}{*}[-1.5ex]{\makecell[c]{\texttt{DeepSeek-v4-}\\[-3pt]\texttt{flash-0731}}} & 43.97$_{\pm 0.70}$ & 41.09$_{\pm 1.79}$ & 43.97$_{\pm 2.28}$ & \underline{45.40$_{\pm 0.50}$} & 44.54$_{\pm 3.98}$ & 42.53$_{\pm 0.99}$ & \textbf{47.70$_{\pm 0.50}$} \\[3pt]
\cmidrule{1-1}\cmidrule{3-9}
\makecell[l]{AutomationBench\\[-3pt]{\scriptsize Workflow Automation}} & & 30.28$_{\pm 1.74}$ & 41.25$_{\pm 1.77}$ & 39.58$_{\pm 0.59}$ & 33.33$_{\pm 2.36}$ & 41.67$_{\pm 2.36}$ & \underline{43.17$_{\pm 2.36}$} & \textbf{44.69$_{\pm 2.36}$} \\[3pt]
\midrule
\multicolumn{2}{l|}{\textbf{Average}} & \makecell[c]{48.52\\[-2pt]{\scriptsize [47.54, 49.51]}} & \makecell[c]{51.43\\[-2pt]{\scriptsize [50.33, 52.53]}} & \makecell[c]{52.72\\[-2pt]{\scriptsize [51.63, 53.80]}} & \makecell[c]{50.23\\[-2pt]{\scriptsize [49.26, 51.20]}} & \makecell[c]{\underline{53.65}\\[-2pt]{\scriptsize \underline{[52.09, 55.20]}}} & \makecell[c]{53.18\\[-2pt]{\scriptsize [51.93, 54.44]}} & \makecell[c]{\textbf{56.17}\\[-2pt]{\scriptsize \textbf{[54.98, 57.37]}}} \\
\bottomrule
\end{tabular}%
}
\endgroup
\end{table}

\paragraph{{Runtime learning.}}
{MemQ also leads final SR and CSR on each of these benchmarks. Relative to MemRL, its final-SR gains reach 5.83~pp on LiveCodeBench, 3.77~pp on BFCL, 3.52~pp on DSBench, and 4.82~pp on AutomationBench. MemQ reaches a CSR of 88.67\% on DSBench, compared with Self-RAG's 88.29\%, and 67.29\% on AutomationBench, compared with 65.60\% for Mem0. The evolution of SR and CSR during runtime learning across all eight benchmarks is shown in Appendix~\ref{app:csr_curves}.}

\begin{table}[t]
\centering
\caption{{Runtime learning results (\%). Within each benchmark, rows show final-epoch success rate (Final SR) and cumulative success rate (CSR), respectively, on the memory-accumulation split. CSR is the fraction of tasks solved at least once. Entries are mean $\pm$ sample SD. Average rows cover the five benchmarks, with approximate 95\% CIs below. Best in \textbf{bold}; second-best \underline{underlined}.}}
\label{tab:runtime_results}
\begingroup
\setlength{\tabcolsep}{5pt}
\resizebox{\linewidth}{!}{%
\begin{tabular}{l | c | ccccccc}
\toprule
\textbf{Benchmark} & \textbf{Model} & \textbf{No Mem.} & \textbf{RAG} & \textbf{Self-RAG} & \textbf{Mem0} & \textbf{MemP} & \textbf{MemRL} & \emph{\textbf{MemQ}} \\
\midrule
\multirow{2}{*}{\makecell[l]{LLAB\\[-3pt]{\scriptsize OS Interaction}}} & \multirow{2}{*}{\texttt{4o-mini}} & 63.60$_{\pm 0.28}$ & 65.20$_{\pm 1.14}$ & 62.33$_{\pm 0.62}$ & 62.53$_{\pm 0.25}$ & 76.07$_{\pm 0.52}$ & \underline{77.13$_{\pm 0.68}$} & \textbf{80.34$_{\pm 2.22}$} \\
 & & -- & 68.47$_{\pm 0.84}$ & 75.27$_{\pm 0.74}$ & 78.40$_{\pm 0.43}$ & 79.27$_{\pm 0.52}$ & \underline{80.80$_{\pm 0.91}$} & \textbf{82.27$_{\pm 1.88}$} \\
\midrule
\multirow{2}{*}{\makecell[l]{LiveCodeBench\\[-3pt]{\scriptsize Coding}}} & \multirow{2}{*}{\makecell[c]{\texttt{Gemma-}\\[-3pt]\texttt{4-E4B-it}}} & 47.14$_{\pm 1.43}$ & 51.96$_{\pm 2.11}$ & 45.89$_{\pm 2.78}$ & 49.52$_{\pm 1.68}$ & 53.04$_{\pm 0.78}$ & \underline{58.93$_{\pm 0.62}$} & \textbf{64.76$_{\pm 1.49}$} \\
 & & -- & 57.38$_{\pm 0.89}$ & 65.00$_{\pm 2.42}$ & 56.19$_{\pm 0.34}$ & 55.48$_{\pm 0.89}$ & \underline{66.19$_{\pm 1.47}$} & \textbf{69.76$_{\pm 2.18}$} \\
\midrule
\multirow{2}{*}{\makecell[l]{BFCL\\[-3pt]{\scriptsize Function Call}}} & \multirow{2}{*}{\makecell[c]{\texttt{Qwen3.5-}\\[-3pt]\texttt{35B-A3B}}} & 54.52$_{\pm 1.30}$ & 55.28$_{\pm 2.31}$ & 64.91$_{\pm 0.78}$ & 57.37$_{\pm 0.83}$ & 65.91$_{\pm 1.20}$ & \underline{71.27$_{\pm 0.47}$} & \textbf{75.04$_{\pm 0.31}$} \\
 & & -- & 70.77$_{\pm 0.31}$ & 80.99$_{\pm 1.33}$ & 73.62$_{\pm 1.28}$ & 80.15$_{\pm 1.14}$ & \underline{85.85$_{\pm 1.44}$} & \textbf{86.43$_{\pm 1.13}$} \\
\midrule
\multirow{2}{*}{\makecell[l]{DSBench\\[-3pt]{\scriptsize Data Science}}} & \multirow{4}{*}{\makecell[c]{\texttt{DeepSeek-v4-}\\[-3pt]\texttt{flash-0731}}} & 58.36$_{\pm 1.03}$ & 66.38$_{\pm 4.54}$ & 63.81$_{\pm 2.33}$ & 66.19$_{\pm 1.75}$ & 64.67$_{\pm 1.72}$ & \underline{68.00$_{\pm 0.58}$} & \textbf{71.52$_{\pm 0.44}$} \\
 & & -- & 82.38$_{\pm 3.06}$ & \underline{88.29$_{\pm 0.29}$} & 87.43$_{\pm 0.29}$ & 83.43$_{\pm 2.23}$ & 86.76$_{\pm 1.15}$ & \textbf{88.67$_{\pm 1.00}$} \\
\cmidrule{1-1}\cmidrule{3-9}
\multirow{2}{*}{\makecell[l]{AutomationBench\\[-3pt]{\scriptsize Workflow Automation}}} & & 28.75$_{\pm 0.42}$ & 46.46$_{\pm 1.30}$ & 44.79$_{\pm 1.77}$ & 46.56$_{\pm 0.64}$ & 58.96$_{\pm 0.59}$ & \underline{61.01$_{\pm 1.03}$} & \textbf{65.83$_{\pm 1.53}$} \\
 & & -- & 50.63$_{\pm 0.81}$ & 58.02$_{\pm 1.15}$ & \underline{65.60$_{\pm 1.83}$} & 63.33$_{\pm 0.21}$ & 64.17$_{\pm 0.93}$ & \textbf{67.29$_{\pm 1.84}$} \\
\midrule
\multicolumn{2}{l|}{\textbf{Average (Final SR)}} & \makecell[c]{50.47\\[-2pt]{\scriptsize [49.86, 51.09]}} & \makecell[c]{57.06\\[-2pt]{\scriptsize [55.30, 58.81]}} & \makecell[c]{56.35\\[-2pt]{\scriptsize [55.17, 57.52]}} & \makecell[c]{56.43\\[-2pt]{\scriptsize [55.67, 57.20]}} & \makecell[c]{63.73\\[-2pt]{\scriptsize [63.05, 64.41]}} & \makecell[c]{\underline{67.27}\\[-2pt]{\scriptsize \underline{[66.84, 67.69]}}} & \makecell[c]{\textbf{71.50}\\[-2pt]{\scriptsize \textbf{[70.59, 72.40]}}} \\
\multicolumn{2}{l|}{\textbf{Average (CSR)}} & -- & \makecell[c]{65.93\\[-2pt]{\scriptsize [64.68, 67.17]}} & \makecell[c]{73.51\\[-2pt]{\scriptsize [72.57, 74.45]}} & \makecell[c]{72.25\\[-2pt]{\scriptsize [71.51, 72.98]}} & \makecell[c]{72.33\\[-2pt]{\scriptsize [71.46, 73.20]}} & \makecell[c]{\underline{76.75}\\[-2pt]{\scriptsize \underline{[76.05, 77.46]}}} & \makecell[c]{\textbf{78.88}\\[-2pt]{\scriptsize \textbf{[77.89, 79.88]}}} \\
\bottomrule
\end{tabular}%
}
\endgroup
\end{table}

\subsection{\texorpdfstring{{RQ2: How does provenance credit assignment affect performance?}}{RQ2: How does provenance credit assignment affect performance?}}
\label{sec:controlled_attribution}
\label{sec:ablations}

{{We compare full MemQ with variants without bootstrapping or multi-depth propagation on LiveCodeBench, BFCL, and DSBench (Figure~\ref{fig:credit_assignment_ablation}).} The discount $\gamma$ weights future memory value in the bootstrap target, while $\lambda$ controls the decay of observed TD errors; together they give the ancestor weight $(\gamma\lambda)^d$ at provenance depth $d$. Setting $\gamma=0$ removes bootstrapping and all positive-depth contributions. Setting $\lambda=0$ retains the one-step bootstrap but stops the observed TD error from reaching earlier provenance ancestors. We also sweep each parameter while holding the other fixed.}

{Full MemQ achieves the highest point estimate on all three metrics across all three benchmarks (Figure~\ref{fig:credit_assignment_ablation}). The $\gamma=0$ variant reduces evaluation SR by 5.72~pp on LiveCodeBench and 2.43~pp on BFCL; the $\lambda=0$ variant reduces it by 8.57 and 4.41~pp. {On DSBench, the $\gamma=0$ variant falls below full MemQ by 3.52~pp in final SR and 3.73~pp in evaluation SR; the corresponding gaps for $\lambda=0$ are 5.71 and 4.88~pp. Both variants also have lower CSR. Since $\lambda=0$ retains the one-step bootstrap, its lower runtime and held-out scores support the value of propagating feedback beyond the directly retrieved memories.}}

\begin{figure}[t]
\centering
\includegraphics[width=0.32\linewidth]{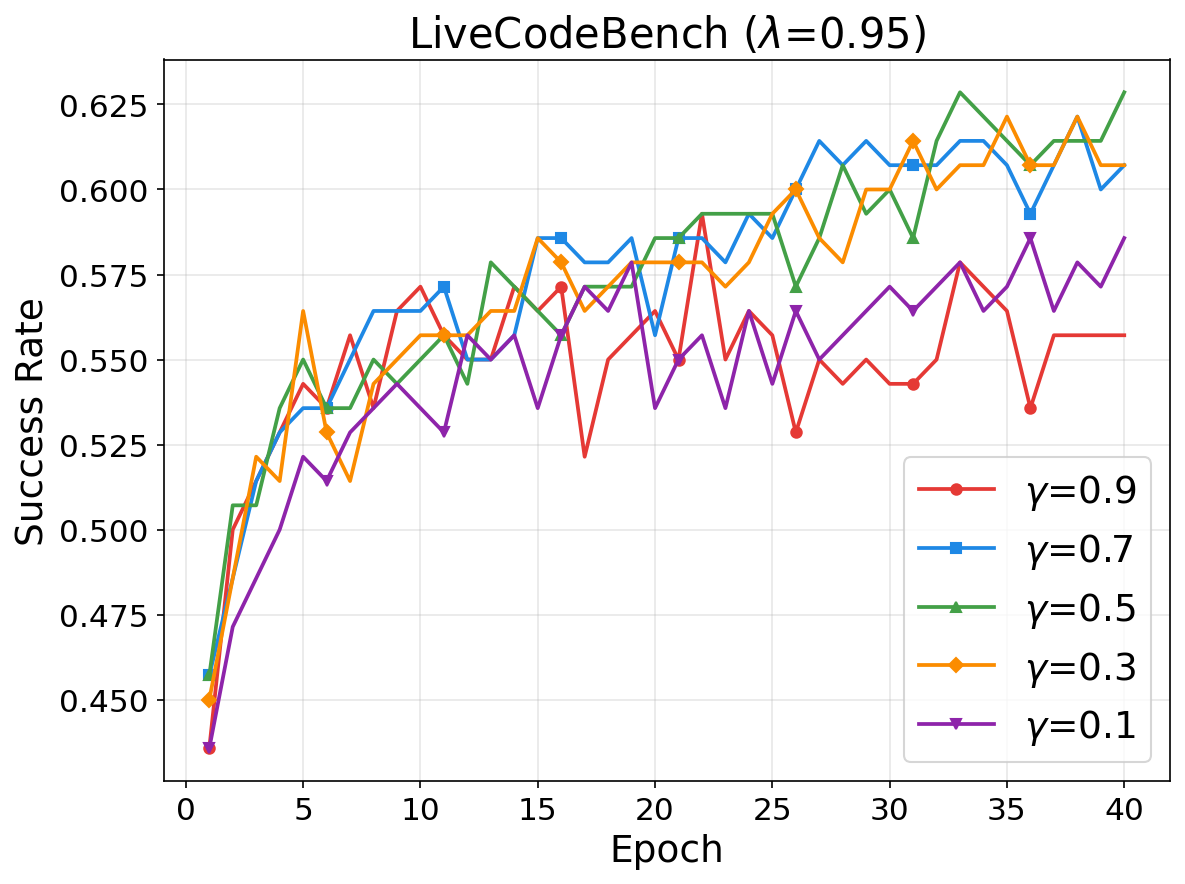}
\hfill
\includegraphics[width=0.32\linewidth]{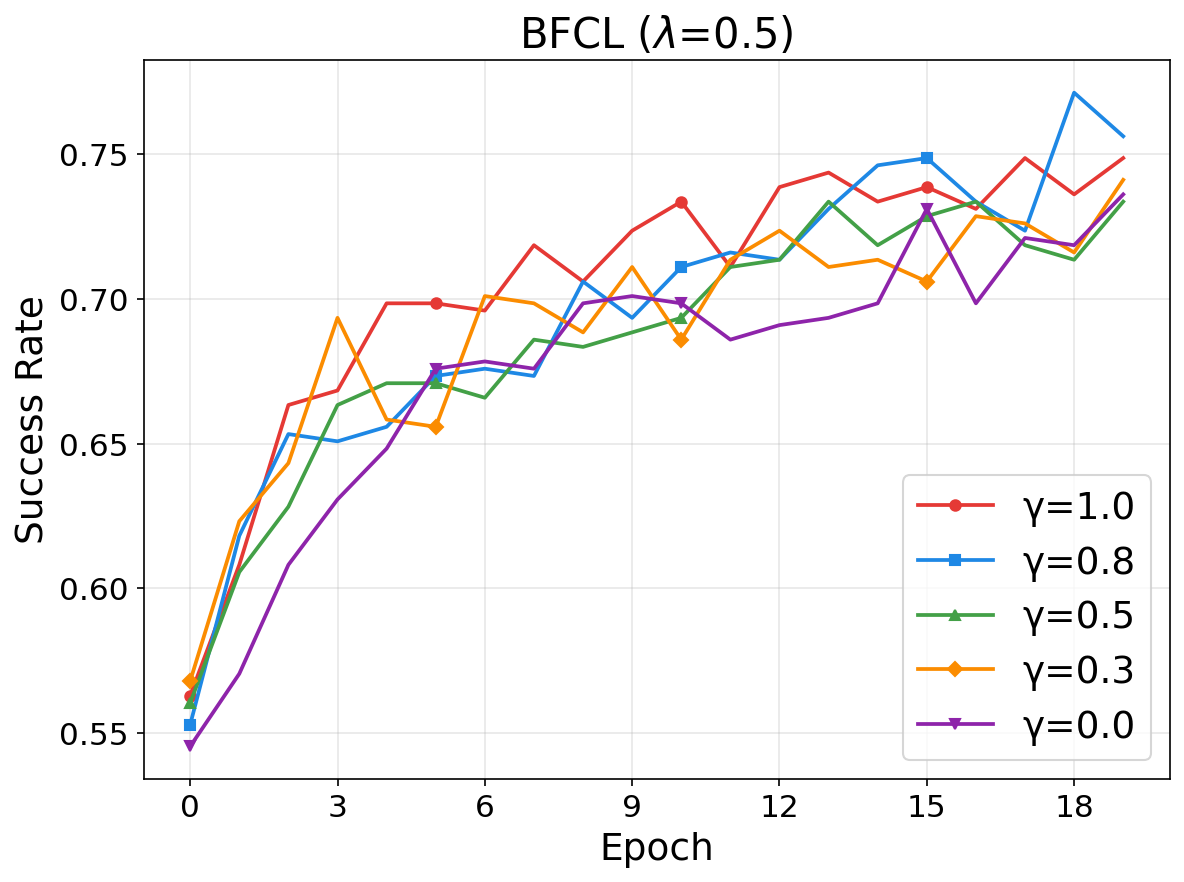}
\hfill
\includegraphics[width=0.32\linewidth]{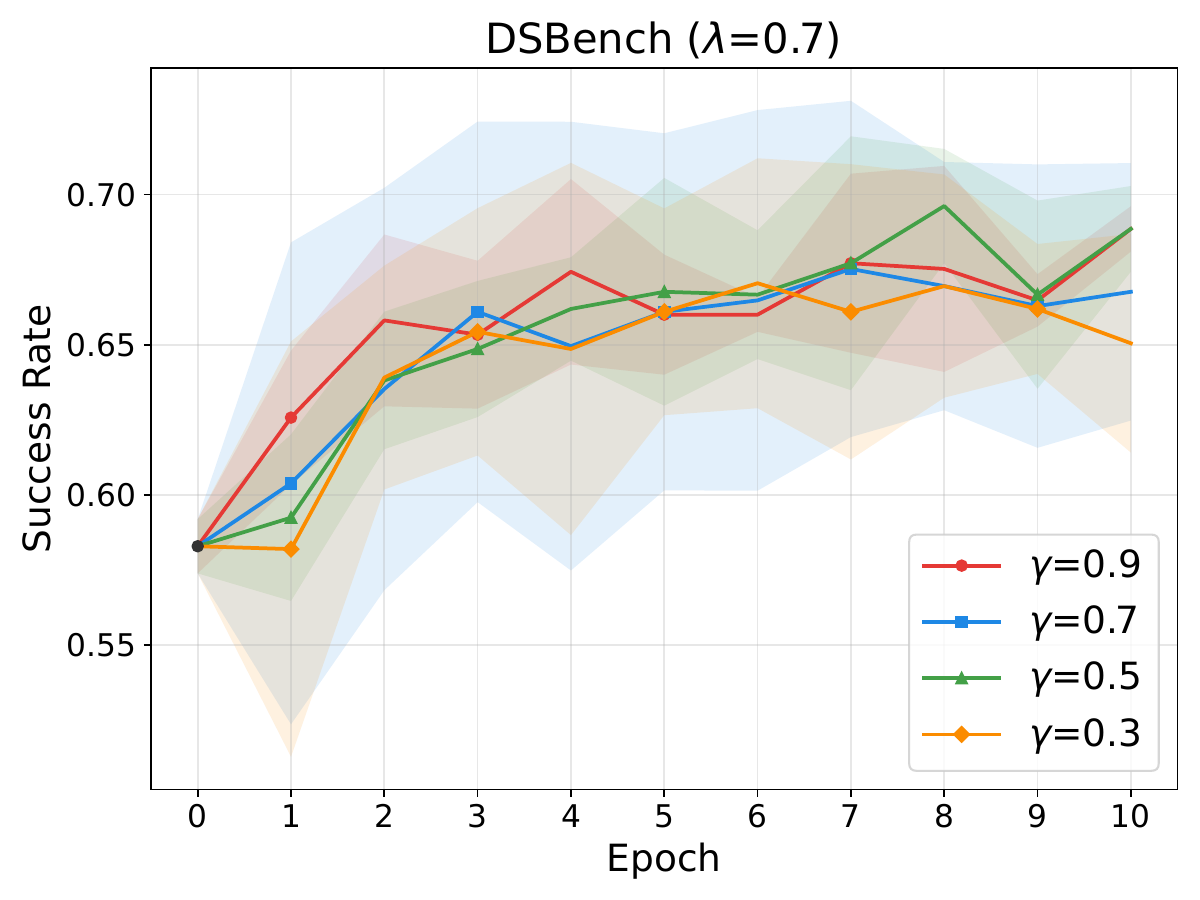}

\vspace{0.8em}
\includegraphics[width=0.32\linewidth]{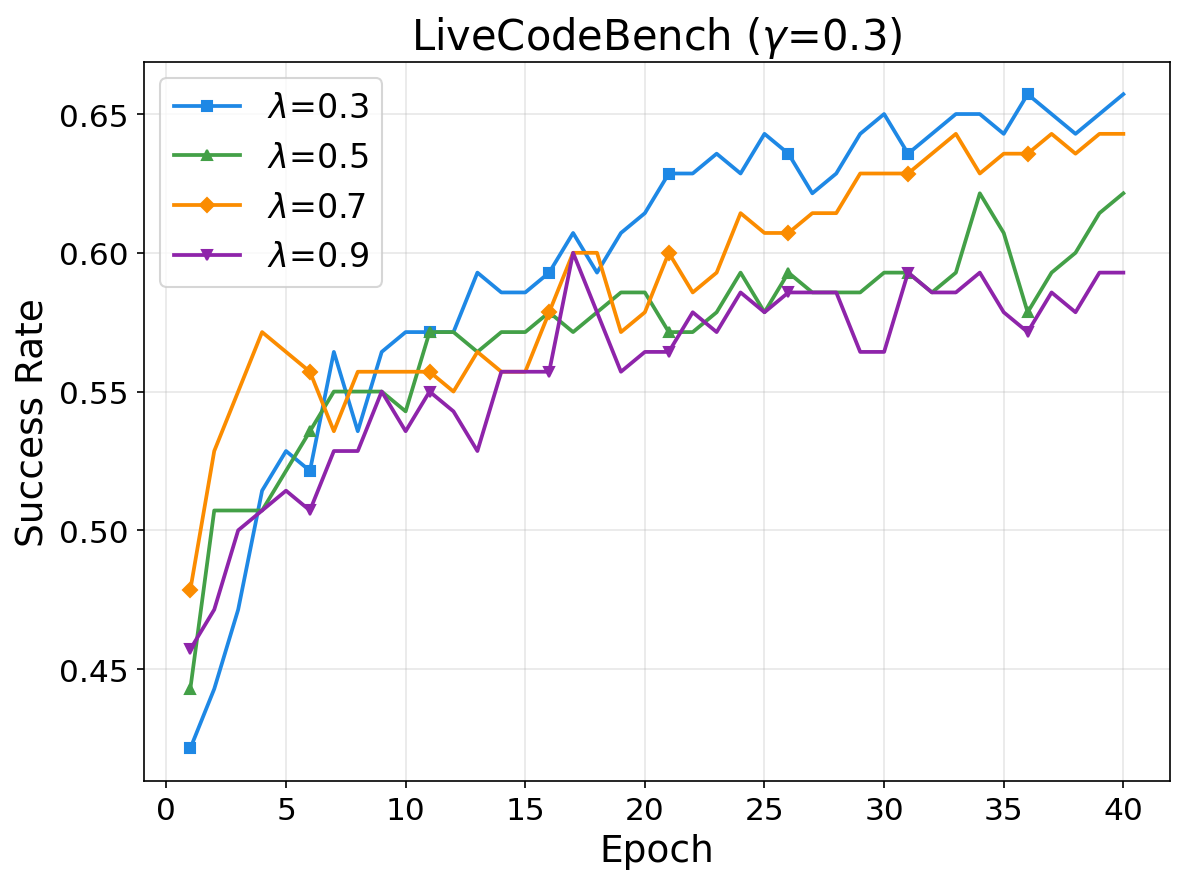}
\hfill
\includegraphics[width=0.32\linewidth]{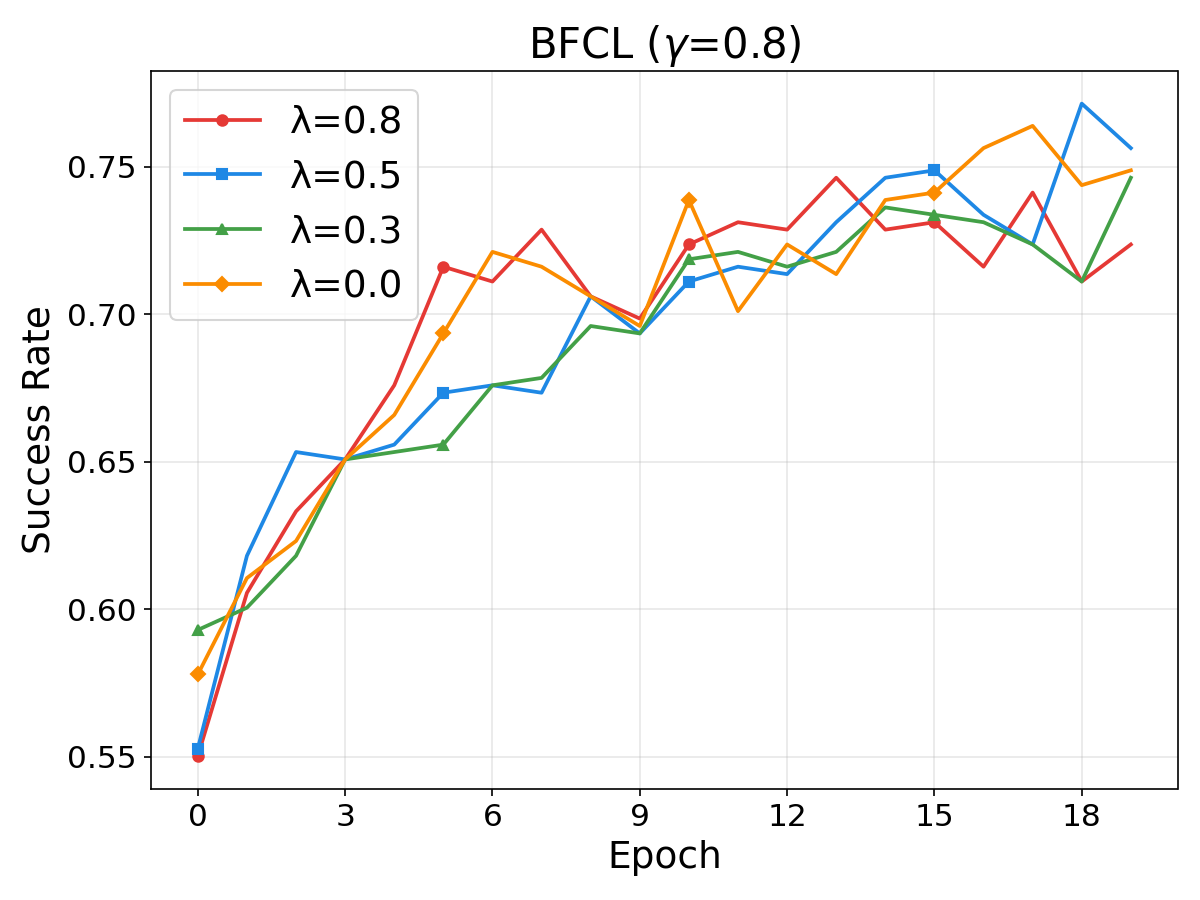}
\hfill
\includegraphics[width=0.32\linewidth]{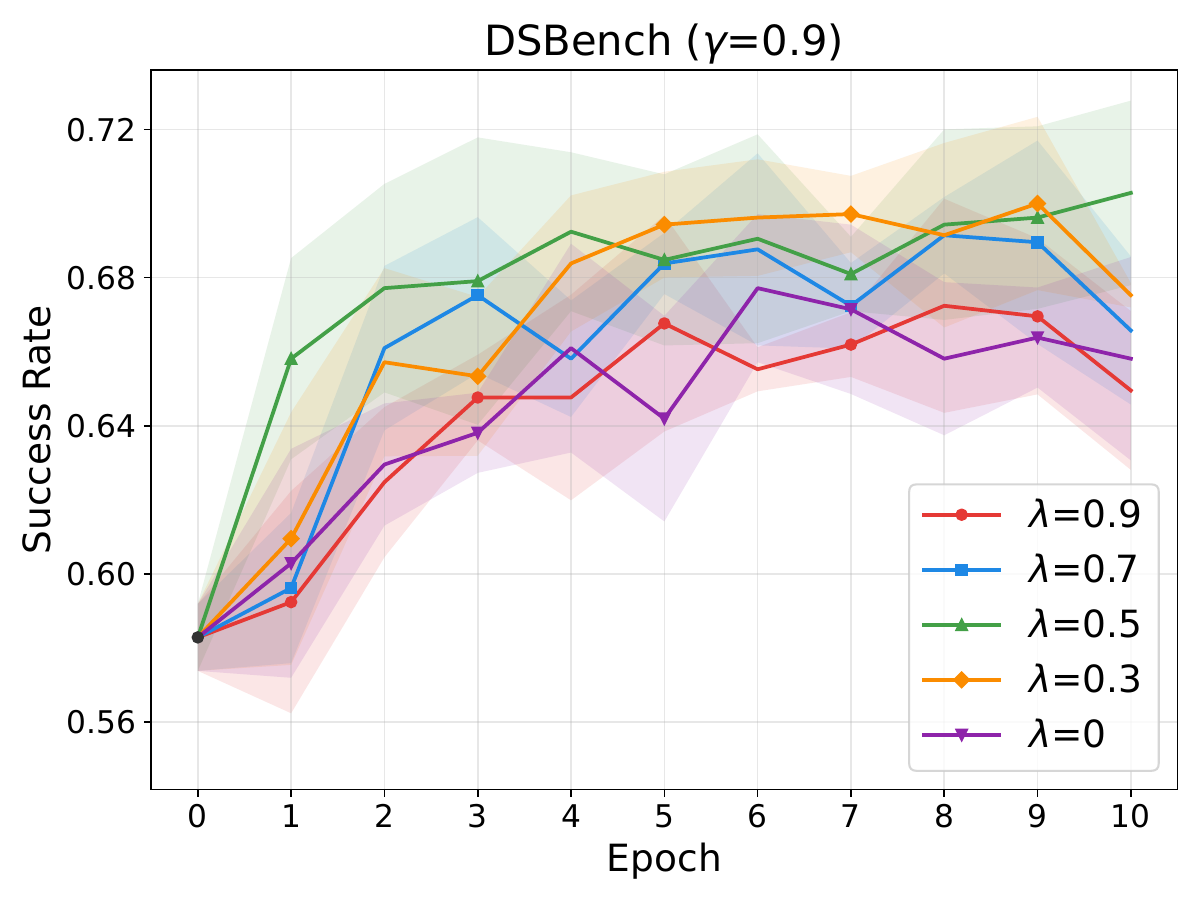}
\caption{{Ablations on $\gamma$ (top) and $\lambda$ (bottom): Success Rate on LiveCodeBench, BFCL, and DSBench.}}
\label{fig:gamma_lambda_sweeps}
\label{fig:gamma_sweep}
\label{fig:lambda_sweep_main}
\end{figure}

\paragraph{{The benefit of a larger $\gamma$ depends on the task.}}
{The discount $\gamma$ shapes both the bootstrap $\gamma Q(m_{\mathrm{new}})$ and structural weight $(\gamma\lambda)^d$ (Eqs.~\ref{eq:td_error}--\ref{eq:general_update}). Figure~\ref{fig:gamma_lambda_sweeps} (top) shows a reversal: LiveCodeBench peaks near $\gamma=0.5$ and declines sharply at $0.9$ ($\sim$63\% vs.\ $\sim$56\%); BFCL favors $\gamma\in[0.8,1.0]$ ($\sim$76\% vs.\ $\sim$73\% at $\gamma=0$). These contrasting responses suggest that the benefit of longer-range credit propagation depends on the task. On DSBench with $\lambda=0.7$, $\gamma=0.5$ and $0.9$ both reach 68.86\% final mean Train SR, versus 65.05\% at $0.3$: moderate and high discounts are effective, with no further final-SR gain from $0.5$ to $0.9$.}

\paragraph{{Moderate trace decay balances credit propagation and attenuation.}}
{Trace decay $\lambda$ governs backward credit, not bootstrapping, weighting depth-$d$ ancestors by $(\gamma\lambda)^d$ (Eq.~\ref{eq:general_update}). On LiveCodeBench at $\gamma=0.3$ (Figure~\ref{fig:gamma_lambda_sweeps}, bottom), $\lambda=0.3$ gives the highest SR ($\sim$65.8\%) and $0.9$ the lowest ($\sim$59.5\%). BFCL (Figures~\ref{fig:gamma_lambda_sweeps}, bottom, and~\ref{fig:lambda_sweep_bfcl}) favors $\lambda=0.5$ over both $0.0$ and $0.8$. DSBench likewise favors intermediate decay: at $\gamma=0.9$, $\lambda=0.5$ reaches 70.29\% final mean Train SR, versus 65.81\% at $0$ and 64.95\% at $0.9$. Memory evolution carries experience across exogenous tasks. The sweep results favor intermediate trace decay, suggesting that credit beyond direct parents is useful when longer-path contributions are attenuated.}

\newpage
\subsection{\texorpdfstring{{RQ3: How does MemQ use memory during runtime learning?}}{RQ3: How does MemQ use memory during runtime learning?}}
\label{sec:memory_values}
\label{sec:practical_analysis}
\label{sec:computational_cost}

{We examine how learned Q-values guide memory retrieval, how task coverage evolves during runtime learning, and how many LLM tokens this process consumes.}

\begin{figure}[t]
\centering
\begin{minipage}[t]{0.49\linewidth}
  \vspace{0pt}
  \begin{subfigure}[t]{\linewidth}
    \centering
    \includegraphics[width=\linewidth]{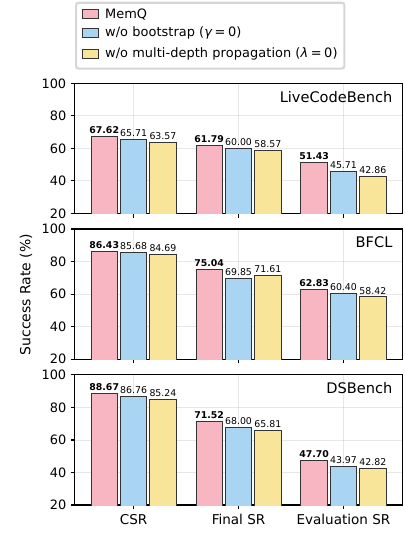}
    \caption{Bootstrapping and multi-depth propagation.}
    \label{fig:credit_assignment_ablation}
  \end{subfigure}
\end{minipage}
\hfill
\begin{minipage}[t]{0.49\linewidth}
  \vspace{0pt}
\begin{subfigure}[t]{\linewidth}
  \centering
  \includegraphics[width=\linewidth]{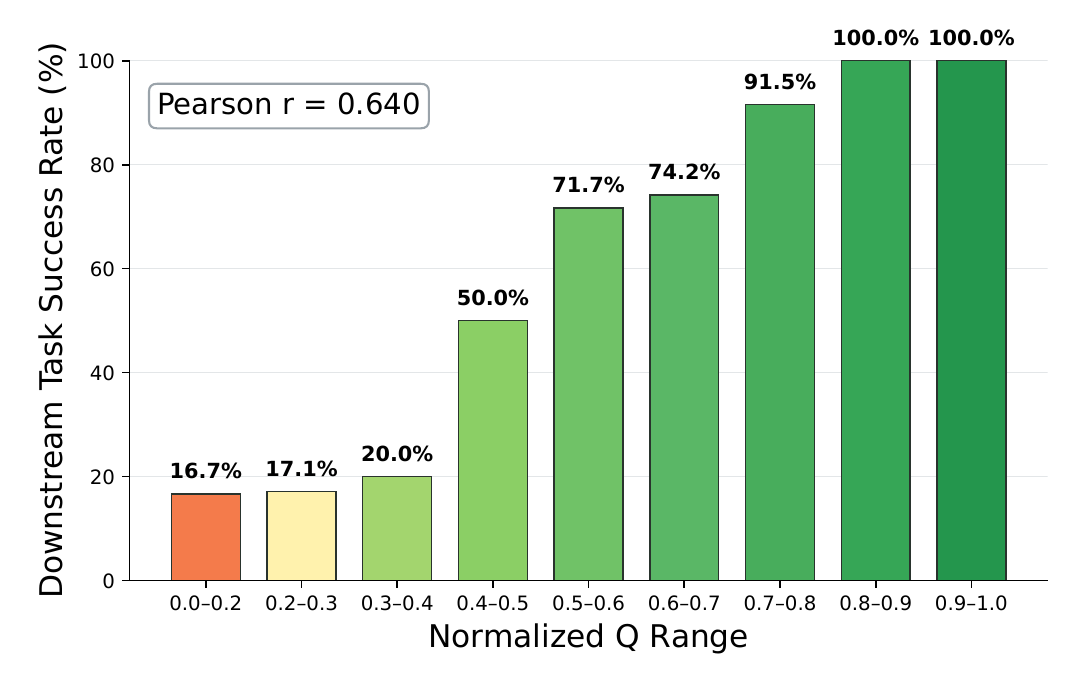}
  \caption{{Q-value and downstream success.}}
  \label{fig:automationbench_q_predictive_a}
\end{subfigure}
\par\medskip
\begin{subfigure}[t]{\linewidth}
  \centering
  \includegraphics[width=\linewidth]{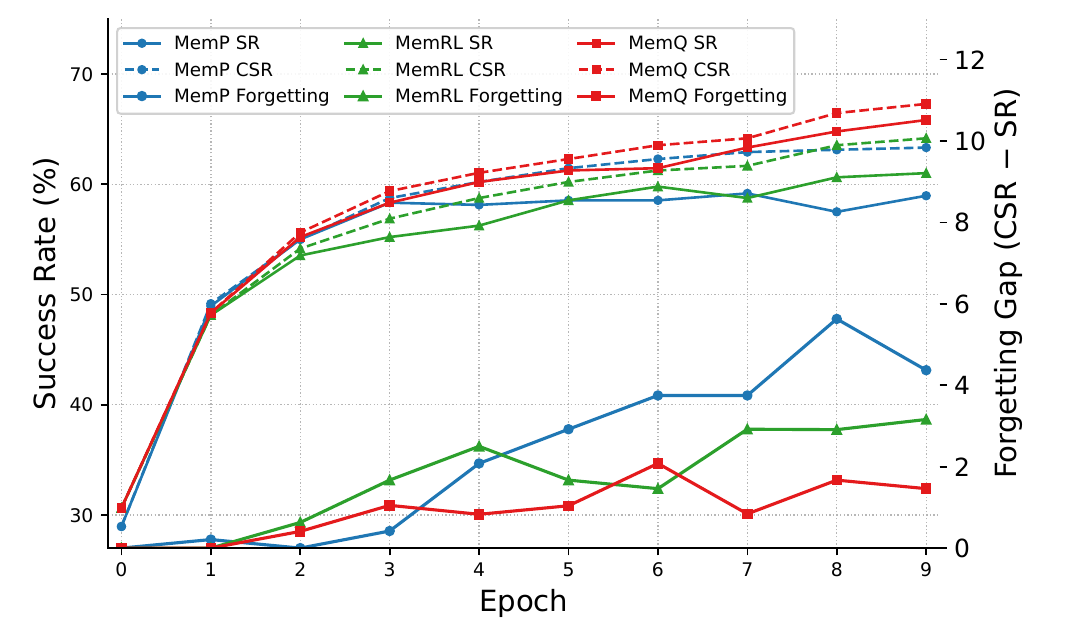}
  \caption{{Runtime learning dynamics.}}
  \label{fig:automationbench_stability}
\end{subfigure}
\end{minipage}
\caption{{Credit-assignment ablations and memory analysis. (a) Removing bootstrapping or multi-depth propagation. (b) Downstream success by the top-ranked memory's normalized retrieval-time Q-value on AutomationBench (480 rollouts). (c) Runtime SR and CSR on AutomationBench.}}
\label{fig:automationbench_value_dynamics}
\end{figure}

\paragraph{{Learned Q-values are associated with subsequent task success.}}
{On AutomationBench, the top-ranked memory's normalized retrieval-time Q-value correlates with exact task success (Pearson $r=0.640$, 480 rollouts). Success rises from 16.7\% in the lowest Q range to 91.5\% at 0.7--0.8 (Figure~\ref{fig:automationbench_q_predictive_a}); values recorded before execution thus distinguish contexts with different observed success rates. High-Q ranges contain a higher proportion of memories from successful trajectories but retain some failed trajectories; Appendix~\ref{app:automationbench_analysis} gives the checkpoint distributions for MemQ and MemRL.}

\paragraph{{Q-values complement semantic similarity in retrieval.}}
{The retrieval-weight ablations examine how this value signal contributes to selecting memories within the locality-filtered candidate set. On LiveCodeBench, evaluation SR is 42.86\% with similarity-only ranking and 45.71\% with Q-only ranking, compared with 51.43\% for the reported mixed ranking. On BFCL, the two endpoints reach 56.77\% and 59.41\%, and several mixtures exceed both. The best BFCL weight differs between evaluation and runtime metrics (Tables~\ref{tab:retrieval_weights_lcb} and~\ref{tab:retrieval_weights_bfcl}). Learned value and semantic relevance therefore provide complementary ranking signals.}

\paragraph{{MemQ maintains more of its historical task coverage.}}
{We next examine whether accumulated task coverage is maintained as learning proceeds. The gap between cumulative and final SR measures the fraction of all tasks solved at least once but not in the final epoch. This gap captures the loss of historical coverage, including stochastic failures.}

\begingroup
\begin{table}[t]

\centering
\begin{minipage}[t]{0.49\linewidth}
\vspace{0pt}
\centering
\captionsetup{font=footnotesize,skip=5pt}
\caption{{Historical coverage and final-epoch success (\%). Values are taken from Table~\ref{tab:runtime_results}; the gap is CSR minus final SR, in percentage points.}}
\label{tab:bfcl_dynamics}
\footnotesize
\setlength{\tabcolsep}{2.5pt}
\begin{tabular}{@{}llccc@{}}
\toprule
Benchmark & Method & CSR & Final SR & Gap $\downarrow$ \\
\midrule
\multirow{3}{*}{BFCL}
 & MemP & 80.15 & 65.91 & \underline{14.24} \\
 & MemRL & \underline{85.85} & \underline{71.27} & 14.58 \\
 & MemQ & \textbf{86.43} & \textbf{75.04} & \textbf{11.39} \\
\midrule
\multirow{3}{*}{AutomationBench}
 & MemP & 63.33 & 58.96 & 4.37 \\
 & MemRL & \underline{64.17} & \underline{61.01} & \underline{3.16} \\
 & MemQ & \textbf{67.29} & \textbf{65.83} & \textbf{1.46} \\
\bottomrule
\end{tabular}
\end{minipage}
\hfill
\begin{minipage}[t]{0.49\linewidth}
\vspace{0pt}
\centering
\captionsetup{font=footnotesize,skip=5pt}
\caption{{DSBench LLM token consumption over 10 training epochs, excluding evaluation and embedding tokens (mean $\pm$ sample SD across three seeds). All values are in millions of tokens. Counts include agent/judge and memory LLM calls; Input and Output denote prompt and completion tokens.}}
\label{tab:token_consumption}
\footnotesize
\setlength{\tabcolsep}{1.75pt}
\begin{tabular}{@{}lccc@{}}
\toprule
Method & \makecell[c]{Input\\(w. memory) $\downarrow$} & Output $\downarrow$ & Total $\downarrow$ \\
\midrule
MemP & $340.01 \pm 25.82$ & $33.34 \pm 1.31$ & $373.35 \pm 25.78$ \\
MemRL & $317.11 \pm 28.37$ & $35.27 \pm 1.11$ & $352.38 \pm 27.65$ \\
MemQ & $\boldsymbol{280.68 \pm 37.85}$ & $\boldsymbol{32.35 \pm 2.73}$ & $\boldsymbol{313.03 \pm 40.01}$ \\
\bottomrule
\end{tabular}
\end{minipage}
\end{table}
\endgroup

{On BFCL, MemQ combines the highest cumulative and final SR with a gap of 11.39~pp, compared with 14.58~pp for MemRL (Table~\ref{tab:bfcl_dynamics}). Its net SR gain from epoch 0 to the final epoch is also larger: 19.38 versus 15.99~pp. The larger gain and smaller gap show that MemQ expands task coverage while maintaining more of that coverage at the end of learning.}

{The AutomationBench trajectories show how this pattern develops during learning (Figure~\ref{fig:automationbench_stability}). MemP's epoch SR plateaus earlier, while MemRL and MemQ continue to improve. At the final epoch, MemQ reaches 65.83\% SR and 67.29\% CSR, with a gap of 1.46~pp; the gaps for MemRL and MemP are 3.16 and 4.37~pp. Across both benchmarks, broader cumulative coverage is accompanied by final performance closer to that coverage. Runtime-evolving SR and cumulative curves for all eight benchmarks are included in Appendix~\ref{app:csr_curves}.}

\paragraph{{MemQ uses memory more effectively and efficiently.}}
{We compare the LLM cost of runtime learning on DSBench for MemP, MemRL, and MemQ, which all use memory proceduralization. Over 10 training epochs, MemQ's mean total token consumption is 16.2\% lower than MemP's and 11.2\% lower than MemRL's (Table~\ref{tab:token_consumption}). Most of the reduction is in input tokens, while MemQ achieves higher DSBench success rates (Tables~\ref{tab:eval_results} and~\ref{tab:runtime_results}). A possible explanation is that more compact, informative memories provide more direct guidance for subsequent decisions, reducing contextual overhead.}

\subsection{\texorpdfstring{{RQ4: When are MemQ's gains limited?}}{RQ4: When are MemQ's gains limited?}}
\label{sec:multiple_choice}

{The preceding experiments demonstrate MemQ's effectiveness across challenging agentic tasks. To explore where its gains are more limited, we additionally evaluate MemQ on three multiple-choice benchmarks: MMMU Pro, ERQA, and GPQA Diamond. The results reveal smaller gains on held-out questions: MemQ has the highest mean on each benchmark, but its margins over the strongest baseline for each task are only 0.19, 0.83, and 0.84~pp, respectively (Table~\ref{tab:multiple_choice_results}).}

\begin{table}[t]
\centering
\caption{{Multiple-choice reasoning results with Gemma-4-E4B-it (\%). For each benchmark, the three rows report Eval SR, Final SR, and CSR, respectively. Entries are mean $\pm$ sample SD. Average across the three benchmarks is reported with approximate 95\% CIs below. Best in \textbf{bold}; second-best \underline{underlined}.}}
\label{tab:multiple_choice_results}
\begingroup
\setlength{\tabcolsep}{5pt}
\resizebox{\linewidth}{!}{%
\begin{tabular}{l | ccccccc}
\toprule
\textbf{Benchmark} & \textbf{No Mem.} & \textbf{RAG} & \textbf{Self-RAG} & \textbf{Mem0} & \textbf{MemP} & \textbf{MemRL} & \emph{\textbf{MemQ}} \\
\midrule
\multirow{3}{*}{\makecell[l]{MMMU Pro\\[-3pt]{\scriptsize Multimodal}}} & 48.46$_{\pm 0.76}$ & 53.76$_{\pm 1.04}$ & 53.57$_{\pm 0.84}$ & 49.90$_{\pm 1.37}$ & \underline{54.24$_{\pm 1.42}$} & 53.66$_{\pm 1.01}$ & \textbf{54.43$_{\pm 0.88}$} \\
 & 48.92$_{\pm 0.50}$ & 49.98$_{\pm 0.79}$ & 49.33$_{\pm 0.62}$ & 49.18$_{\pm 0.77}$ & 51.37$_{\pm 0.72}$ & \underline{57.68$_{\pm 0.66}$} & \textbf{57.83$_{\pm 0.32}$} \\
 & -- & 51.81$_{\pm 0.68}$ & 62.04$_{\pm 0.36}$ & 53.40$_{\pm 0.72}$ & 53.47$_{\pm 1.04}$ & \underline{62.93$_{\pm 0.61}$} & \textbf{63.39$_{\pm 0.19}$} \\
\midrule
\multirow{3}{*}{\makecell[l]{ERQA\\[-3pt]{\scriptsize Embodied Reasoning}}} & 41.67$_{\pm 1.24}$ & 44.58$_{\pm 1.44}$ & \underline{50.42$_{\pm 2.89}$} & 48.33$_{\pm 1.91}$ & 45.00$_{\pm 3.31}$ & 46.67$_{\pm 1.91}$ & \textbf{51.25$_{\pm 2.17}$} \\
 & 39.38$_{\pm 0.00}$ & 39.61$_{\pm 1.66}$ & 40.31$_{\pm 1.02}$ & 37.81$_{\pm 1.02}$ & 41.80$_{\pm 0.95}$ & \underline{57.11$_{\pm 2.50}$} & \textbf{61.33$_{\pm 1.76}$} \\
 & -- & 50.78$_{\pm 0.90}$ & 62.42$_{\pm 1.80}$ & 49.17$_{\pm 1.03}$ & 50.31$_{\pm 1.55}$ & \underline{77.08$_{\pm 1.31}$} & \textbf{83.02$_{\pm 1.85}$} \\
\midrule
\multirow{3}{*}{\makecell[l]{GPQA Diamond\\[-3pt]{\scriptsize Science QA}}} & 47.50$_{\pm 3.01}$ & 58.33$_{\pm 3.82}$ & 58.33$_{\pm 3.82}$ & 58.33$_{\pm 1.18}$ & 59.17$_{\pm 1.44}$ & \underline{60.83$_{\pm 5.20}$} & \textbf{61.67$_{\pm 2.89}$} \\
 & 52.53$_{\pm 1.10}$ & 65.19$_{\pm 3.62}$ & 85.02$_{\pm 1.30}$ & 64.86$_{\pm 1.31}$ & 93.46$_{\pm 1.30}$ & \underline{97.05$_{\pm 0.52}$} & \textbf{98.52$_{\pm 1.37}$} \\
 & -- & 68.99$_{\pm 3.23}$ & 95.99$_{\pm 0.30}$ & 71.94$_{\pm 1.66}$ & 95.99$_{\pm 1.49}$ & \underline{97.89$_{\pm 0.30}$} & \textbf{98.95$_{\pm 0.60}$} \\
\midrule
\textbf{Average (Eval SR)} & \makecell[c]{45.88\\[-2pt]{\scriptsize [43.81, 47.95]}} & \makecell[c]{52.22\\[-2pt]{\scriptsize [49.58, 54.86]}} & \makecell[c]{\underline{54.11}\\[-2pt]{\scriptsize \underline{[51.50, 56.72]}}} & \makecell[c]{52.19\\[-2pt]{\scriptsize [50.89, 53.48]}} & \makecell[c]{52.80\\[-2pt]{\scriptsize [50.62, 54.99]}} & \makecell[c]{53.72\\[-2pt]{\scriptsize [50.04, 57.40]}} & \makecell[c]{\textbf{55.78}\\[-2pt]{\scriptsize \textbf{[53.82, 57.75]}}} \\
\textbf{Average (Final SR)} & \makecell[c]{46.94\\[-2pt]{\scriptsize [46.17, 47.72]}} & \makecell[c]{51.59\\[-2pt]{\scriptsize [49.12, 54.07]}} & \makecell[c]{58.22\\[-2pt]{\scriptsize [57.33, 59.11]}} & \makecell[c]{50.62\\[-2pt]{\scriptsize [49.72, 51.52]}} & \makecell[c]{62.21\\[-2pt]{\scriptsize [61.33, 63.09]}} & \makecell[c]{\underline{70.61}\\[-2pt]{\scriptsize \underline{[68.78, 72.45]}}} & \makecell[c]{\textbf{72.56}\\[-2pt]{\scriptsize \textbf{[71.35, 73.77]}}} \\
\textbf{Average (CSR)} & -- & \makecell[c]{57.19\\[-2pt]{\scriptsize [54.84, 59.55]}} & \makecell[c]{73.48\\[-2pt]{\scriptsize [72.11, 74.86]}} & \makecell[c]{58.17\\[-2pt]{\scriptsize [57.08, 59.26]}} & \makecell[c]{66.59\\[-2pt]{\scriptsize [65.44, 67.74]}} & \makecell[c]{\underline{79.30}\\[-2pt]{\scriptsize \underline{[78.41, 80.19]}}} & \makecell[c]{\textbf{81.79}\\[-2pt]{\scriptsize \textbf{[80.43, 83.15]}}} \\
\bottomrule
\end{tabular}%
}
\endgroup
\end{table}

{The relationship between runtime and held-out performance varies across these tasks. On GPQA, MemQ reaches 98.52\% final SR and 98.95\% CSR, compared with MemRL's 97.05\% and 97.89\%; their held-out scores are 61.67\% and 60.83\%. On MMMU Pro, MemQ and MemRL differ by less than one percentage point on all three metrics. ERQA shows larger improvements over MemRL: 4.58~pp in held-out SR, 4.22~pp in final SR, and 5.94~pp in CSR. Its smaller 0.83~pp lead over Self-RAG reflects the stronger held-out baseline on this task.}

{These benchmarks present largely self-contained questions: most task context is provided in the question, and answering it relies mainly on the model's knowledge and reasoning. Without extended multi-turn interaction with an environment, procedural experience from earlier tasks has fewer opportunities to guide subsequent actions. This may explain the smaller additional gains from MemQ's accumulated experience.}

\section{Conclusion}
\label{sec:conclusions}

\rev{In this paper, we studied multi-step credit assignment in episodic memory for LLM agents, where memories retrieved for one task contribute to the construction of memories used in later tasks. We formalized this setting as an Exogenous-Context MDP and developed MemQ, which propagates TD($\lambda$)-inspired credit along provenance paths, replacing temporal distance with structural depth. Our analysis establishes convergence on fixed provenance DAGs and fixed-point preservation for full-buffer updates. Experiments across eight benchmarks show improvements in both held-out performance and runtime learning, with the largest held-out gains over each task's strongest baseline on DSBench and AutomationBench and smaller gains on multiple-choice reasoning. Ablations support the contribution of bootstrapping and multi-depth propagation, while value and learning-dynamics analyses associate learned Q-values with subsequent success and show better retention of historical task coverage. MemQ also uses fewer training tokens than MemP and MemRL on DSBench. These findings demonstrate the value of accounting for memory-creation dependencies when learning to retrieve and reuse experience.}

%%%%%%%%%%%%%%%%%%%%%%%%%%%%%%%%%%%%%%%%%%%%%%%%%%%%%%%%%%%%%%%%
%% Bibliography
%%%%%%%%%%%%%%%%%%%%%%%%%%%%%%%%%%%%%%%%%%%%%%%%%%%%%%%%%%%%%%%%
\bibliography{main}

@inproceedings{zhang2023rememberer,
title={Large Language Models Are Semi-Parametric Reinforcement Learning Agents},
author={Danyang Zhang and Lu Chen and Situo Zhang and Hongshen Xu and Zihan Zhao and Kai Yu},
booktitle={Thirty-seventh Conference on Neural Information Processing Systems},
year={2023},
url={https://openreview.net/forum?id=ZcJa1R6j3v}
}

@misc{wang2025memento2,
      title={Memento 2: Learning by Stateful Reflective Memory}, 
      author={Jun Wang},
      year={2026},
      eprint={2512.22716},
      archivePrefix={arXiv},
      primaryClass={cs.AI},
      url={https://arxiv.org/abs/2512.22716}, 
}

@misc{zhang2026memrl,
      title={MemRL: Self-Evolving Agents via Runtime Reinforcement Learning on Episodic Memory}, 
      author={Shengtao Zhang and Jiaqian Wang and Ruiwen Zhou and Junwei Liao and Yuchen Feng and Zhuo Li and Yujie Zheng and Weinan Zhang and Ying Wen and Zhiyu Li and Feiyu Xiong and Yutao Qi and Bo Tang and Muning Wen},
      year={2026},
      eprint={2601.03192},
      archivePrefix={arXiv},
      primaryClass={cs.CL},
      url={https://arxiv.org/abs/2601.03192}, 
}

@inproceedings{park2023generative,
author = {Park, Joon Sung and O'Brien, Joseph and Cai, Carrie Jun and Morris, Meredith Ringel and Liang, Percy and Bernstein, Michael S.},
title = {Generative Agents: Interactive Simulacra of Human Behavior},
year = {2023},
isbn = {9798400701320},
publisher = {Association for Computing Machinery},
address = {New York, NY, USA},
url = {https://doi.org/10.1145/3586183.3606763},
doi = {10.1145/3586183.3606763},
booktitle = {Proceedings of the 36th Annual ACM Symposium on User Interface Software and Technology},
articleno = {2},
numpages = {22},
location = {San Francisco, CA, USA},
series = {UIST '23}
}

@misc{packer2023memgpt,
      title={MemGPT: Towards LLMs as Operating Systems}, 
      author={Charles Packer and Sarah Wooders and Kevin Lin and Vivian Fang and Shishir G. Patil and Ion Stoica and Joseph E. Gonzalez},
      year={2024},
      eprint={2310.08560},
      archivePrefix={arXiv},
      primaryClass={cs.AI},
      url={https://arxiv.org/abs/2310.08560}, 
}

@article{sumers2023cognitive,
title={Cognitive Architectures for Language Agents},
author={Theodore Sumers and Shunyu Yao and Karthik R Narasimhan and Thomas L. Griffiths},
journal={Transactions on Machine Learning Research},
issn={2835-8856},
year={2024},
url={https://openreview.net/forum?id=1i6ZCvflQJ},
note={Survey Certification, Featured Certification}
}

@inproceedings{shinn2023reflexion,
title={Reflexion: language agents with verbal reinforcement learning},
author={Noah Shinn and Federico Cassano and Ashwin Gopinath and Karthik R Narasimhan and Shunyu Yao},
booktitle={Thirty-seventh Conference on Neural Information Processing Systems},
year={2023},
url={https://openreview.net/forum?id=vAElhFcKW6}
}

@inproceedings{zhao2024expel,
author = {Zhao, Andrew and Huang, Daniel and Xu, Quentin and Lin, Matthieu and Liu, Yong-Jin and Huang, Gao},
title = {ExpeL: LLM agents are experiential learners},
year = {2024},
isbn = {978-1-57735-887-9},
publisher = {AAAI Press},
url = {https://doi.org/10.1609/aaai.v38i17.29936},
doi = {10.1609/aaai.v38i17.29936},
booktitle = {Proceedings of the Thirty-Eighth AAAI Conference on Artificial Intelligence and Thirty-Sixth Conference on Innovative Applications of Artificial Intelligence and Fourteenth Symposium on Educational Advances in Artificial Intelligence},
articleno = {2188},
numpages = {11},
series = {AAAI'24/IAAI'24/EAAI'24}
}

@article{wang2023voyager,
title={Voyager: An Open-Ended Embodied Agent with Large Language Models},
author={Guanzhi Wang and Yuqi Xie and Yunfan Jiang and Ajay Mandlekar and Chaowei Xiao and Yuke Zhu and Linxi Fan and Anima Anandkumar},
journal={Transactions on Machine Learning Research},
issn={2835-8856},
year={2024},
url={https://openreview.net/forum?id=ehfRiF0R3a},
note={}
}

@inproceedings{zhong2023memorybank,
author = {Zhong, Wanjun and Guo, Lianghong and Gao, Qiqi and Ye, He and Wang, Yanlin},
title = {MemoryBank: enhancing large language models with long-term memory},
year = {2024},
isbn = {978-1-57735-887-9},
publisher = {AAAI Press},
url = {https://doi.org/10.1609/aaai.v38i17.29946},
doi = {10.1609/aaai.v38i17.29946},
booktitle = {Proceedings of the Thirty-Eighth AAAI Conference on Artificial Intelligence and Thirty-Sixth Conference on Innovative Applications of Artificial Intelligence and Fourteenth Symposium on Educational Advances in Artificial Intelligence},
articleno = {2198},
numpages = {8},
series = {AAAI'24/IAAI'24/EAAI'24}
}

@misc{kynoch2023recallm,
      title={RecallM: An Adaptable Memory Mechanism with Temporal Understanding for Large Language Models}, 
      author={Brandon Kynoch and Hugo Latapie and Dwane van der Sluis},
      year={2023},
      eprint={2307.02738},
      archivePrefix={arXiv},
      primaryClass={cs.AI},
      url={https://arxiv.org/abs/2307.02738}, 
}

@misc{blundell2016model,
      title={Model-Free Episodic Control}, 
      author={Charles Blundell and Benigno Uria and Alexander Pritzel and Yazhe Li and Avraham Ruderman and Joel Z Leibo and Jack Rae and Daan Wierstra and Demis Hassabis},
      year={2016},
      eprint={1606.04460},
      archivePrefix={arXiv},
      primaryClass={stat.ML},
      url={https://arxiv.org/abs/1606.04460}, 
}

@inproceedings{lin2018episodic,
author = {Lin, Zichuan and Zhao, Tianqi and Yang, Guangwen and Zhang, Lintao},
title = {Episodic memory deep Q-networks},
year = {2018},
isbn = {9780999241127},
publisher = {AAAI Press},
booktitle = {Proceedings of the 27th International Joint Conference on Artificial Intelligence},
pages = {2433–2439},
numpages = {7},
location = {Stockholm, Sweden},
series = {IJCAI'18}
}

@article{borkar2000ode,
author = {Borkar, V. S. and Meyn, S. P.},
title = {The O.D. E. Method for Convergence of Stochastic Approximation and Reinforcement Learning},
year = {2000},
issue_date = {Jan. 2000},
publisher = {Society for Industrial and Applied Mathematics},
address = {USA},
volume = {38},
number = {2},
issn = {0363-0129},
url = {https://doi.org/10.1137/S0363012997331639},
doi = {10.1137/S0363012997331639},
journal = {SIAM J. Control Optim.},
month = jan,
pages = {447–469},
numpages = {23}
}

@article{sutton1988learning,
  title={Learning to Predict by the Methods of Temporal Differences},
  author={Sutton, Richard S},
  journal={Machine Learning},
  volume={3},
  number={1},
  pages={9--44},
  year={1988}
}

@article{singh1996reinforcement,
author = {Singh, Satinder P. and Sutton, Richard S.},
title = {Reinforcement learning with replacing eligibility traces},
year = {1996},
issue_date = {Jan./Feb./March 1996},
publisher = {Kluwer Academic Publishers},
address = {USA},
volume = {22},
number = {1–3},
issn = {0885-6125},
url = {https://doi.org/10.1007/BF00114726},
doi = {10.1007/BF00114726},
journal = {Mach. Learn.},
month = jan,
pages = {123–158},
numpages = {36}
}

@book{sutton2018reinforcement,
  title={Reinforcement Learning: An Introduction},
  author={Sutton, Richard S and Barto, Andrew G},
  edition={2nd},
  publisher={MIT Press},
  year={2018}
}

@InProceedings{vanseijen2014true,
  title = 	 {True Online TD(lambda)},
  author = 	 {Seijen, Harm and Sutton, Rich},
  booktitle = 	 {Proceedings of the 31st International Conference on Machine Learning},
  pages = 	 {692--700},
  year = 	 {2014},
  editor = 	 {Xing, Eric P. and Jebara, Tony},
  volume = 	 {32},
  number =       {1},
  series = 	 {Proceedings of Machine Learning Research},
  address = 	 {Bejing, China},
  month = 	 {22--24 Jun},
  publisher =    {PMLR},
  url = 	 {https://proceedings.mlr.press/v32/seijen14.html}
}

@phdthesis{watkins1989learning,
  title={Learning from Delayed Rewards},
  author={Watkins, Christopher John Cornish Hellaby},
  school={King's College, Cambridge},
  year={1989}
}

@article{peng1996incremental,
  title={Incremental Multi-Step {Q}-Learning},
  author={Peng, Jing and Williams, Ronald J},
  journal={Machine Learning},
  volume={22},
  number={1--3},
  pages={283--290},
  year={1996}
}

@inproceedings{munos2016safe,
  title={Safe and Efficient Off-Policy Reinforcement Learning},
  author={Munos, R{\'e}mi and Stepleton, Tom and Harutyunyan, Anna and Bellemare, Marc G},
  booktitle={Advances in Neural Information Processing Systems},
  volume={29},
  year={2016}
}

@inproceedings{schulman2016gae,
  title={High-Dimensional Continuous Control Using Generalized Advantage Estimation},
  author={Schulman, John and Moritz, Philipp and Levine, Sergey and Jordan, Michael I and Abbeel, Pieter},
  booktitle={International Conference on Learning Representations},
  year={2016}
}

@inproceedings{espeholt2018impala,
  title={{IMPALA}: Scalable Distributed Deep-{RL} with Importance Weighted Actor-Learner Architectures},
  author={Espeholt, Lasse and Soyer, Hubert and Munos, R{\'e}mi and Simonyan, Karen and Mnih, Volodymyr and Ward, Tom and Doron, Yotam and Firoiu, Vlad and Harley, Tim and Dunning, Iain and Legg, Shane and Kavukcuoglu, Koray},
  booktitle={International Conference on Machine Learning},
  year={2018}
}

@misc{graves2014neural,
      title={Neural Turing Machines}, 
      author={Alex Graves and Greg Wayne and Ivo Danihelka},
      year={2014},
      eprint={1410.5401},
      archivePrefix={arXiv},
      primaryClass={cs.NE},
      url={https://arxiv.org/abs/1410.5401}, 
}

@InProceedings{pritzel2017neural,
  title = 	 {Neural Episodic Control},
  author =       {Alexander Pritzel and Benigno Uria and Sriram Srinivasan and Adri{\`a} Puigdom{\`e}nech Badia and Oriol Vinyals and Demis Hassabis and Daan Wierstra and Charles Blundell},
  booktitle = 	 {Proceedings of the 34th International Conference on Machine Learning},
  pages = 	 {2827--2836},
  year = 	 {2017},
  editor = 	 {Precup, Doina and Teh, Yee Whye},
  volume = 	 {70},
  series = 	 {Proceedings of Machine Learning Research},
  month = 	 {06--11 Aug},
  publisher =    {PMLR},
  url = 	 {https://proceedings.mlr.press/v70/pritzel17a.html}
}

@inproceedings{schaul2016prioritized,
  title={Prioritized Experience Replay},
  author={Schaul, Tom and Quan, John and Antonoglou, Ioannis and Silver, David},
  booktitle={International Conference on Learning Representations},
  year={2016}
}

@inproceedings{guu2020realm,
author = {Guu, Kelvin and Lee, Kenton and Tung, Zora and Pasupat, Panupong and Chang, Ming-Wei},
title = {REALM: retrieval-augmented language model pre-training},
year = {2020},
publisher = {JMLR.org},
booktitle = {Proceedings of the 37th International Conference on Machine Learning},
articleno = {368},
numpages = {10},
series = {ICML'20}
}

@misc{yan2025memoryr1,
  title={{Memory-R1}: Enhancing Large Language Model Agents to Manage and Utilize Memories via Reinforcement Learning},
  author={Yan, Sikuan and Yang, Xiufeng and Huang, Zuchao and Nie, Ercong and Ding, Zifeng and Li, Zonggen and Ma, Xiaowen and Bi, Jinhe and Kersting, Kristian and Pan, Jeff Z. and Sch{\"u}tze, Hinrich and Tresp, Volker and Ma, Yunpu},
  year={2026},
  eprint={2508.19828},
  archivePrefix={arXiv},
  primaryClass={cs.CL},
  url={https://arxiv.org/abs/2508.19828}
}

@misc{zhang2025memact,
      title={Memory as Action: Autonomous Context Curation for Long-Horizon Agentic Tasks}, 
      author={Yuxiang Zhang and Jiangming Shu and Ye Ma and Xueyuan Lin and Shangxi Wu and Jitao Sang},
      year={2026},
      eprint={2510.12635},
      archivePrefix={arXiv},
      primaryClass={cs.AI},
      url={https://arxiv.org/abs/2510.12635}, 
}

@misc{ma2026finemem,
      title={Fine-Mem: Fine-Grained Feedback Alignment for Long-Horizon Memory Management}, 
      author={Weitao Ma and Xiaocheng Feng and Lei Huang and Xiachong Feng and Zhanyu Ma and Jun Xu and Jiuchong Gao and Jinghua Hao and Renqing He and Bing Qin},
      year={2026},
      eprint={2601.08435},
      archivePrefix={arXiv},
      primaryClass={cs.CL},
      url={https://arxiv.org/abs/2601.08435}, 
}

@misc{shen2026membuilder,
      title={MemBuilder: Reinforcing LLMs for Long-Term Memory Construction via Attributed Dense Rewards}, 
      author={Zhiyu Shen and Ziming Wu and Fuming Lai and Shaobing Lian and Yanghui Rao},
      year={2026},
      eprint={2601.05488},
      archivePrefix={arXiv},
      primaryClass={cs.CL},
      url={https://arxiv.org/abs/2601.05488}, 
}

@misc{zhang2026retroagent,
      title={RetroAgent: From Solving to Evolving via Retrospective Dual Intrinsic Feedback}, 
      author={Xiaoying Zhang and Zichen Liu and Yipeng Zhang and Xia Hu and Wenqi Shao},
      year={2026},
      eprint={2603.08561},
      archivePrefix={arXiv},
      primaryClass={cs.AI},
      url={https://arxiv.org/abs/2603.08561}, 
}

@inproceedings{rein2024gpqa,
title={{GPQA}: A Graduate-Level Google-Proof Q\&A Benchmark},
author={David Rein and Betty Li Hou and Asa Cooper Stickland and Jackson Petty and Richard Yuanzhe Pang and Julien Dirani and Julian Michael and Samuel R. Bowman},
booktitle={First Conference on Language Modeling},
year={2024},
url={https://openreview.net/forum?id=Ti67584b98}
}

@misc{qwen3.5,
    title  = {{Qwen3.5}: Towards Native Multimodal Agents},
    author = {{Qwen Team}},
    month  = {February},
    year   = {2026},
    url    = {https://qwen.ai/blog?id=qwen3.5}
}

@misc{gemma_4_e4b_it,
  author       = {Gemma Team, Google},
  title        = {Gemma 4 E4B-it},
  year         = {2026},
  publisher    = {Hugging Face},
  howpublished = {\url{https://huggingface.co/google/gemma-4-E4B-it}},
  note         = {Vertex AI / Hugging Face Model Hub}
}

@misc{geminiroboticsteam2025geminiroboticsbringingai,
      title={Gemini Robotics: Bringing AI into the Physical World}, 
      author={Gemini Robotics Team and Saminda Abeyruwan and Joshua Ainslie and Jean-Baptiste Alayrac and Montserrat Gonzalez Arenas and Travis Armstrong and Ashwin Balakrishna and Robert Baruch and Maria Bauza and Michiel Blokzijl and Steven Bohez and Konstantinos Bousmalis and Anthony Brohan and Thomas Buschmann and Arunkumar Byravan and Serkan Cabi and Ken Caluwaerts and Federico Casarini and Oscar Chang and Jose Enrique Chen and Xi Chen and Hao-Tien Lewis Chiang and Krzysztof Choromanski and David D'Ambrosio and Sudeep Dasari and Todor Davchev and Coline Devin and Norman Di Palo and Tianli Ding and Adil Dostmohamed and Danny Driess and Yilun Du and Debidatta Dwibedi and Michael Elabd and Claudio Fantacci and Cody Fong and Erik Frey and Chuyuan Fu and Marissa Giustina and Keerthana Gopalakrishnan and Laura Graesser and Leonard Hasenclever and Nicolas Heess and Brandon Hernaez and Alexander Herzog and R. Alex Hofer and Jan Humplik and Atil Iscen and Mithun George Jacob and Deepali Jain and Ryan Julian and Dmitry Kalashnikov and M. Emre Karagozler and Stefani Karp and Chase Kew and Jerad Kirkland and Sean Kirmani and Yuheng Kuang and Thomas Lampe and Antoine Laurens and Isabel Leal and Alex X. Lee and Tsang-Wei Edward Lee and Jacky Liang and Yixin Lin and Sharath Maddineni and Anirudha Majumdar and Assaf Hurwitz Michaely and Robert Moreno and Michael Neunert and Francesco Nori and Carolina Parada and Emilio Parisotto and Peter Pastor and Acorn Pooley and Kanishka Rao and Krista Reymann and Dorsa Sadigh and Stefano Saliceti and Pannag Sanketi and Pierre Sermanet and Dhruv Shah and Mohit Sharma and Kathryn Shea and Charles Shu and Vikas Sindhwani and Sumeet Singh and Radu Soricut and Jost Tobias Springenberg and Rachel Sterneck and Razvan Surdulescu and Jie Tan and Jonathan Tompson and Vincent Vanhoucke and Jake Varley and Grace Vesom and Giulia Vezzani and Oriol Vinyals and Ayzaan Wahid and Stefan Welker and Paul Wohlhart and Fei Xia and Ted Xiao and Annie Xie and Jinyu Xie and Peng Xu and Sichun Xu and Ying Xu and Zhuo Xu and Yuxiang Yang and Rui Yao and Sergey Yaroshenko and Wenhao Yu and Wentao Yuan and Jingwei Zhang and Tingnan Zhang and Allan Zhou and Yuxiang Zhou},
      year={2025},
      eprint={2503.20020},
      archivePrefix={arXiv},
      primaryClass={cs.RO},
      url={https://arxiv.org/abs/2503.20020}, 
}

@inproceedings{asai2024selfrag,
title={Self-{RAG}: Learning to Retrieve, Generate, and Critique through Self-Reflection},
author={Akari Asai and Zeqiu Wu and Yizhong Wang and Avirup Sil and Hannaneh Hajishirzi},
booktitle={The Twelfth International Conference on Learning Representations},
year={2024},
url={https://openreview.net/forum?id=hSyW5go0v8}
}

@misc{fang2026mempexploringagentprocedural,
      title={Memp: Exploring Agent Procedural Memory}, 
      author={Runnan Fang and Yuan Liang and Xiaobin Wang and Jialong Wu and Shuofei Qiao and Pengjun Xie and Fei Huang and Huajun Chen and Ningyu Zhang},
      year={2026},
      eprint={2508.06433},
      archivePrefix={arXiv},
      primaryClass={cs.CL},
      url={https://arxiv.org/abs/2508.06433}, 
}

@misc{mem0,
      title={Mem0: Building Production-Ready AI Agents with Scalable Long-Term Memory}, 
      author={Prateek Chhikara and Dev Khant and Saket Aryan and Taranjeet Singh and Deshraj Yadav},
      year={2025},
      eprint={2504.19413},
      archivePrefix={arXiv},
      primaryClass={cs.CL},
      url={https://arxiv.org/abs/2504.19413}, 
}

@misc{zhou2025memento,
      title={Memento: Fine-tuning LLM Agents without Fine-tuning LLMs}, 
      author={Huichi Zhou and Yihang Chen and Siyuan Guo and Xue Yan and Kin Hei Lee and Zihan Wang and Ka Yiu Lee and Guchun Zhang and Kun Shao and Linyi Yang and Jun Wang},
      year={2025},
      eprint={2508.16153},
      archivePrefix={arXiv},
      primaryClass={cs.LG},
      url={https://arxiv.org/abs/2508.16153}, 
}

@misc{nan2025nemori,
      title={What Deserves Memory: Adaptive Memory Distillation for LLM Agents}, 
      author={Wenquan Ma and Jiayan Nan and Wenlong Wu and Yize Chen},
      year={2026},
      eprint={2508.03341},
      archivePrefix={arXiv},
      primaryClass={cs.AI},
      url={https://arxiv.org/abs/2508.03341}, 
}

@misc{zhang2025memskill,
      title={MemSkill: Learning and Evolving Memory Skills for Self-Evolving Agents}, 
      author={Haozhen Zhang and Quanyu Long and Jianzhu Bao and Tao Feng and Weizhi Zhang and Haodong Yue and Wenya Wang},
      year={2026},
      eprint={2602.02474},
      archivePrefix={arXiv},
      primaryClass={cs.CL},
      url={https://arxiv.org/abs/2602.02474}, 
}

@misc{zhang2026liveevo,
      title={Live-Evo: Online Evolution of Agentic Memory from Continuous Feedback}, 
      author={Yaolun Zhang and Yiran Wu and Yijiong Yu and Qingyun Wu and Huazheng Wang},
      year={2026},
      eprint={2602.02369},
      archivePrefix={arXiv},
      primaryClass={cs.AI},
      url={https://arxiv.org/abs/2602.02369}, 
}

@misc{zhu2026helamem,
      title={HeLa-Mem: Hebbian Learning and Associative Memory for LLM Agents}, 
      author={Jinchang Zhu and Jindong Li and Cheng Zhang and Jiahong Liu and Menglin Yang},
      year={2026},
      eprint={2604.16839},
      archivePrefix={arXiv},
      primaryClass={cs.CL},
      url={https://arxiv.org/abs/2604.16839}, 
}

@misc{cao2025reme,
      title={Remember Me, Refine Me: A Dynamic Procedural Memory Framework for Experience-Driven Agent Evolution}, 
      author={Zouying Cao and Jiaji Deng and Li Yu and Weikang Zhou and Zhaoyang Liu and Bolin Ding and Hai Zhao},
      year={2026},
      eprint={2512.10696},
      archivePrefix={arXiv},
      primaryClass={cs.AI},
      url={https://arxiv.org/abs/2512.10696}, 
}

@misc{zhou2025mem1,
      title={MEM1: Learning to Synergize Memory and Reasoning for Efficient Long-Horizon Agents}, 
      author={Zijian Zhou and Ao Qu and Zhaoxuan Wu and Sunghwan Kim and Alok Prakash and Daniela Rus and Jinhua Zhao and Bryan Kian Hsiang Low and Paul Pu Liang},
      year={2025},
      eprint={2506.15841},
      archivePrefix={arXiv},
      primaryClass={cs.CL},
      url={https://arxiv.org/abs/2506.15841}, 
}

@misc{yue2026memt,
      title={Mem-T: Densifying Rewards for Long-Horizon Memory Agents}, 
      author={Yanwei Yue and Boci Peng and Xuanbo Fan and Jiaxin Guo and Qiankun Li and Yan Zhang},
      year={2026},
      eprint={2601.23014},
      archivePrefix={arXiv},
      primaryClass={cs.LG},
      url={https://arxiv.org/abs/2601.23014}, 
}

@misc{wei2025evomemory,
      title={Evo-Memory: Benchmarking LLM Agent Test-time Learning with Self-Evolving Memory}, 
      author={Tianxin Wei and Noveen Sachdeva and Benjamin Coleman and Zhankui He and Yuanchen Bei and Xuying Ning and Mengting Ai and Yunzhe Li and Jingrui He and Ed H. Chi and Chi Wang and Shuo Chen and Fernando Pereira and Wang-Cheng Kang and Derek Zhiyuan Cheng},
      year={2025},
      eprint={2511.20857},
      archivePrefix={arXiv},
      primaryClass={cs.CL},
      url={https://arxiv.org/abs/2511.20857}, 
}

@inproceedings{ouyang2025reasoningbank,
title={ReasoningBank: Scaling Agent Self-Evolving with Reasoning Memory},
author={Siru Ouyang and Jun Yan and I-Hung Hsu and Yanfei Chen and Ke Jiang and Zifeng Wang and Rujun Han and Long Le and Samira Daruki and Xiangru Tang and Vishy Tirumalashetty and George Lee and Mahsan Rofouei and Hangfei Lin and Jiawei Han and Chen-Yu Lee and Tomas Pfister},
booktitle={The Fourteenth International Conference on Learning Representations},
year={2026},
url={https://openreview.net/forum?id=jL7fwchScm}
}

@misc{zhang2025memevolve,
      title={MemEvolve: Meta-Evolution of Agent Memory Systems}, 
      author={Guibin Zhang and Haotian Ren and Chong Zhan and Zhenhong Zhou and Junhao Wang and He Zhu and Wangchunshu Zhou and Shuicheng Yan},
      year={2025},
      eprint={2512.18746},
      archivePrefix={arXiv},
      primaryClass={cs.CL},
      url={https://arxiv.org/abs/2512.18746}, 
}

@misc{yuan2025memsearcher,
title={MemSearcher: Training {LLM}s to Reason, Search and Manage Memory via End-to-End Reinforcement Learning},
author={Qianhao Yuan and Jie Lou and Zichao Li and Jiawei Chen and Yaojie Lu and Hongyu Lin and Le Sun and Debing Zhang and Xianpei Han},
year={2026},
url={https://openreview.net/forum?id=EWIAx3NgvA}
}

@misc{zheng2026lifelongagentbench,
      title={LifelongAgentBench: Evaluating LLM Agents as Lifelong Learners}, 
      author={Junhao Zheng and Xidi Cai and Qiuke Li and Duzhen Zhang and ZhongZhi Li and Yingying Zhang and Le Song and Qianli Ma},
      year={2025},
      eprint={2505.11942},
      archivePrefix={arXiv},
      primaryClass={cs.AI},
      url={https://arxiv.org/abs/2505.11942}, 
}

@inproceedings{patil2025bfcl,
title={The Berkeley Function Calling Leaderboard ({BFCL}): From Tool Use to Agentic Evaluation of Large Language Models},
author={Shishir G Patil and Huanzhi Mao and Fanjia Yan and Charlie Cheng-Jie Ji and Vishnu Suresh and Ion Stoica and Joseph E. Gonzalez},
booktitle={Forty-second International Conference on Machine Learning},
year={2025},
url={https://openreview.net/forum?id=2GmDdhBdDk}
}

@misc{jing2024dsbench,
  title={DSBench: How Far Are Data Science Agents to Becoming Data Science Experts?},
  author={Liqiang Jing and Zhehui Huang and Xiaoyang Wang and Wenlin Yao and Wenhao Yu and Kaixin Ma and Hongming Zhang and Xinya Du and Dong Yu},
  year={2024},
  eprint={2409.07703},
  archivePrefix={arXiv},
  url={https://arxiv.org/abs/2409.07703}
}

@misc{automationbench2026,
  title={AutomationBench},
  author={Daniel Shepard and Robin Salimans},
  year={2026},
  eprint={2604.18934},
  archivePrefix={arXiv},
  url={https://arxiv.org/abs/2604.18934}
}

@misc{deepseekv4flash,
  title={{DeepSeek-V4}: Towards Highly Efficient Million-Token Context Intelligence},
  author={{DeepSeek-AI}},
  year={2026},
  eprint={2606.19348},
  archivePrefix={arXiv},
  url={https://arxiv.org/abs/2606.19348}
}

@inproceedings{jain2025livecodebench,
title={LiveCodeBench: Holistic and Contamination Free Evaluation of Large Language Models for Code},
author={Naman Jain and King Han and Alex Gu and Wen-Ding Li and Fanjia Yan and Tianjun Zhang and Sida Wang and Armando Solar-Lezama and Koushik Sen and Ion Stoica},
booktitle={The Thirteenth International Conference on Learning Representations},
year={2025},
url={https://openreview.net/forum?id=chfJJYC3iL}
}

@inproceedings{yue-etal-2025-mmmu,
    title = "{MMMU}-Pro: A More Robust Multi-discipline Multimodal Understanding Benchmark",
    author = "Yue, Xiang  and
      Zheng, Tianyu  and
      Ni, Yuansheng  and
      Wang, Yubo  and
      Zhang, Kai  and
      Tong, Shengbang  and
      Sun, Yuxuan  and
      Yu, Botao  and
      Zhang, Ge  and
      Sun, Huan  and
      Su, Yu  and
      Chen, Wenhu  and
      Neubig, Graham",
    editor = "Che, Wanxiang  and
      Nabende, Joyce  and
      Shutova, Ekaterina  and
      Pilehvar, Mohammad Taher",
    booktitle = "Proceedings of the 63rd Annual Meeting of the Association for Computational Linguistics (Volume 1: Long Papers)",
    month = jul,
    year = "2025",
    address = "Vienna, Austria",
    publisher = "Association for Computational Linguistics",
    url = "https://aclanthology.org/2025.acl-long.736/",
    doi = "10.18653/v1/2025.acl-long.736",
    pages = "15134--15186",
    ISBN = "979-8-89176-251-0"
}

@incollection{robbins1971convergence,
  author = {Robbins, Herbert and Siegmund, David},
  title = {A Convergence Theorem for Non Negative Almost Supermartingales and Some Applications},
  booktitle = {Optimizing Methods in Statistics},
  publisher = {Academic Press},
  pages = {233--257},
  year = {1971},
  doi = {10.1016/B978-0-12-604550-5.50015-8},
  url = {https://www.sciencedirect.com/science/article/pii/B9780126045505500158}
}
\bibliographystyle{rlc}

%%%%%%%%%%%%%%%%%%%%%%%%%%%%%%%%%%%%%%%%%%%%%%%%%%%%%%%%%%%%%%%%
%% Appendices
%%%%%%%%%%%%%%%%%%%%%%%%%%%%%%%%%%%%%%%%%%%%%%%%%%%%%%%%%%%%%%%%
\newpage
\appendix
\begingroup

\section{Theoretical Analysis of MemQ}
\label{app:convergence}

We study the value recursion in Section~\ref{sec:credit_propagation} for a fixed set of memory-construction events in the EC-MDP of Section~\ref{sec:problem_formulation}.

\subsection{Preliminaries}
\label{app:convergence_setup}

Let $\mathcal T=\{(A_j,c_j,R_j)\}_j$ be a finite transition buffer, where $c_j=m_{\mathrm{new},j}$ is the child of the retrieved set $A_j$, and let $\mathcal G$ be the provenance DAG consisting of its recorded parent--child edges. Assume that every edge joins an older parent to a strictly newer child and that every nonterminal vertex occurs as a retrieved root. Index the $n\ge1$ nonterminal memories in chronological order and write $q_i=Q(m_i)$. The terminal scores $g_c$ are fixed, with $|R_j|\le R_{\max}$ and $|g_c|\le G_{\max}$. Throughout, $0\le\gamma<1$, $\lambda\in[0,1]$, and $\beta=\gamma\lambda$. We consider unclipped value updates on this fixed graph.

The stored-child bootstrap in Eq.~\ref{eq:td_error} is
\[
v_c(q)=
\begin{cases}
q_c,&c\text{ is nonterminal},\\
g_c,&c\text{ is terminal}.
\end{cases}
\]
Thus nonterminal children contribute their current values. Let $n_u=|\{j:u\in A_j\}|>0$ and define
\[
P_{uv}=\frac1{n_u}\sum_{j:u\in A_j}\mathbf 1\{c_j=v\},
\qquad
\bar r_u=\frac1{n_u}\sum_{j:u\in A_j}
\left[R_j+\gamma\mathbf 1\{c_j\text{ is terminal}\}g_{c_j}\right],
\]
where the terminal summand is zero otherwise. The matrix $P$ is nonnegative, substochastic, and strictly upper triangular. Define the memory-lineage Bellman operator by
\[
\mathcal T_{\mathcal G}q=\bar r+\gamma Pq.
\]

For finite $D$, fix $\mathcal L\subseteq\{0,\ldots,D\}$ with $0\in\mathcal L$. Let $\mathcal P_{\mathcal L}(u\rightsquigarrow i)$ denote the set of all backward paths from $u$ to $i$ whose lengths belong to $\mathcal L$, including the zero-length path for $i=u$. Distinct paths between the same pair of vertices are counted separately. The unpruned case has $\mathcal L=\{0,\ldots,D\}$; fixed geometric-weight pruning amounts to restricting $\mathcal L$.

Let $B_{iu}=1$ if $i$ is an immediate parent of $u$, and zero otherwise. Define
\begin{equation}
\label{eq:app_path_matrix}
W_{iu}=\sum_{p\in\mathcal P_{\mathcal L}(u\rightsquigarrow i)}\beta^{|p|},
\qquad
W=\sum_{d\in\mathcal L}\beta^dB^d,
\end{equation}
with the convention $\beta^0=1$ also at $\beta=0$. Put
\[
N_i=\sum_j\sum_{u\in A_j}|\mathcal P_{\mathcal L}(u\rightsquigarrow i)|,
\qquad
C=\operatorname{diag}(N_1,\ldots,N_n),
\qquad
D_n=\operatorname{diag}(n_1,\ldots,n_n).
\]
Then $N_i\ge n_i>0$. The normalized update corresponding to Eqs.~\ref{eq:td_error} and~\ref{eq:general_update} is
\begin{equation}
\label{eq:app_normalized_update}
q_{t+1}=q_t+\alpha_t h(q_t),
\qquad
h_i(q)=\frac1{N_i}\sum_j\sum_{u\in A_j}
\sum_{p\in\mathcal P_{\mathcal L}(u\rightsquigarrow i)}
\beta^{|p|}\bigl(R_j+\gamma v_{c_j}(q)-q_u\bigr).
\end{equation}
The residual in each summand is evaluated at the retrieved root $u$.

\newpage
\subsection{Convergence}
\label{app:overall_theorem}

\begin{theorem}[Convergence and fixed-point preservation]
\label{thm:memq_overall}
Under the preceding assumptions, the following hold.

\begin{enumerate}[label=(\roman*),leftmargin=*,itemsep=4pt]
\item The full-buffer update satisfies $h(q)=b-Hq$, where
\begin{equation}
\label{eq:app_factorization}
H=C^{-1}WD_n(I-\gamma P),
\qquad
b=C^{-1}WD_n\bar r.
\end{equation}
Its unique fixed point is
\begin{equation}
\label{eq:app_target}
q_{\mathcal G}^{\dagger}
=(I-\gamma P)^{-1}\bar r
=\sum_{k=0}^{n-1}\gamma^kP^k\bar r.
\end{equation}
This is also the fixed point of the depth-zero update. With $\mathcal T$ and the terminal scores fixed, it is independent of $\lambda$, $\mathcal L$, and the choice of positive diagonal normalizer.

\item For the path-count normalizer $C$ defined above, $H$ is upper triangular with eigenvalues $n_i/N_i\in(0,1]$. From any finite $q_0$, the recursion converges to $q_{\mathcal G}^{\dagger}$ for either a constant step
\[
0<\alpha<\frac{2}{\max_i(n_i/N_i)}
\]
(in particular, any $0<\alpha<2$), or deterministic positive steps with $\sum_t\alpha_t=\infty$ and $\sum_t\alpha_t^2<\infty$.

\item Consider random epochs of records from $\mathcal G$, with the same stored-child bootstrap and path-count normalization, and write $q_{t+1}=q_t+\alpha_t(b_t-H_tq_t)$. Then
\begin{equation}
\label{eq:app_coefficient_bound}
\|H_t\|_\infty\le1+\gamma,
\qquad
\|b_t\|_\infty\le R_{\max}+\gamma G_{\max}.
\end{equation}
Let $\mathcal F_t$ be the history before epoch $t$, so that $q_t$ is $\mathcal F_t$-measurable. Assume there exist deterministic, time-independent $\bar H,\bar b$ such that
\begin{equation}
\label{eq:app_stationary_mean}
\mathbb E[H_t\mid\mathcal F_t]=\bar H,
\qquad
\mathbb E[b_t\mid\mathcal F_t]=\bar b,
\end{equation}
and suppose
\[
d_i=\mathbb E\!\left[\frac{n_{t,i}}{N_{t,i}}\,\middle|\,\mathcal F_t\right]>0.
\]
Here $n_{t,i}$ and $N_{t,i}$ are the epoch's direct-root and path counts, respectively, with $n_{t,i}/N_{t,i}=0$ when $N_{t,i}=0$. If $q_0$ is finite and deterministic and the deterministic steps satisfy $\alpha_t>0$, $\sum_t\alpha_t=\infty$, and $\sum_t\alpha_t^2<\infty$, then
\[
q_t\longrightarrow\bar H^{-1}\bar b
\qquad\text{almost surely}.
\]
In particular, $\sup_t\|q_t\|<\infty$ almost surely. If $\bar H=H$ and $\bar b=b$, the limit is $q_{\mathcal G}^{\dagger}$.
\end{enumerate}
\end{theorem}

\begin{proof}
For each root $u$,
\[
\sum_{j:u\in A_j}\bigl(R_j+\gamma v_{c_j}(q)-q_u\bigr)
=n_u\bigl(\bar r_u+\gamma(Pq)_u-q_u\bigr).
\]
Since $(B^d)_{iu}$ counts length-$d$ paths from $i$ to $u$, Eqs.~\ref{eq:app_path_matrix} and~\ref{eq:app_normalized_update} imply
\[
h(q)=C^{-1}WD_n(\bar r+\gamma Pq-q).
\]
This gives Eq.~\ref{eq:app_factorization}.

The matrices $B$ and $P$ are strictly upper triangular, while $W$ is upper triangular with unit diagonal. Since $C$ and $D_n$ have positive diagonals, $C^{-1}WD_n$ is invertible. Hence
\[
h(q)=0
\quad\Longleftrightarrow\quad
\bar r+\gamma Pq-q=0.
\]
As $P^n=0$, we have $(I-\gamma P)^{-1}=\sum_{k=0}^{n-1}\gamma^kP^k$, which proves Eq.~\ref{eq:app_target}. The depth-zero update has $W=I$ and $C=D_n$, and therefore equals $\mathcal T_{\mathcal G}q-q$. Replacing $C$ by any positive diagonal matrix leaves the zero of $h$ unchanged. This proves (i).

All factors in Eq.~\ref{eq:app_factorization} are upper triangular, and $W$ and $I-\gamma P$ have unit diagonal. Thus
\[
H_{ii}=\frac{n_i}{N_i}>0,
\qquad
\operatorname{spec}(H)=\{n_i/N_i:1\le i\le n\}\subset(0,1].
\]
In particular, $H$ is positive stable. For a constant step,
\[
q_{t+1}-q_{\mathcal G}^{\dagger}
=(I-\alpha H)(q_t-q_{\mathcal G}^{\dagger}).
\]
The eigenvalues of $I-\alpha H$ are $1-\alpha n_i/N_i$. Under the stated restriction on $\alpha$, their moduli are strictly less than one, so $(I-\alpha H)^t\to0$. The diminishing-step case of (ii) is a special case of (iii), proved below.

For the stochastic recursion, define
\begin{equation}
\label{eq:app_stochastic_weights}
z_{t,j,u}(i)=
\begin{cases}
\displaystyle\frac1{N_{t,i}}
\sum_{p\in\mathcal P_{\mathcal L}(u\rightsquigarrow i)}\beta^{|p|},
&N_{t,i}>0,\\[5pt]
0,&N_{t,i}=0.
\end{cases}
\end{equation}
Each term $\beta^{|p|}$ in Eq.~\ref{eq:app_stochastic_weights} lies in $[0,1]$, whereas $N_{t,i}$ counts paths without weighting. Consequently,
\[
0\le\sum_j\sum_{u\in A_{t,j}}z_{t,j,u}(i)\le1.
\]
Let $e_u$ denote the $u$th coordinate vector and write $v_{c_j}(q)=\psi_{t,j}^{\top}q+\kappa_{t,j}$, with $(\psi_{t,j},\kappa_{t,j})=(e_{c_j},0)$ for a nonterminal child and $(0,g_{c_j})$ otherwise. Then $\|\psi_{t,j}\|_1\le1$, $|\kappa_{t,j}|\le G_{\max}$, and
\begin{align}
H_t&=\sum_j\sum_{u\in A_{t,j}}
z_{t,j,u}(e_u-\gamma\psi_{t,j})^\top,\notag\\
b_t&=\sum_j\sum_{u\in A_{t,j}}
z_{t,j,u}(R_{t,j}+\gamma\kappa_{t,j}).
\label{eq:app_stochastic_coefficients}
\end{align}
The absolute sum of row $i$ of $H_t$ is bounded by
\[
\sum_{j,u}z_{t,j,u}(i)\|e_u-\gamma\psi_{t,j}\|_1
\le(1+\gamma)\sum_{j,u}z_{t,j,u}(i)\le1+\gamma.
\]
The analogous bound for $b_t$ proves Eq.~\ref{eq:app_coefficient_bound}. It also follows that
\[
\|b_t-H_tq\|_\infty
\le R_{\max}+\gamma G_{\max}+(1+\gamma)\|q\|_\infty.
\]

In Eq.~\ref{eq:app_stochastic_coefficients}, $z_{t,j,u}(i)>0$ only if $i=u$ or $i$ is an ancestor of $u$. Nonzero columns of the corresponding summand can occur only at $u$ and, if nonterminal, its strictly newer child. Thus $H_t$ is upper triangular, and its diagonal receives only the zero-length contribution:
\[
(H_t)_{ii}=\frac{n_{t,i}}{N_{t,i}},
\qquad
\bar H_{ii}=d_i>0.
\]
Hence $\bar H$ is upper triangular and positive stable.

It remains to prove almost-sure convergence. We use a standard stochastic-approximation argument \citep{borkar2000ode}. Set $q^\dagger=\bar H^{-1}\bar b$, $e_t=q_t-q^\dagger$, and
\[
M_{t+1}=(b_t-H_tq_t)-(\bar b-\bar Hq_t).
\]
By Eq.~\ref{eq:app_stationary_mean}, $\mathbb E[M_{t+1}\mid\mathcal F_t]=0$. Equation~\ref{eq:app_coefficient_bound} and equivalence of finite-dimensional norms give, for a deterministic $K_0<\infty$,
\begin{equation}
\label{eq:app_noise_bound}
\mathbb E[\|M_{t+1}\|_2^2\mid\mathcal F_t]
\le K_0(1+\|q_t\|_2^2).
\end{equation}
Positive stability of $\bar H$ implies that there exists a symmetric $S\succ0$ satisfying
\begin{equation}
\label{eq:app_lyapunov}
\bar H^\top S+S\bar H=I.
\end{equation}
With $L_t=e_t^\top Se_t$ and $e_{t+1}=e_t-\alpha_t\bar H e_t+\alpha_tM_{t+1}$, Eqs.~\ref{eq:app_noise_bound} and~\ref{eq:app_lyapunov} yield
\begin{align}
\mathbb E[L_{t+1}\mid\mathcal F_t]
&=L_t-\alpha_t\|e_t\|_2^2
+\alpha_t^2e_t^\top\bar H^\top S\bar H e_t
+\alpha_t^2\mathbb E[M_{t+1}^\top SM_{t+1}\mid\mathcal F_t]\notag\\
&\le(1+K\alpha_t^2)L_t-\alpha_t\|e_t\|_2^2+K\alpha_t^2
\label{eq:app_lyapunov_drift}
\end{align}
for a deterministic $K<\infty$. Bounded coefficients and finite deterministic $q_0$ imply finite second moments at every finite time. Since $\sum_t\alpha_t^2<\infty$, the almost-supermartingale theorem of \citet{robbins1971convergence}, applied to Eq.~\ref{eq:app_lyapunov_drift}, gives almost-sure convergence of $L_t$ to a finite limit and
\begin{equation}
\label{eq:app_summability}
\sum_t\alpha_t\|e_t\|_2^2<\infty
\qquad\text{almost surely}.
\end{equation}
On any sample path where $\lim_t L_t>0$, the inequality $L_t\le\lambda_{\max}(S)\|e_t\|_2^2$ bounds $\|e_t\|_2^2$ away from zero for all sufficiently large $t$. Together with $\sum_t\alpha_t=\infty$, this contradicts Eq.~\ref{eq:app_summability}. Thus $L_t\to0$ and, since $S\succ0$, $q_t\to q^\dagger$ almost surely. Boundedness follows.

When $\bar H=H$ and $\bar b=b$, Eq.~\ref{eq:app_target} gives $q^\dagger=q_{\mathcal G}^{\dagger}$. Taking $H_t=H$ and $b_t=b$ deterministically also proves the remaining assertion in (ii).
\end{proof}

\paragraph{Remark.}
The matrix $C^{-1}WD_n$ is an invertible preconditioner of $\mathcal T_{\mathcal G}q-q$, so the full-buffer operator in part~(i) preserves $q_{\mathcal G}^{\dagger}$. Independent epochs from a fixed replay law satisfy Eq.~\ref{eq:app_stationary_mean}. With epoch-wise normalization, part~(iii) gives convergence to $\bar H^{-1}\bar b$ under its stated assumptions; this limit equals $q_{\mathcal G}^{\dagger}$ when $\bar H=H$ and $\bar b=b$.

\newpage
\subsection{Set-Value Approximation Error}
\label{app:set_value_approximation}

Fix a policy $\pi$ and a task--memory state $(s,\mathcal M)$. Let $\mathcal A_s$ be a finite, nonempty collection of candidate retrieval sets $A\subseteq\mathcal M$ satisfying $1\le |A|\le k$. For $A\in\mathcal A_s$, write
\[
Q(A)=Q^\pi(s,A;\mathcal M),
\qquad
\widetilde Q(A)=\frac1{|A|}\sum_{m\in A}q^\pi(m),
\qquad
\xi(A)=\xi^\pi(s,A;\mathcal M).
\]
By Eq.~\ref{eq:value_decomposition}, $Q(A)=\widetilde Q(A)+\xi(A)$. Define $\varepsilon_{\mathrm{int}}=\max_{A\in\mathcal A_s}|\xi(A)|$. A uniform bound on the residual over task--memory states and candidate sets can be used in its place.

\begin{proposition}[Set-value approximation]
\label{prop:set_value_approximation}
Let
\[
A^\star\in\operatorname*{arg\,max}_{A\in\mathcal A_s}Q(A),
\qquad
\widehat A\in\operatorname*{arg\,max}_{A\in\mathcal A_s}\widetilde Q(A).
\]
Then
\[
\max_{A\in\mathcal A_s}|Q(A)-\widetilde Q(A)|\le\varepsilon_{\mathrm{int}},
\qquad
0\le Q(A^\star)-Q(\widehat A)\le2\varepsilon_{\mathrm{int}}.
\]
If
\[
Q(A^\star)-Q(A)>2\varepsilon_{\mathrm{int}}
\qquad\text{for every }A\in\mathcal A_s\setminus\{A^\star\},
\]
then $\widehat A=A^\star$.
\end{proposition}

\begin{proof}
The approximation-error bound follows from the definition of $\varepsilon_{\mathrm{int}}$. Optimality of $A^\star$ gives nonnegative regret. By optimality of $\widehat A$ for $\widetilde Q$,
\[
\begin{aligned}
Q(A^\star)-Q(\widehat A)
&=\xi(A^\star)+\widetilde Q(A^\star)-\widetilde Q(\widehat A)-\xi(\widehat A)\\
&\le |\xi(A^\star)|+|\xi(\widehat A)|
\le2\varepsilon_{\mathrm{int}}.
\end{aligned}
\]
For every $A\ne A^\star$, the margin condition gives
\[
\widetilde Q(A^\star)-\widetilde Q(A)
=Q(A^\star)-Q(A)-\xi(A^\star)+\xi(A)
\ge Q(A^\star)-Q(A)-2\varepsilon_{\mathrm{int}}>0.
\]
Thus $A^\star$ is also the unique maximizer of $\widetilde Q$.
\end{proof}

When $\xi(A)=0$ for every candidate set, the approximation is exact and the selection regret is zero. In particular, for fixed retrieval cardinality $k$, an additive value $Q(A)=\sum_{m\in A}u(m)$ is represented exactly by choosing $q^\pi(m)=k\,u(m)$.

\endgroup

\subsection{\texorpdfstring{{Computational Complexity}}{Computational Complexity}}
\label{app:computational_cost}

{For a bank of $N$ memories, MemQ stores one Q-value and at most $k$ parent IDs per entry, requiring $O(Nk)$ metadata, or linear space for fixed $k$. Relative to MemRL, which already maintains per-memory values, the graph edges are the additional persistent state. Update cost depends on the paths traversed: every-path propagation can accumulate $O(|A_j|\sum_{d=0}^{D}k^d)$ contributions per task. The implementation bounds this work through depth truncation and discount-floor pruning.}
\clearpage

% The following stronger-backbone BFCL result is retained for reference but
% intentionally commented out of the current arXiv version.
% \begin{table}[t]
% \centering
% \caption{BFCL results with DeepSeek-V4-Flash as the shared backbone (\%).}
% \label{tab:strong_backbone}
% \small
% \begin{tabular}{lccc}
% \toprule
% Method & Cumulative SR & Final runtime SR & Evaluation SR \\
% \midrule
% MemQ & \textbf{88.19} & \textbf{85.43} & \textbf{71.29} \\
% MemRL & 87.44 & 84.17 & 63.37 \\
% MemP & 82.66 & 76.38 & 64.36 \\
% RAG & 80.90 & 65.08 & 65.35 \\
% Self-RAG & 84.20 & 65.83 & 59.41 \\
% \bottomrule
% \end{tabular}
% \end{table}

\section{Additional Learning Curves}
\label{app:csr_curves}

\begin{figure}[H]
\centering
\includegraphics[width=0.245\linewidth]{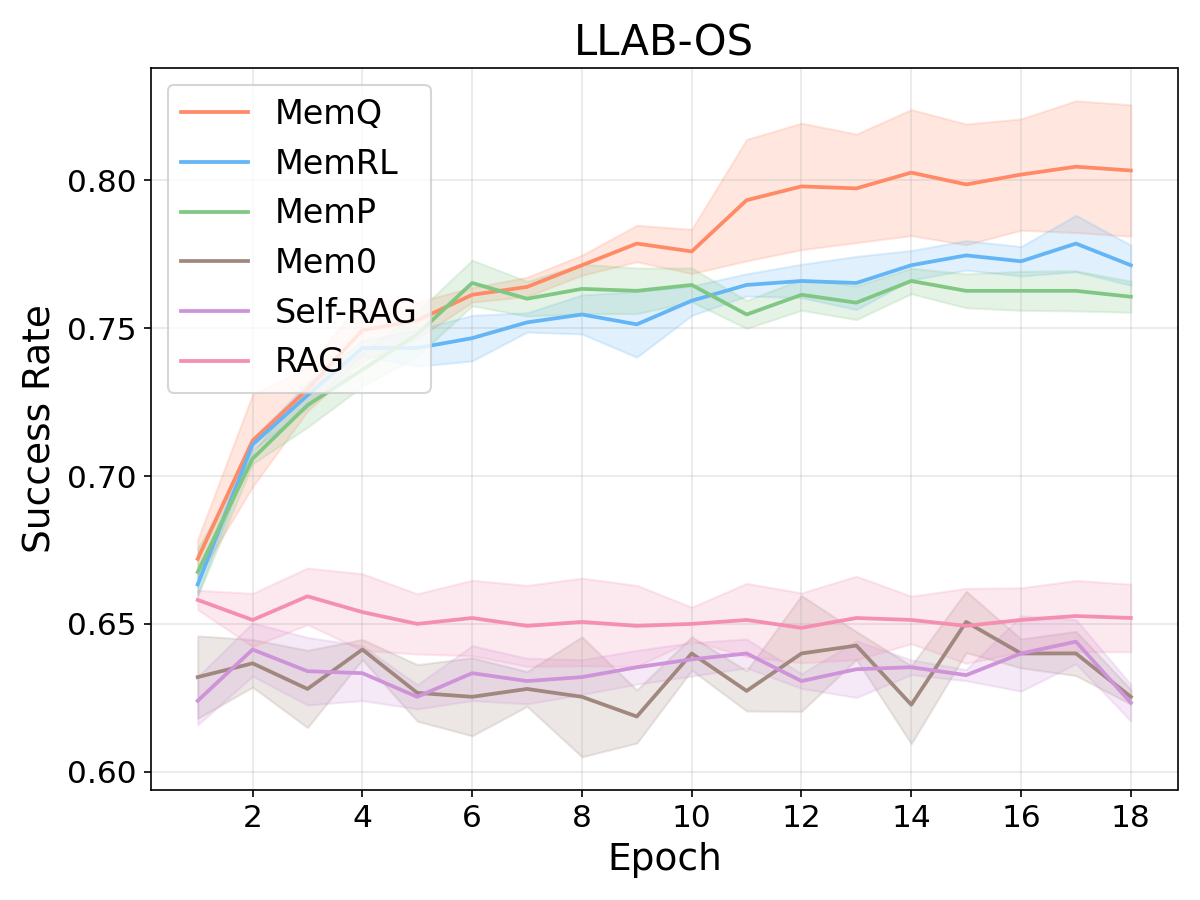}%
\hfill%
\includegraphics[width=0.245\linewidth]{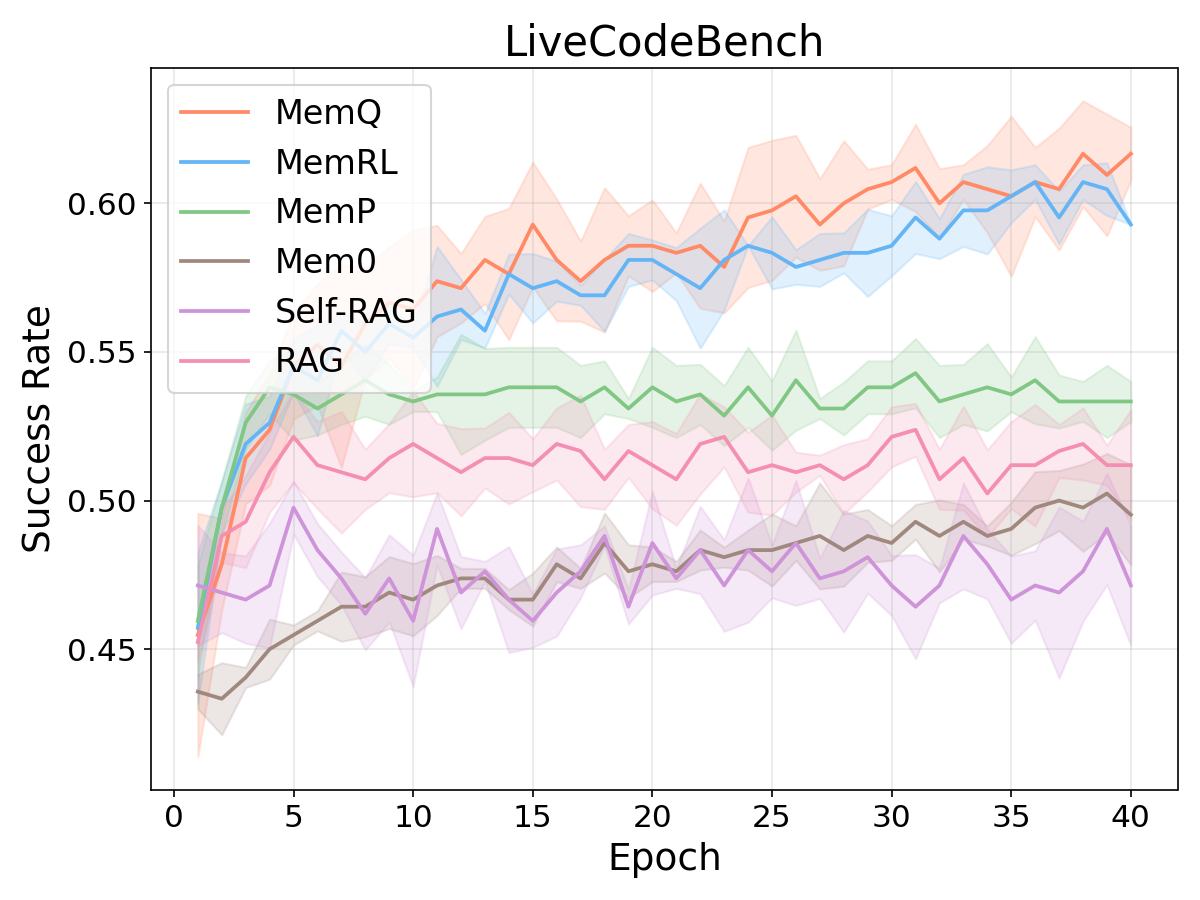}%
\hfill%
\includegraphics[width=0.245\linewidth]{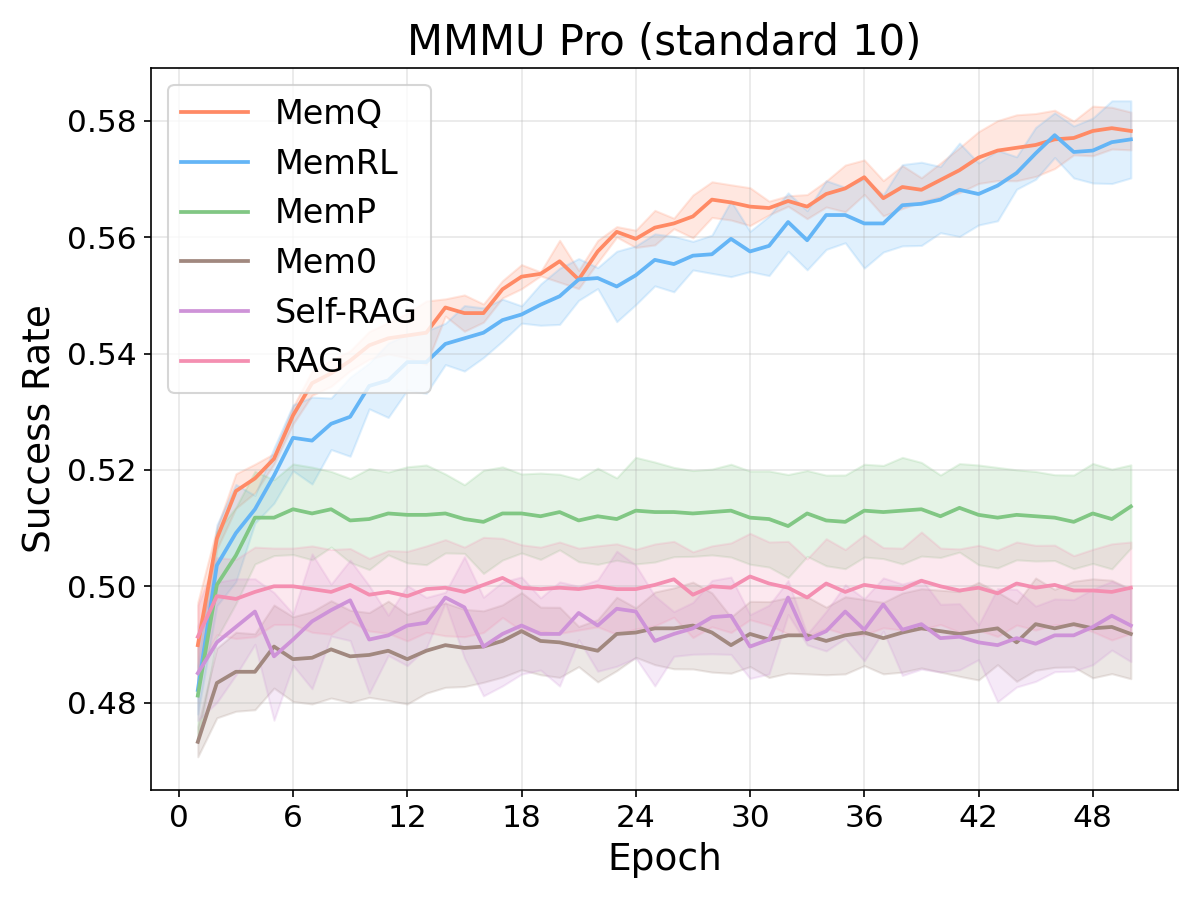}%
\hfill%
\includegraphics[width=0.245\linewidth]{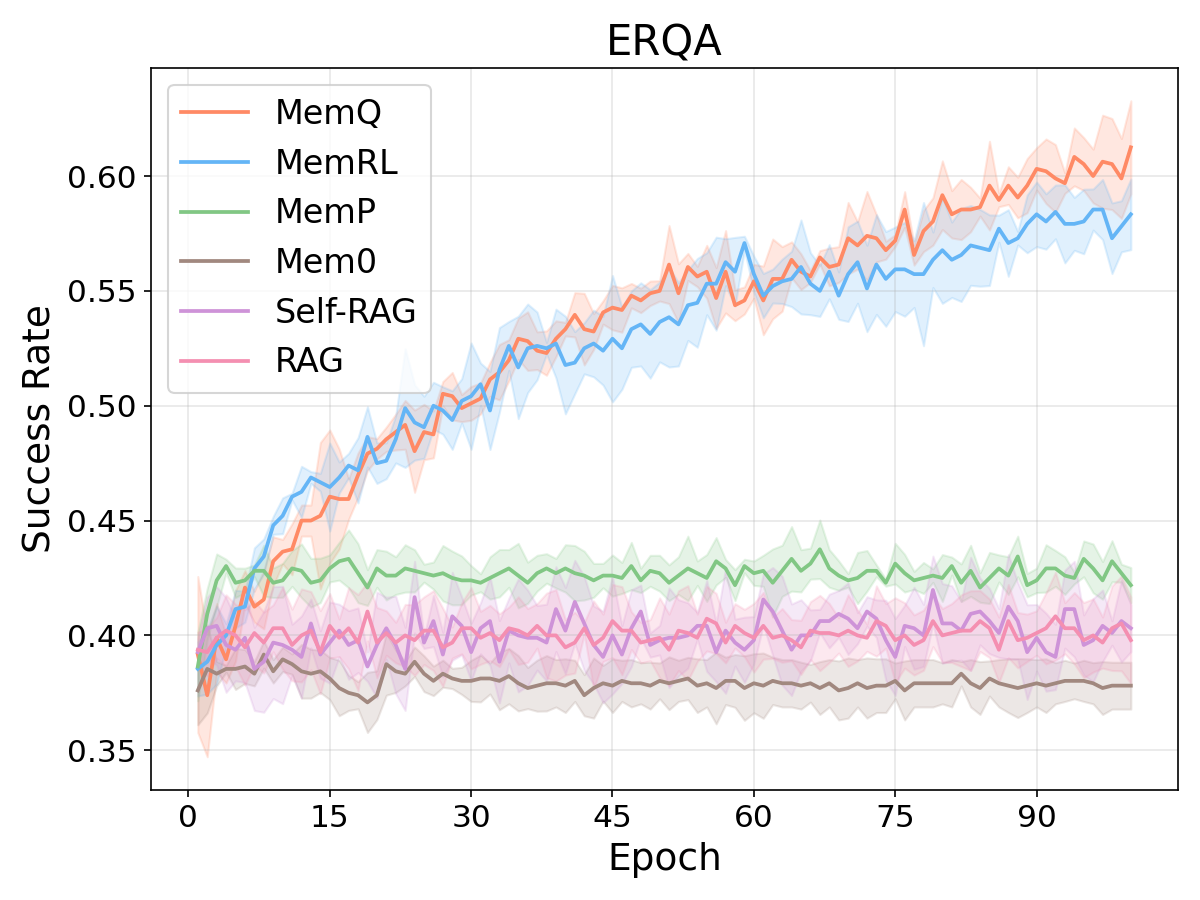}%
\\[5pt]
\includegraphics[width=0.245\linewidth]{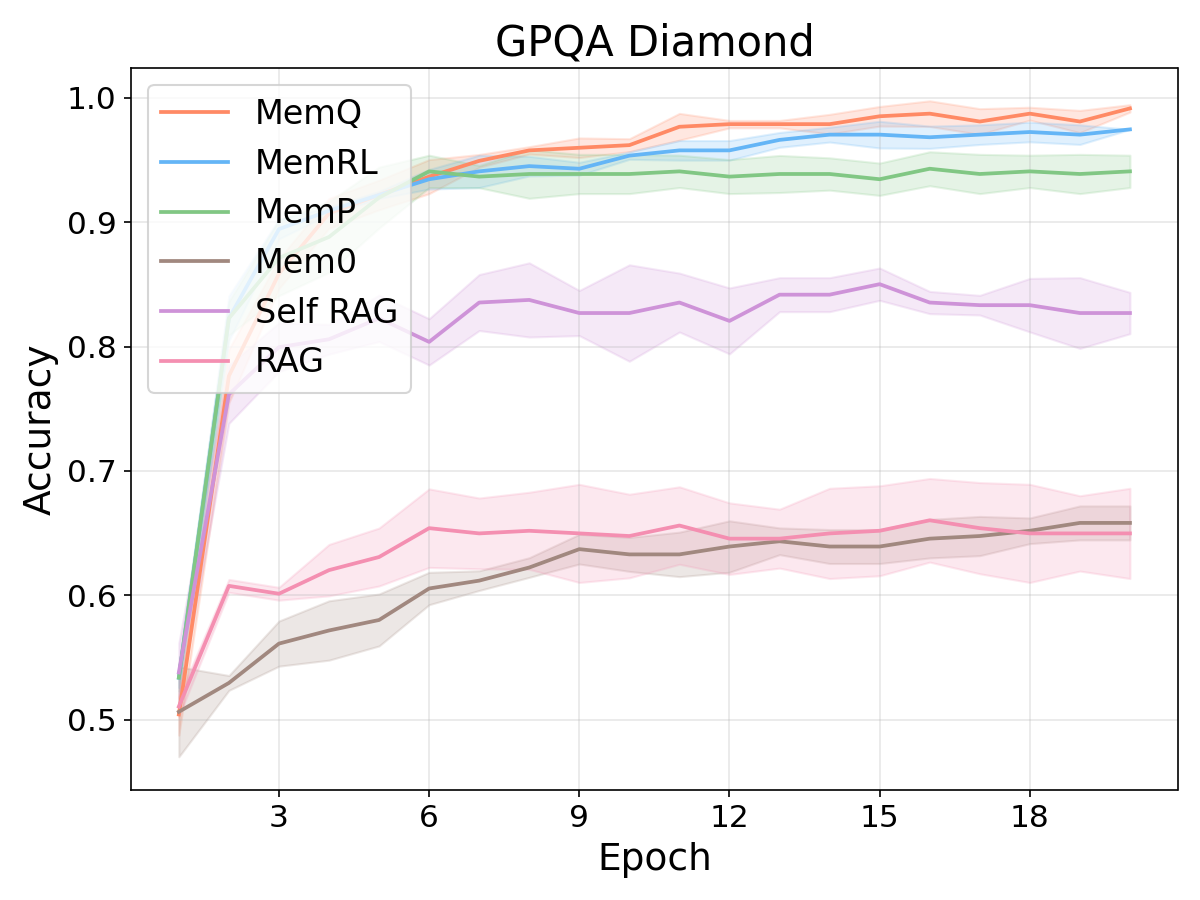}%
\hfill%
\includegraphics[width=0.245\linewidth]{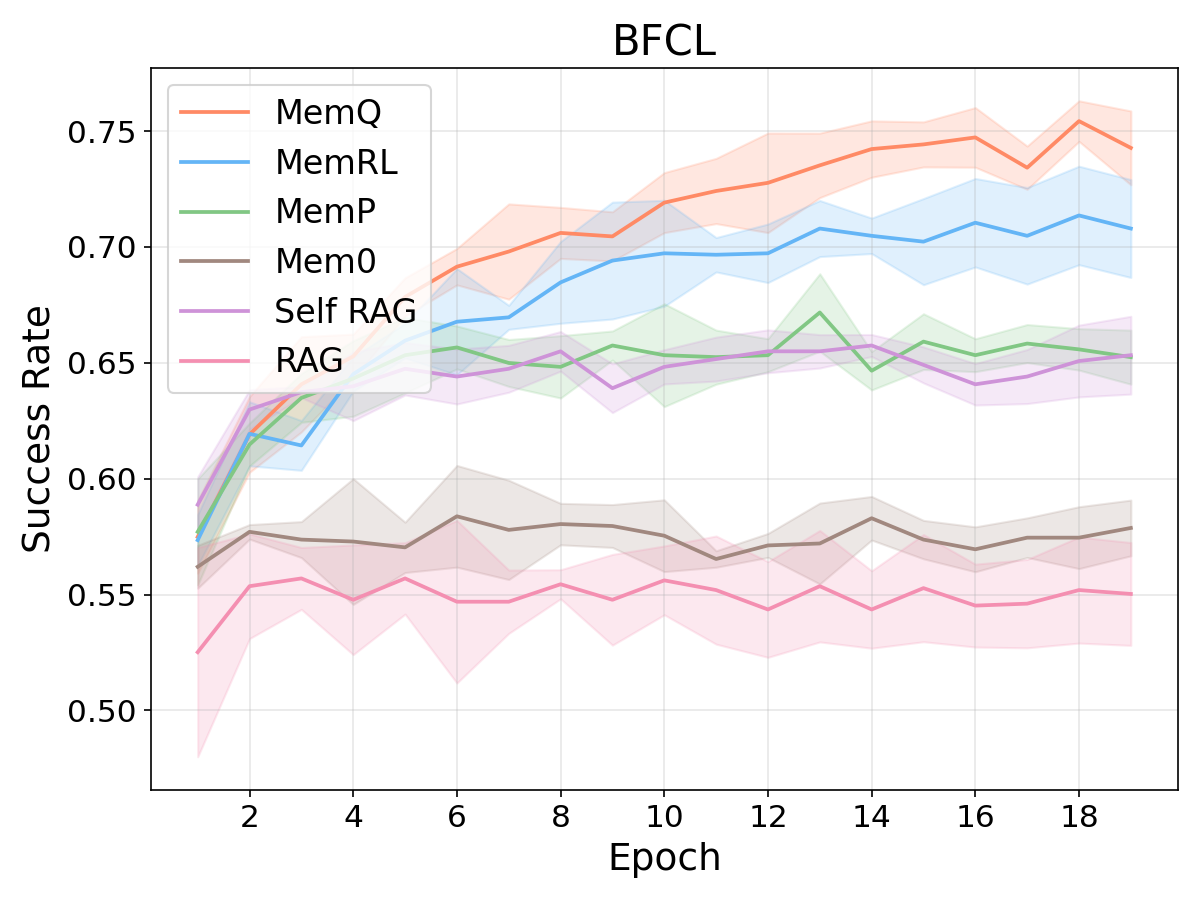}%
\hfill%
\includegraphics[width=0.245\linewidth]{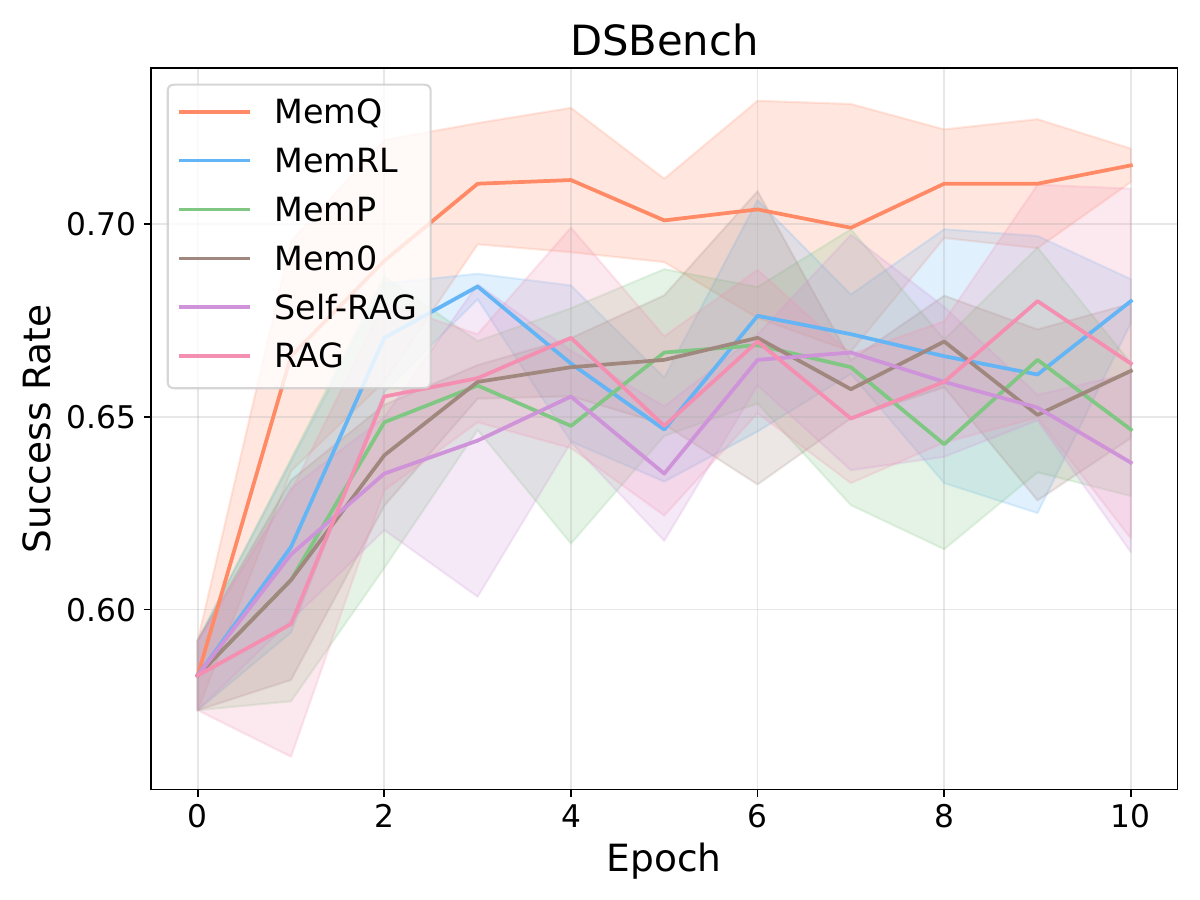}%
\hfill%
\includegraphics[width=0.245\linewidth]{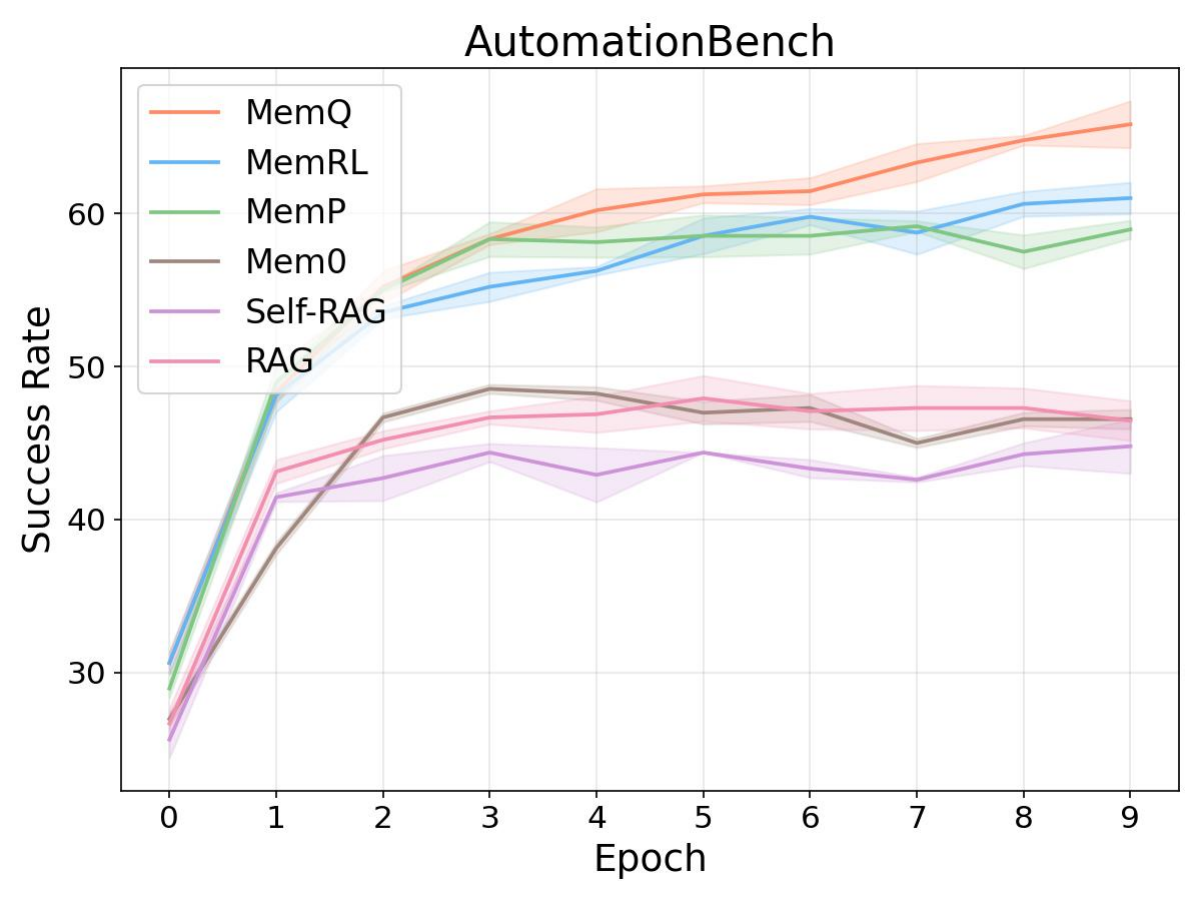}%
\caption{{Runtime learning dynamics (success rate vs.\ epoch) across eight benchmarks. Top row: LLAB, LiveCodeBench, MMMU-Pro, and ERQA. Bottom row: GPQA, BFCL, DSBench, and AutomationBench. DSBench curves show means $\pm1$ sample SD.}}
\label{fig:learning_curves}
\end{figure}

\begin{figure}[H]
\centering
\includegraphics[width=0.245\linewidth]{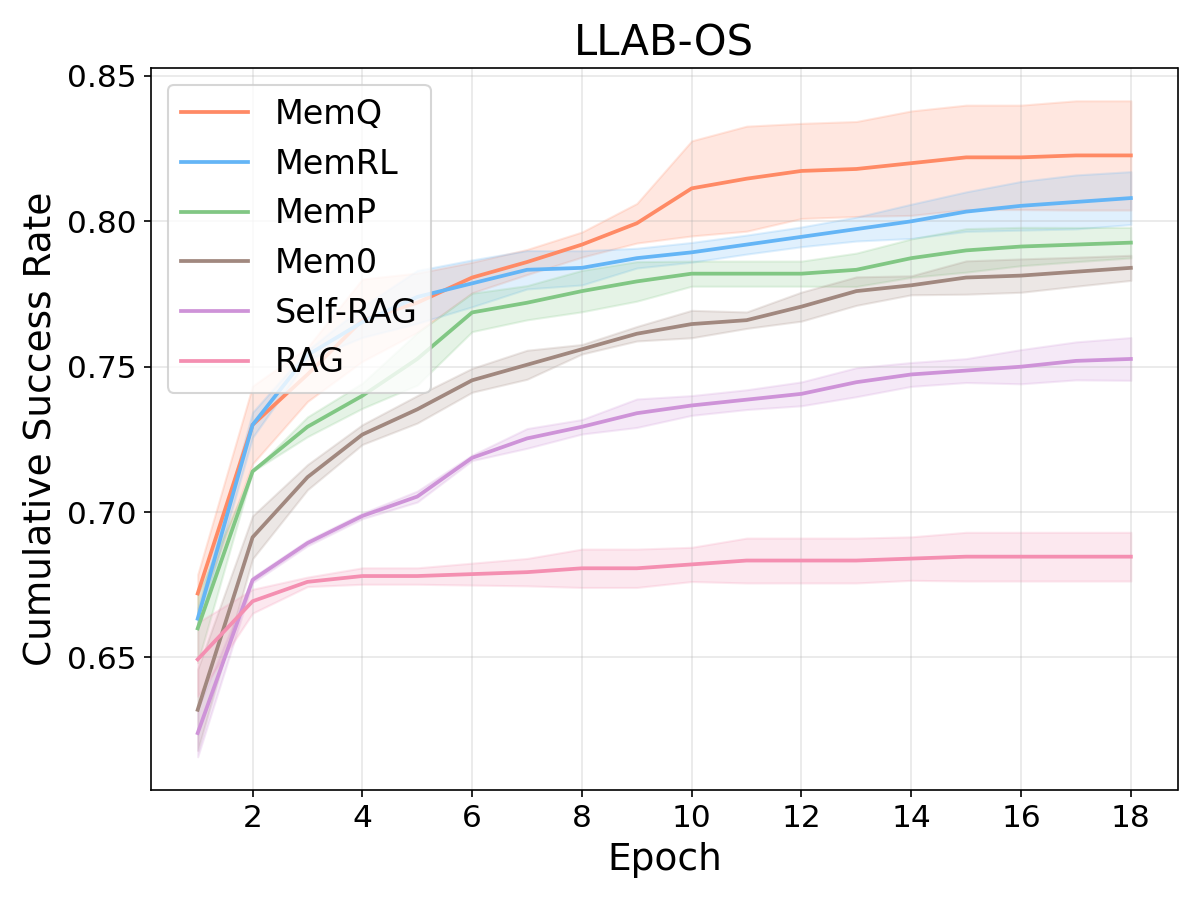}%
\hfill%
\includegraphics[width=0.245\linewidth]{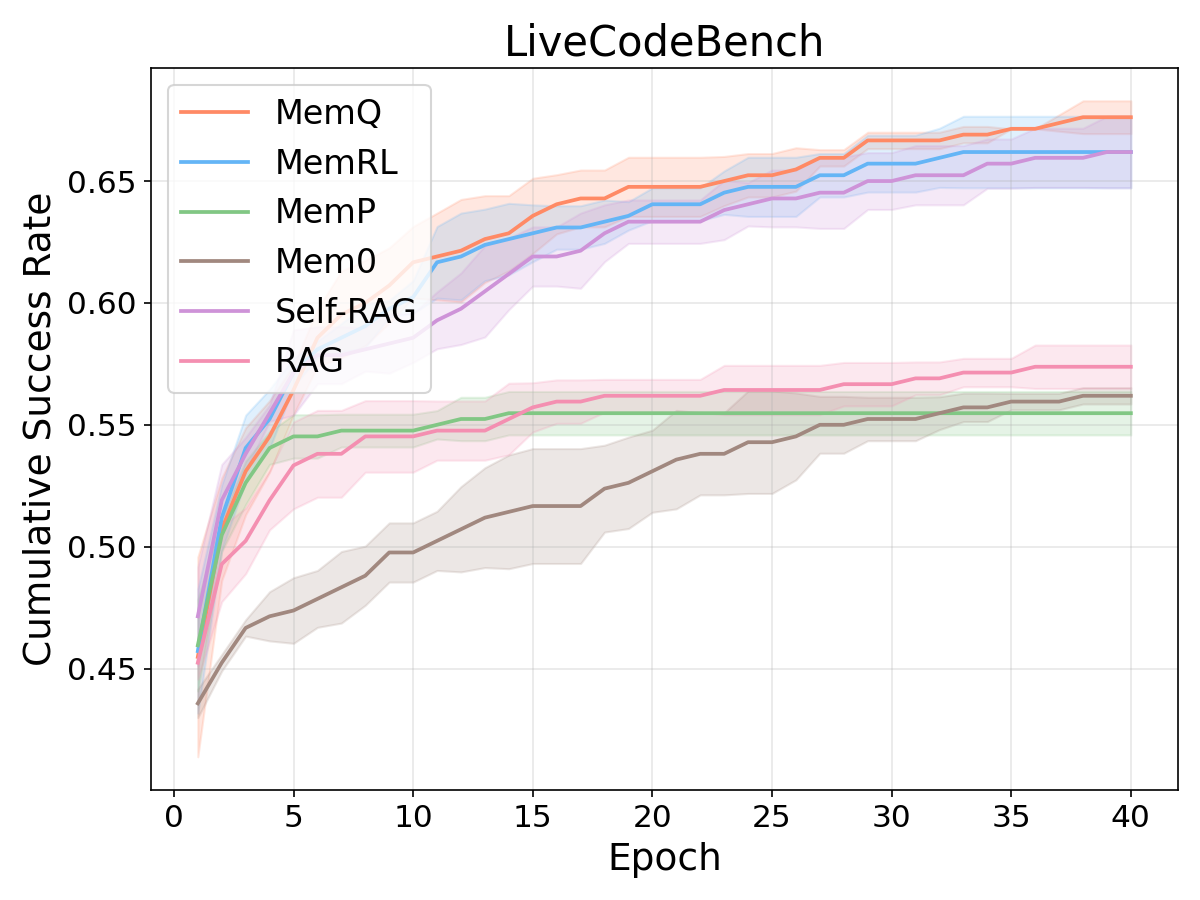}%
\hfill%
\includegraphics[width=0.245\linewidth]{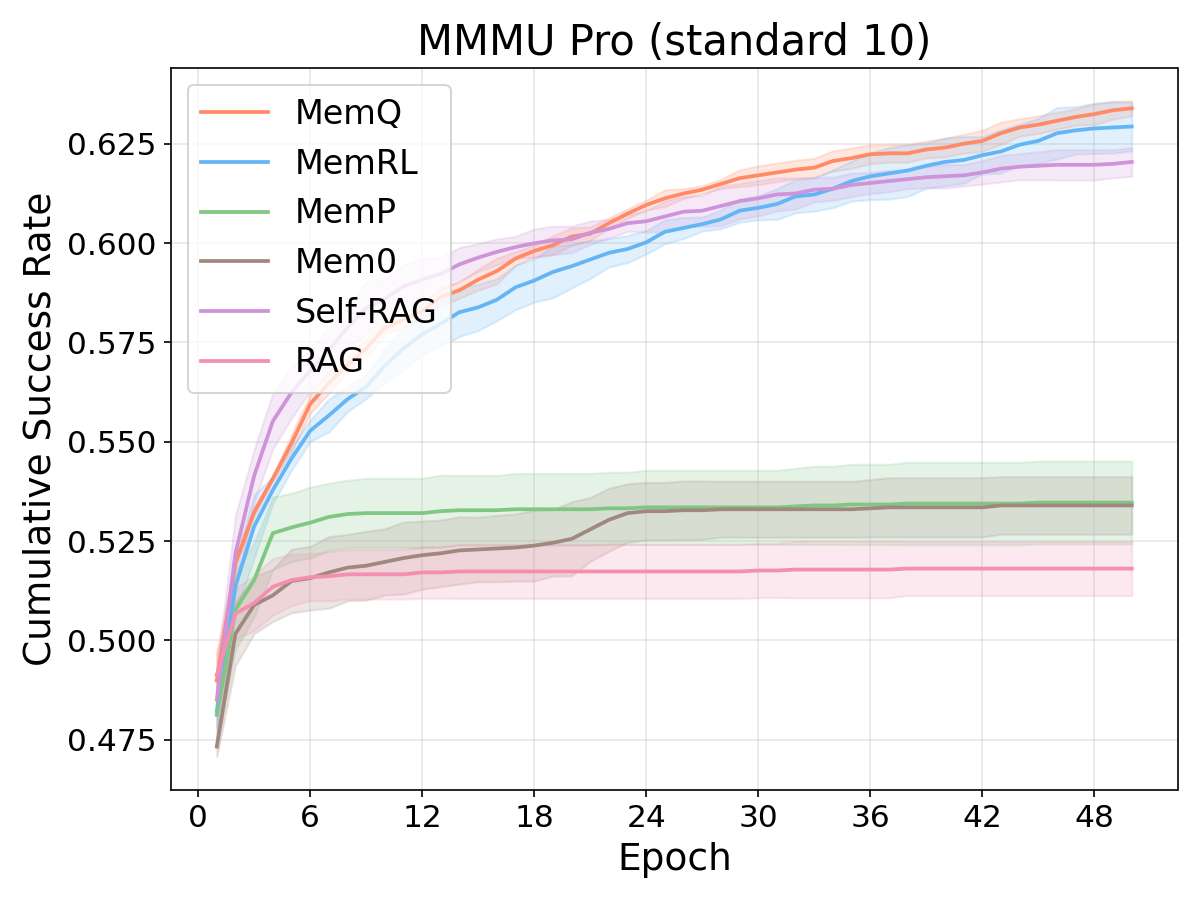}%
\hfill%
\includegraphics[width=0.245\linewidth]{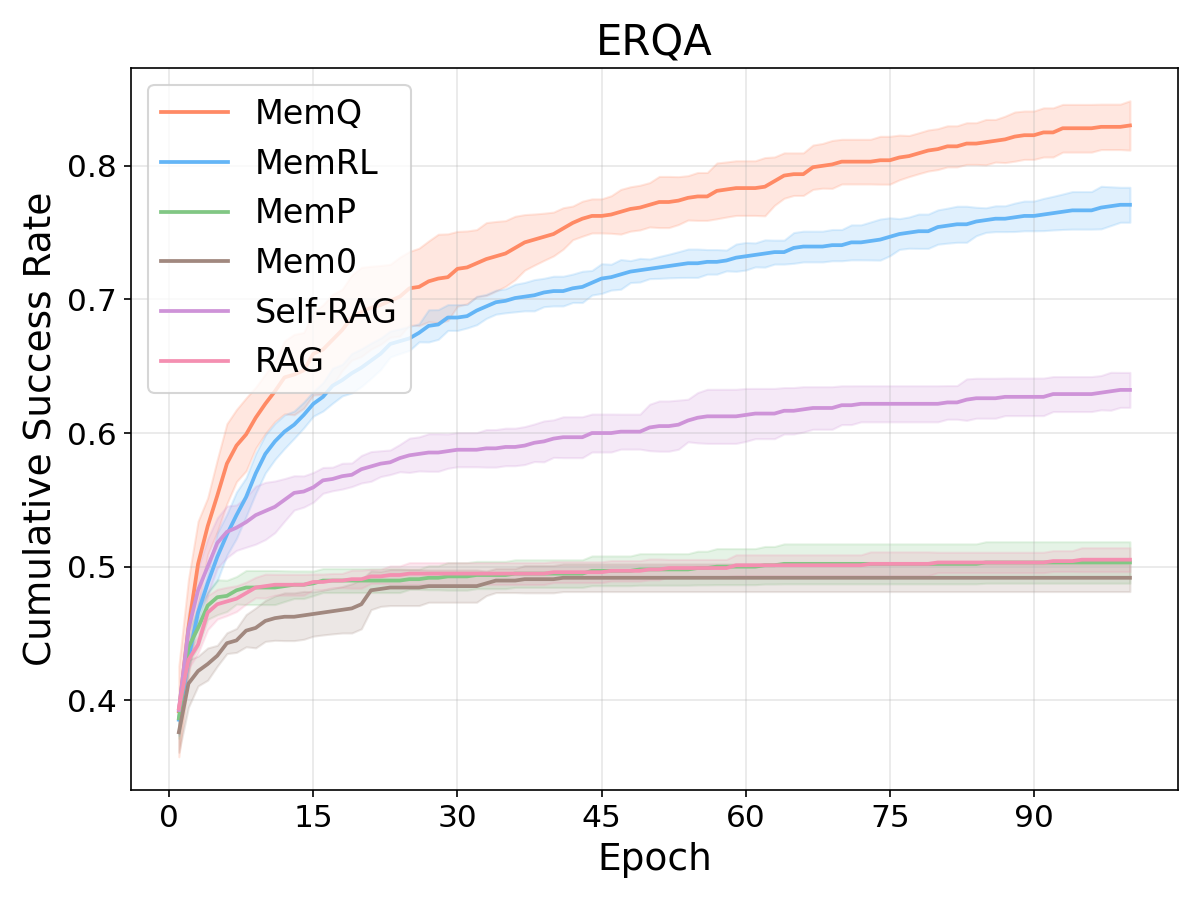}%
\\[5pt]
\includegraphics[width=0.245\linewidth]{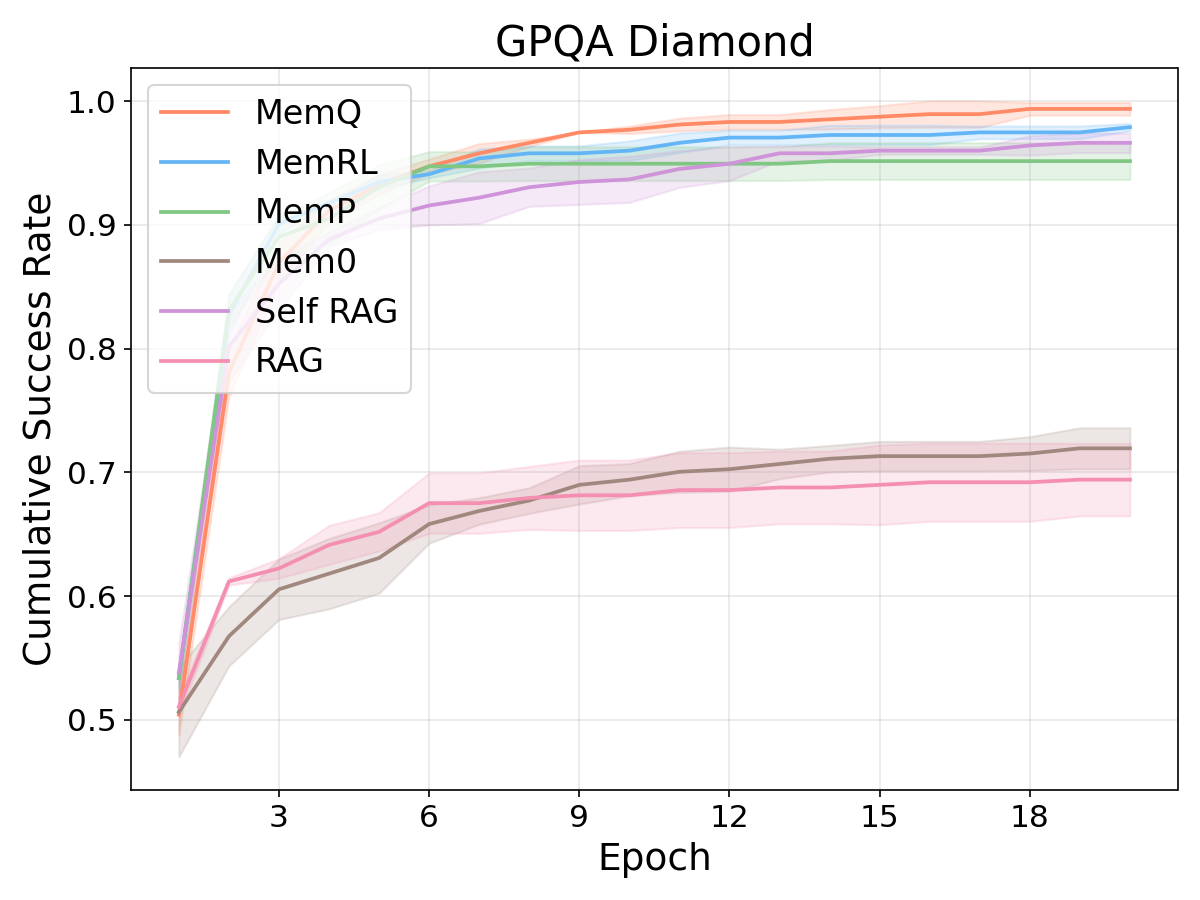}%
\hfill%
\includegraphics[width=0.245\linewidth]{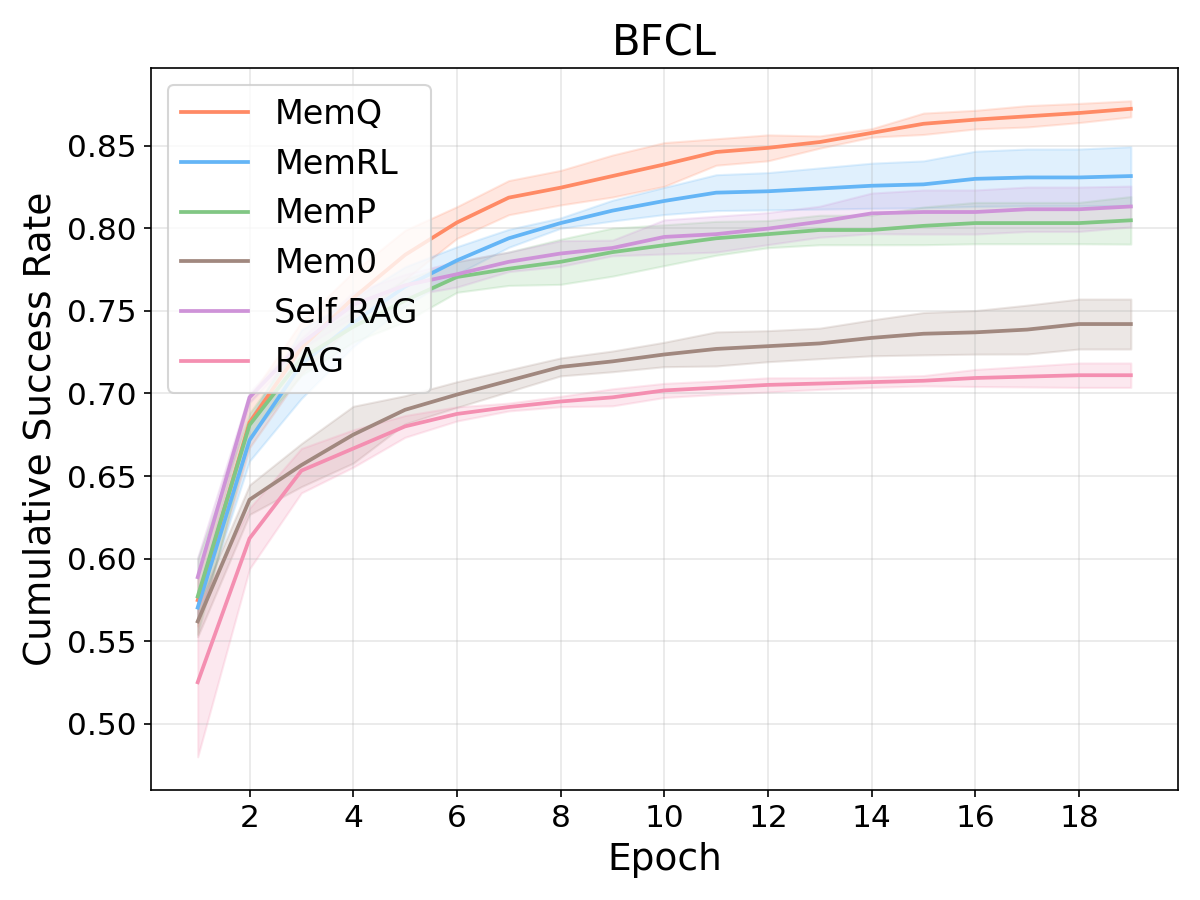}%
\hfill%
\includegraphics[width=0.245\linewidth]{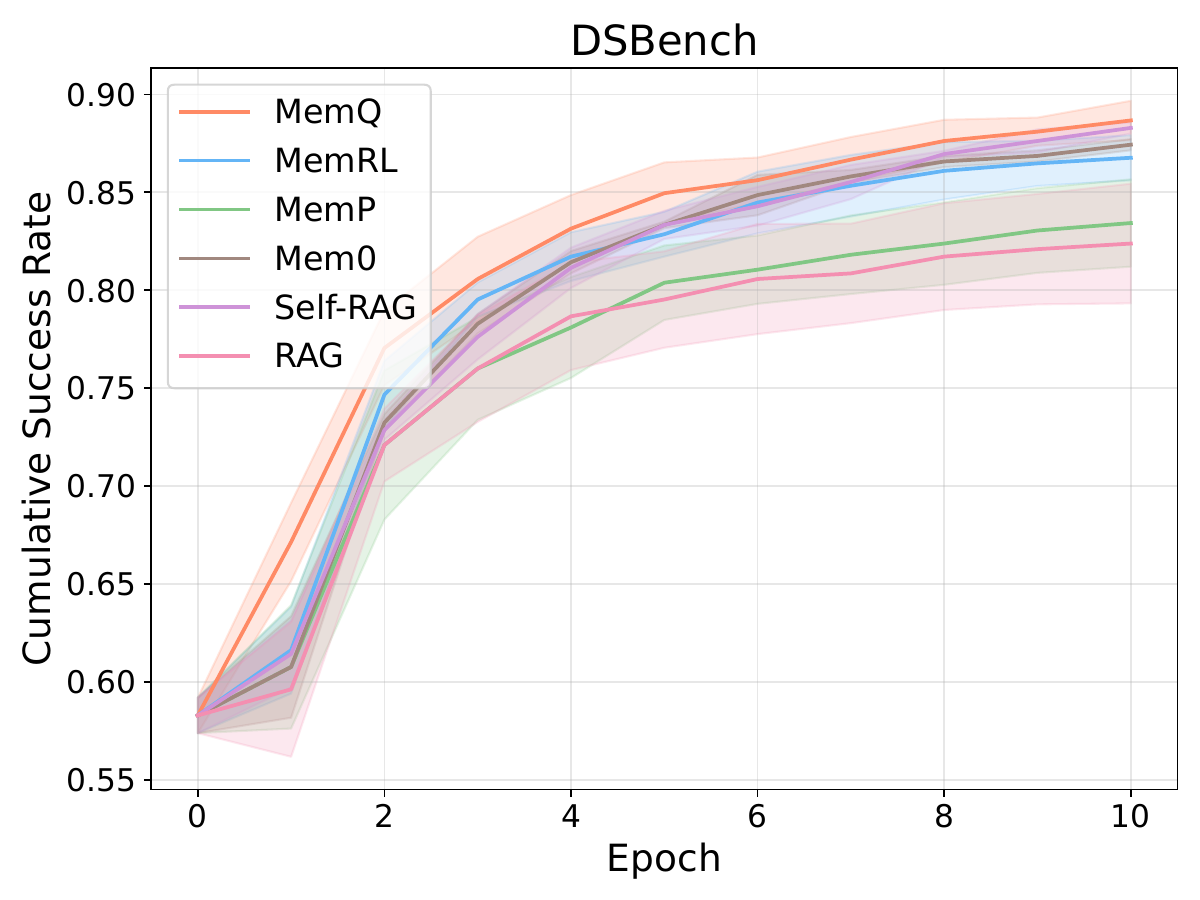}%
\hfill%
\includegraphics[width=0.245\linewidth]{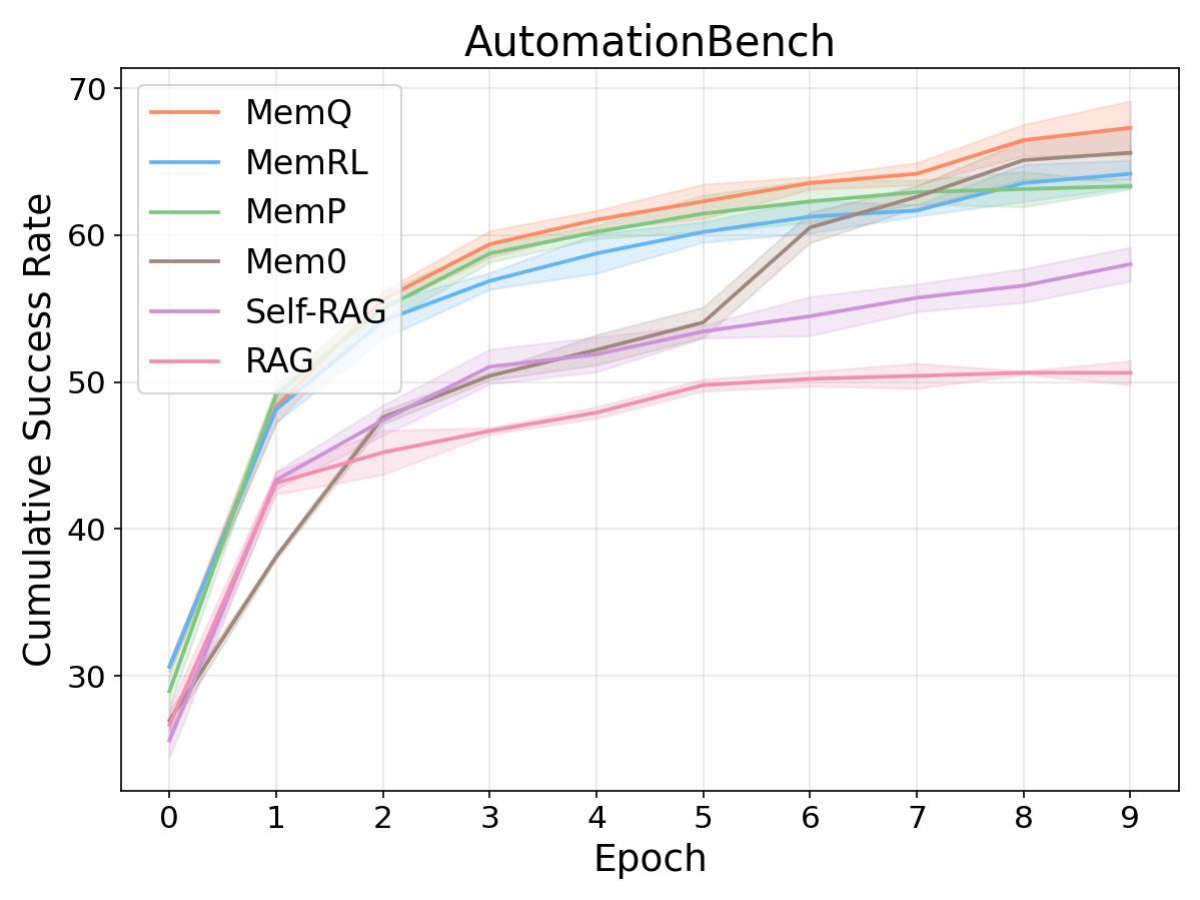}%
\caption{{Cumulative success rate over epochs across eight benchmarks, in the same order as Figure~\ref{fig:learning_curves}. CSR counts tasks solved at least once through each epoch and is nondecreasing by construction. DSBench curves show means $\pm1$ sample SD.}}
\label{fig:learning_curves_csr}
\end{figure}

\section{\texorpdfstring{{Additional AutomationBench Value Analysis}}{Additional AutomationBench Value Analysis}}
\label{app:automationbench_analysis}

{Complementing the retrieval-time analysis in Section~\ref{sec:memory_values}, we examine checkpoint memory composition by Q range. Each memory is labeled by whether its originating trajectory succeeded.}

\begin{figure}[H]
\centering
\begin{subfigure}[t]{0.49\linewidth}
  \centering
  \includegraphics[width=\linewidth]{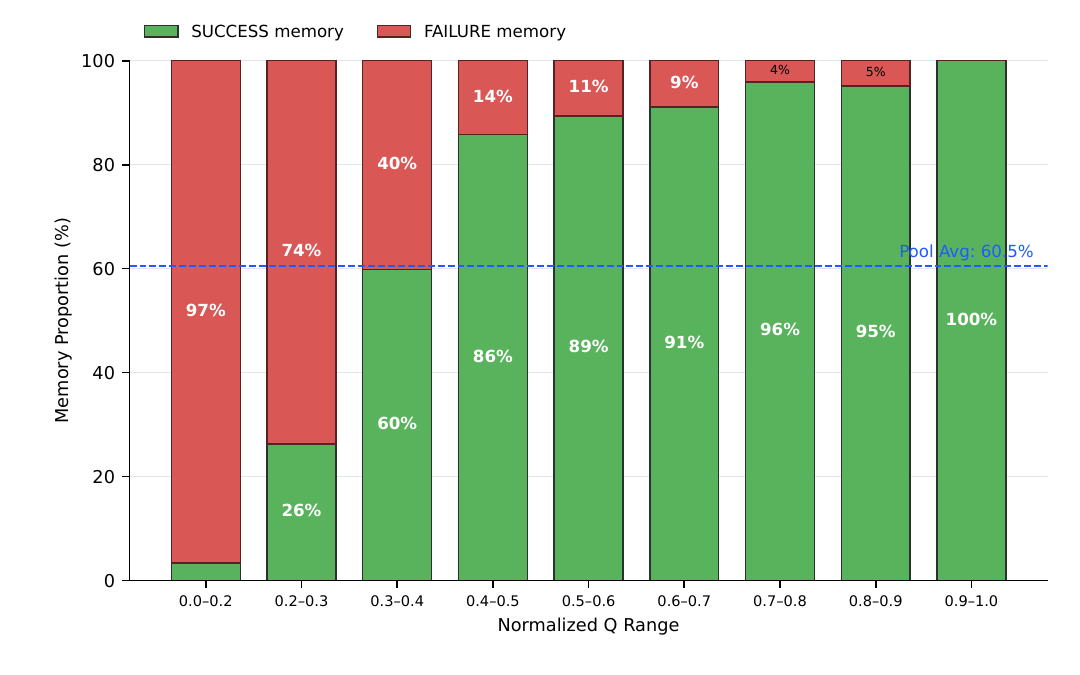}
  \caption{{MemQ checkpoint composition.}}
  \label{fig:automationbench_q_predictive_b}
\end{subfigure}
\hfill
\begin{subfigure}[t]{0.49\linewidth}
  \centering
  \includegraphics[width=\linewidth]{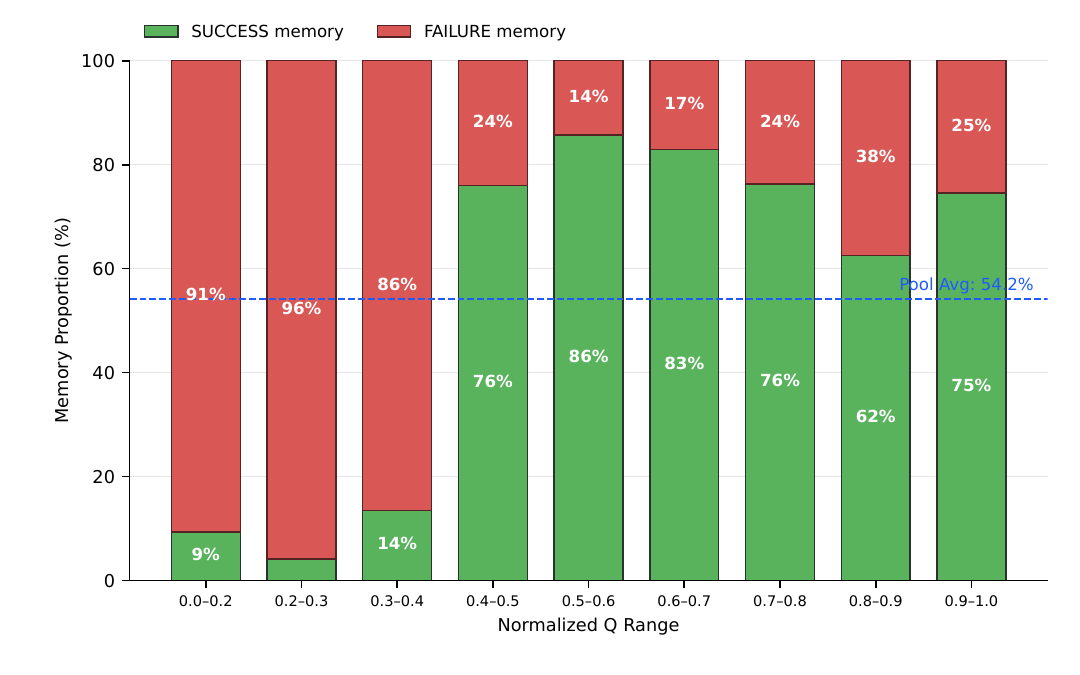}
  \caption{{MemRL checkpoint composition.}}
  \label{fig:automationbench_q_predictive_c}
\end{subfigure}
\caption{{Successful and failed memories by normalized Q range on AutomationBench. MemQ uses a 4,800-memory checkpoint; MemRL uses seed 43 at E9. Dashed lines indicate each checkpoint's overall successful-memory proportion, 60.5\% and 54.2\%, respectively.}}
\label{fig:automationbench_q_predictive}
\end{figure}

{In MemQ's 4,800-memory checkpoint, successful trajectories account for 60.5\% of the bank. Their proportions increase to 95.9\% and 95.2\% in the normalized Q ranges 0.7--0.8 and 0.8--0.9 (Figure~\ref{fig:automationbench_q_predictive_b}), leaving 4.1\% and 4.8\% failed trajectories in those ranges. High-Q ranges thus include memories from both successful and failed trajectories. The highest range, 0.9--1.0, contains only eight memories, all successful.}

{Figure~\ref{fig:automationbench_q_predictive_c} reports the corresponding composition of MemRL's seed-43 checkpoint at E9, whose overall successful-memory proportion is 54.2\%. The two checkpoint views show how trajectory outcomes are distributed across Q ranges in each memory bank.}

\section{\texorpdfstring{{Additional Ablation Studies}}{Additional Ablation Studies}}
\label{app:more_ablations}

\subsection{\texorpdfstring{{Q-Value Initialization}}{Q-Value Initialization}}
\label{app:initialization}

{Table~\ref{tab:init_ablation} compares parent-mean initialization with fixed values of 0.5 and 0 under the same within-benchmark ablation pipeline. Parent mean gives the best evaluation SR in both benchmarks. On LiveCodeBench, a fixed value of 0.5 instead gives higher final and cumulative SR. Parent-mean initialization therefore offers a useful cold-start rule, with an observed tradeoff between runtime and held-out performance. Results are point estimates from the initialization ablation runs.}

\begingroup
\begin{table}[H]
\centering
\caption{{Q-value initialization ablation results (\%).}}
\label{tab:init_ablation}
\small
\begin{tabular}{llccc}
\toprule
Benchmark & Q initialization & CSR & Final SR & Evaluation SR \\
\midrule
\multirow{3}{*}{LiveCodeBench}
 & Mean parent Q & \underline{67.62} & \underline{61.79} & \textbf{51.43} \\
 & 0.5 & \textbf{69.29} & \textbf{63.57} & \underline{45.71} \\
 & 0 & 57.86 & 52.86 & 34.29 \\
\midrule
\multirow{3}{*}{BFCL}
 & Mean parent Q & \textbf{86.43} & \textbf{75.04} & \textbf{62.83} \\
 & 0.5 & \underline{83.92} & \underline{70.60} & \underline{60.73} \\
 & 0 & 82.66 & 68.34 & 57.76 \\
\bottomrule
\end{tabular}
\end{table}
\endgroup

\subsection{\texorpdfstring{{Trace-Decay Sensitivity on LiveCodeBench}}{Trace-Decay Sensitivity on LiveCodeBench}}

\begin{figure}[H]
\centering
\includegraphics[width=0.32\linewidth]{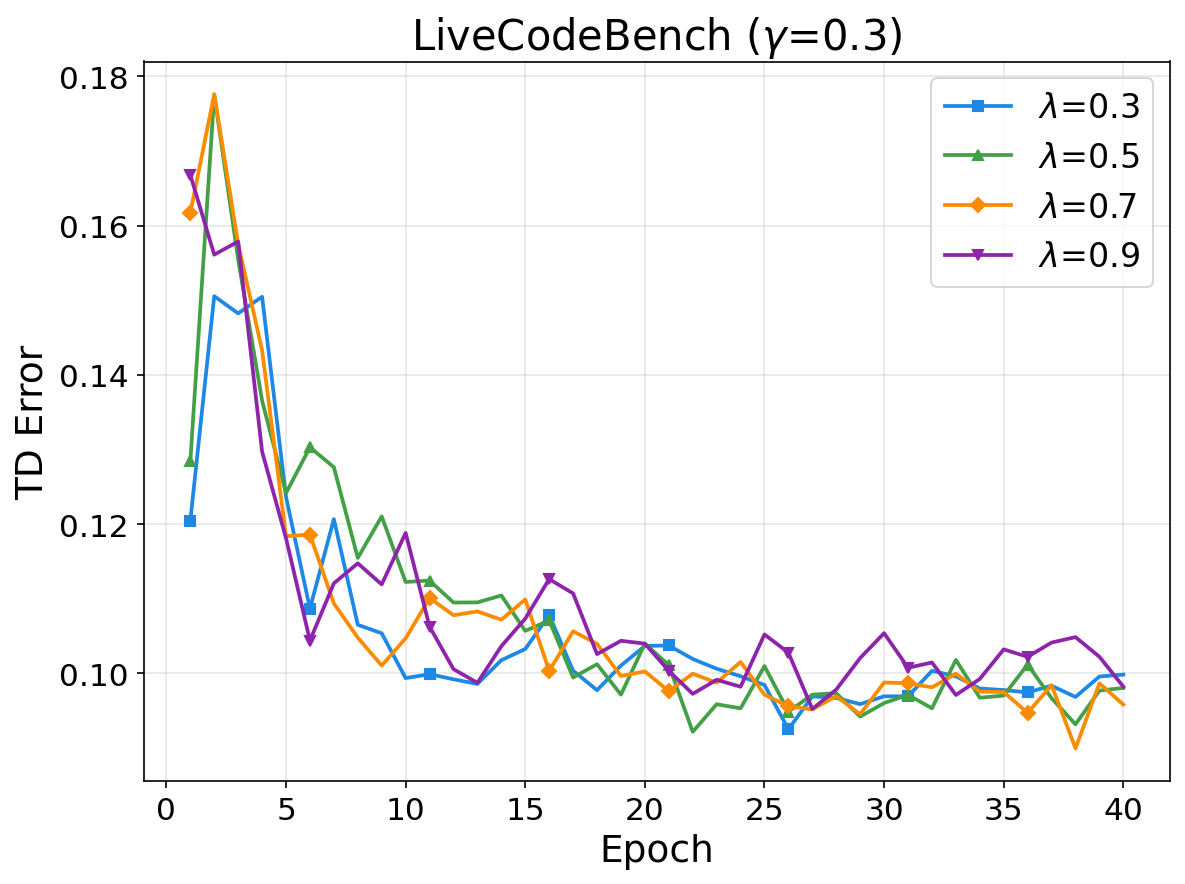}
\hfill
\includegraphics[width=0.32\linewidth]{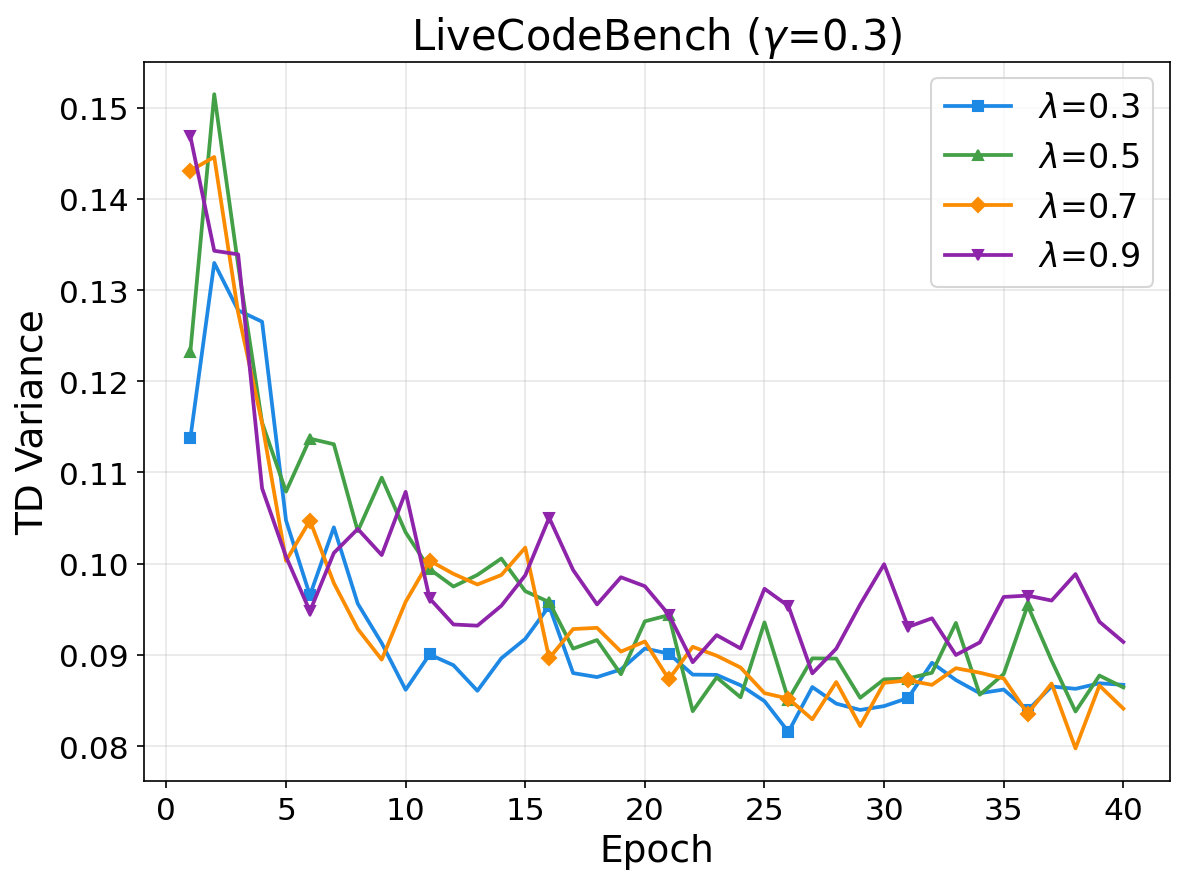}
\hfill
\includegraphics[width=0.32\linewidth]{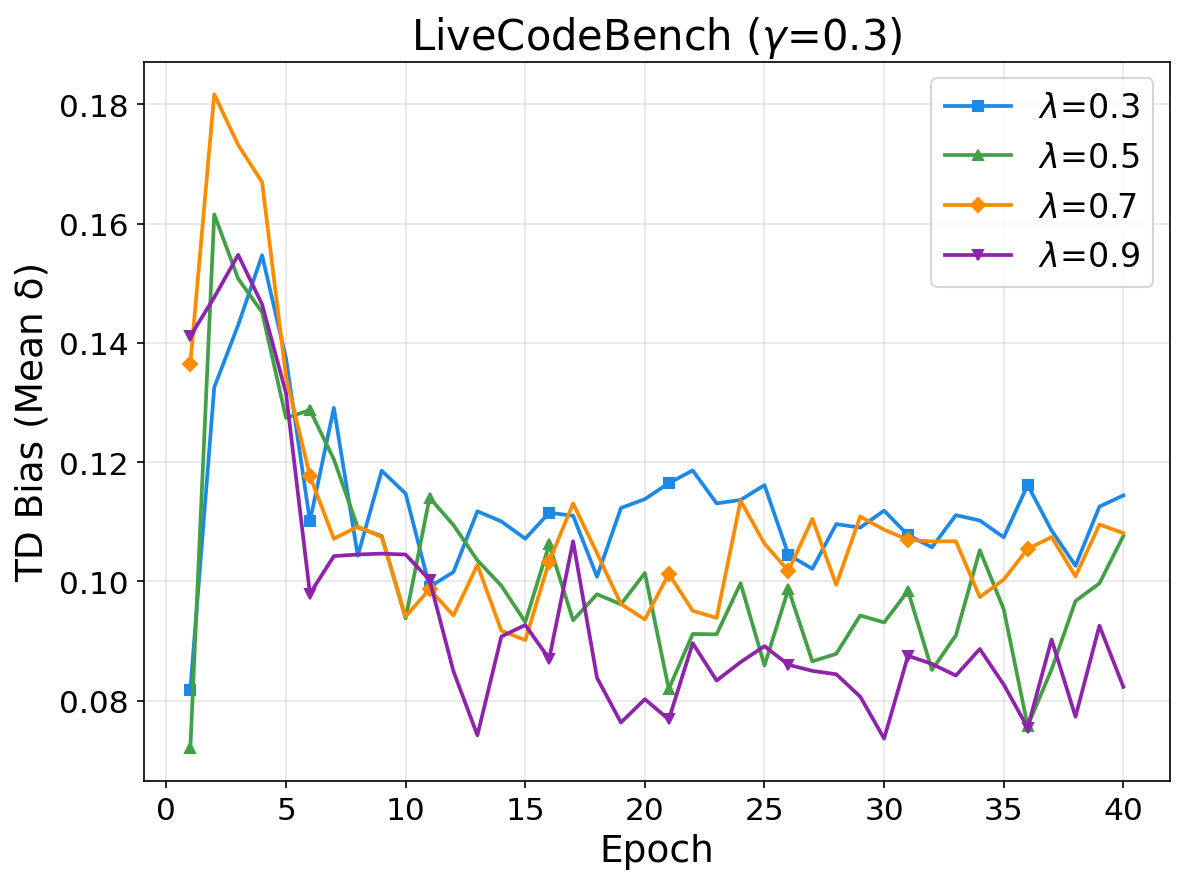}
\caption{{TD error, TD variance, and TD bias for the LiveCodeBench $\lambda$ sweep at $\gamma=0.3$.}}
\label{fig:lambda_sweep}
\end{figure}

{Figure~\ref{fig:lambda_sweep} reports TD diagnostics for the LiveCodeBench $\lambda$ sweep at $\gamma=0.3$. The corresponding Train SR curves appear in Figure~\ref{fig:gamma_lambda_sweeps} (bottom).}

\subsection{\texorpdfstring{{Additional $\gamma$ Sweeps and TD Diagnostics}}{Additional gamma Sweeps and TD Diagnostics}}
\label{app:gamma_td}

\begin{figure}[H]
\centering
\includegraphics[width=0.42\linewidth]{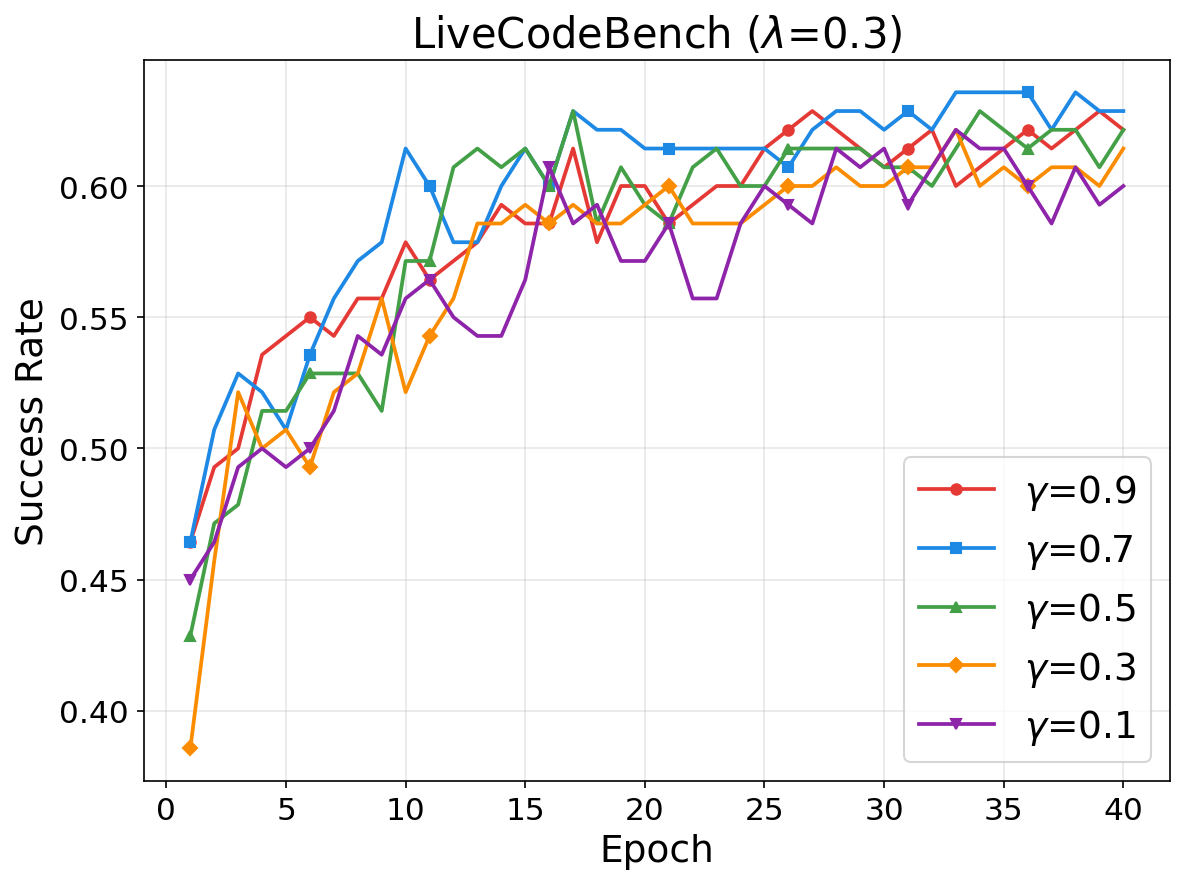}
\hspace{0.04\linewidth}
\includegraphics[width=0.42\linewidth]{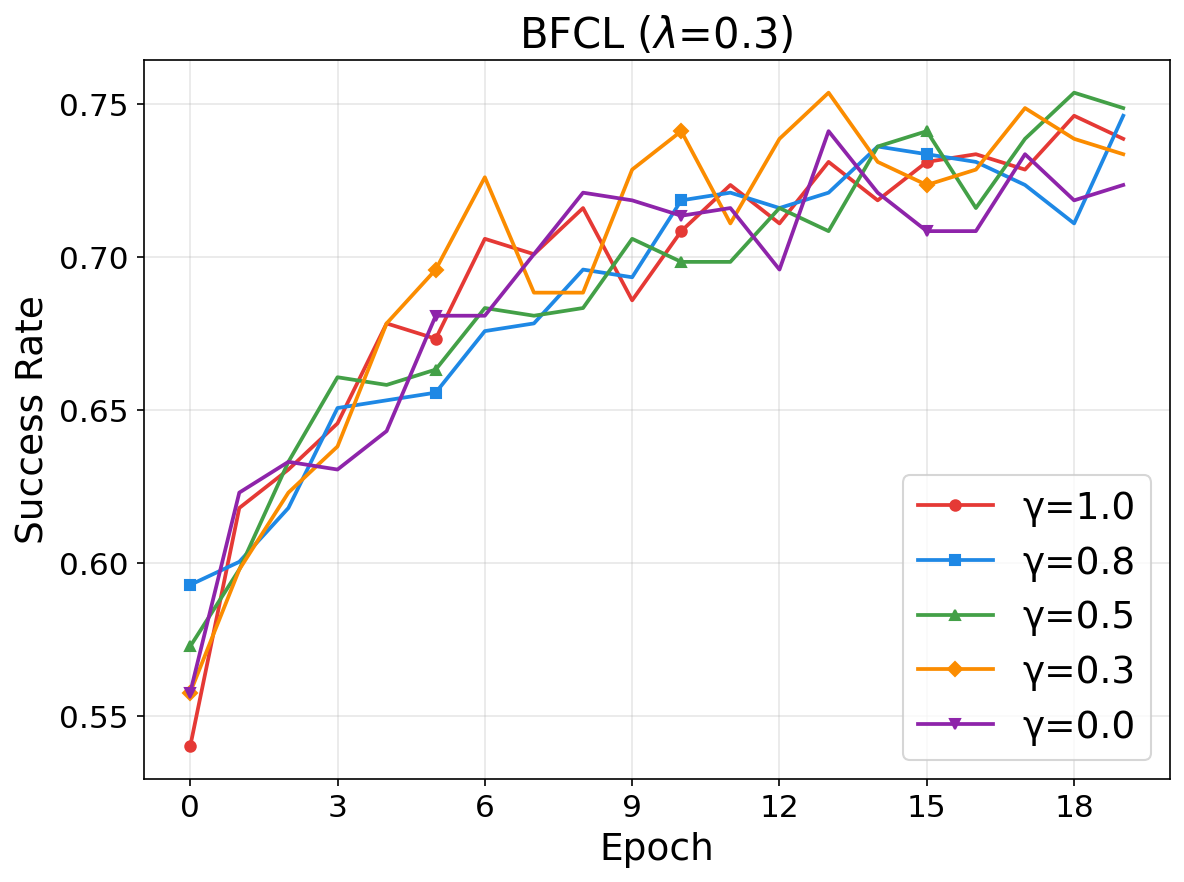}
\caption{{Additional Train SR curves under varying $\gamma$, with $\lambda=0.3$ on LiveCodeBench (left) and BFCL (right).}}
\label{fig:gamma_sweep_additional}
\end{figure}

{Figure~\ref{fig:gamma_sweep_additional} complements Figure~\ref{fig:gamma_lambda_sweeps} (top) with $\lambda=0.3$. LiveCodeBench no longer shows the marked late-stage decline at $\gamma=0.9$ seen with $\lambda=0.95$. On BFCL, intermediate discounts remain competitive with the largest tested discount. These additional sweeps show how the response to $\gamma$ changes with trace decay.}

\begin{figure}[H]
\centering
\includegraphics[width=0.24\linewidth]{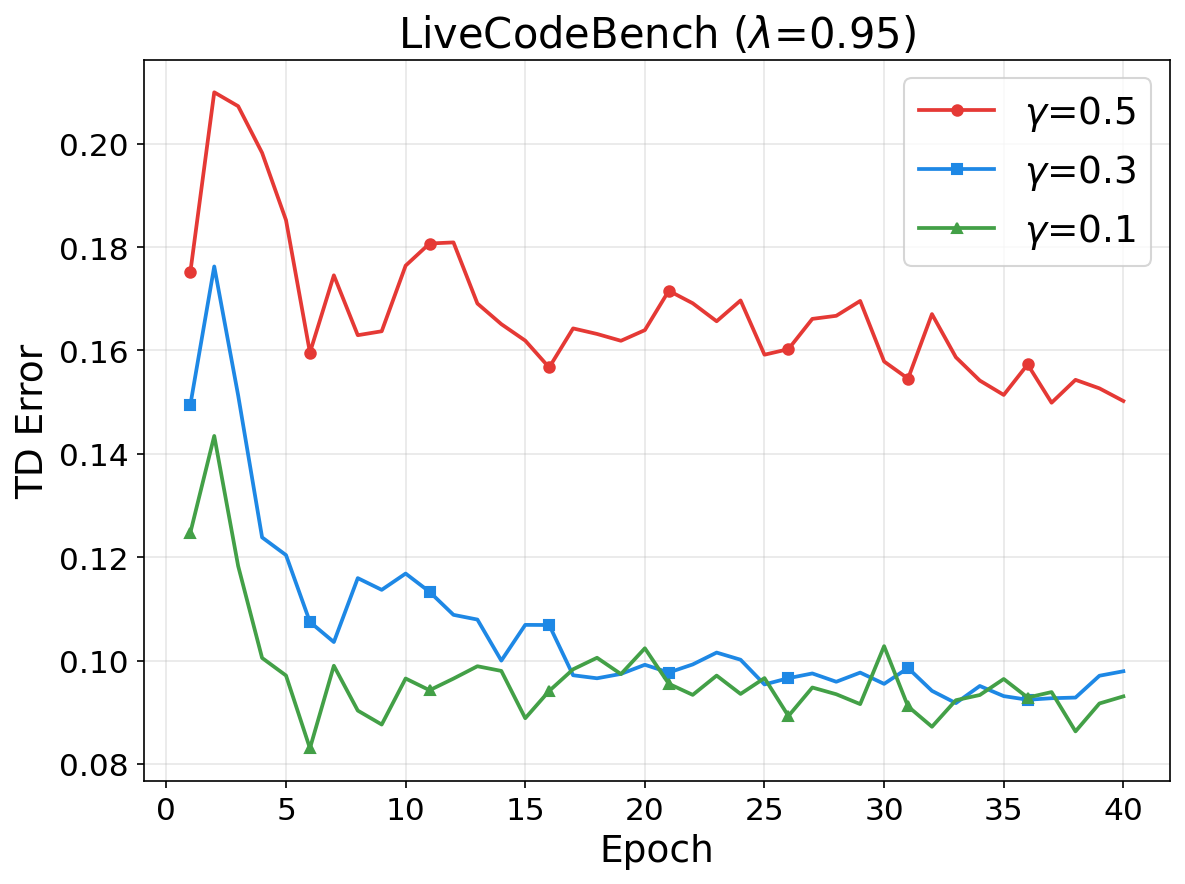}
\hfill
\includegraphics[width=0.24\linewidth]{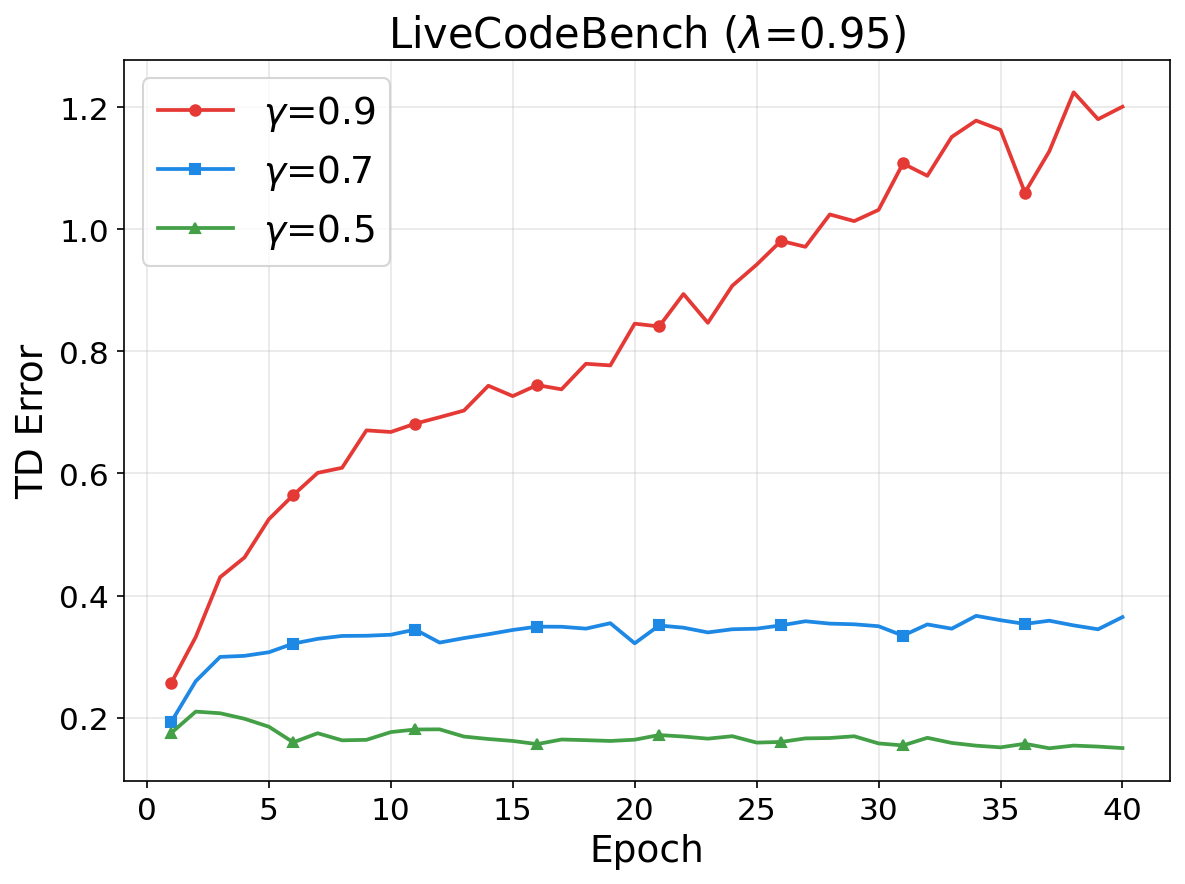}
\hfill
\includegraphics[width=0.24\linewidth]{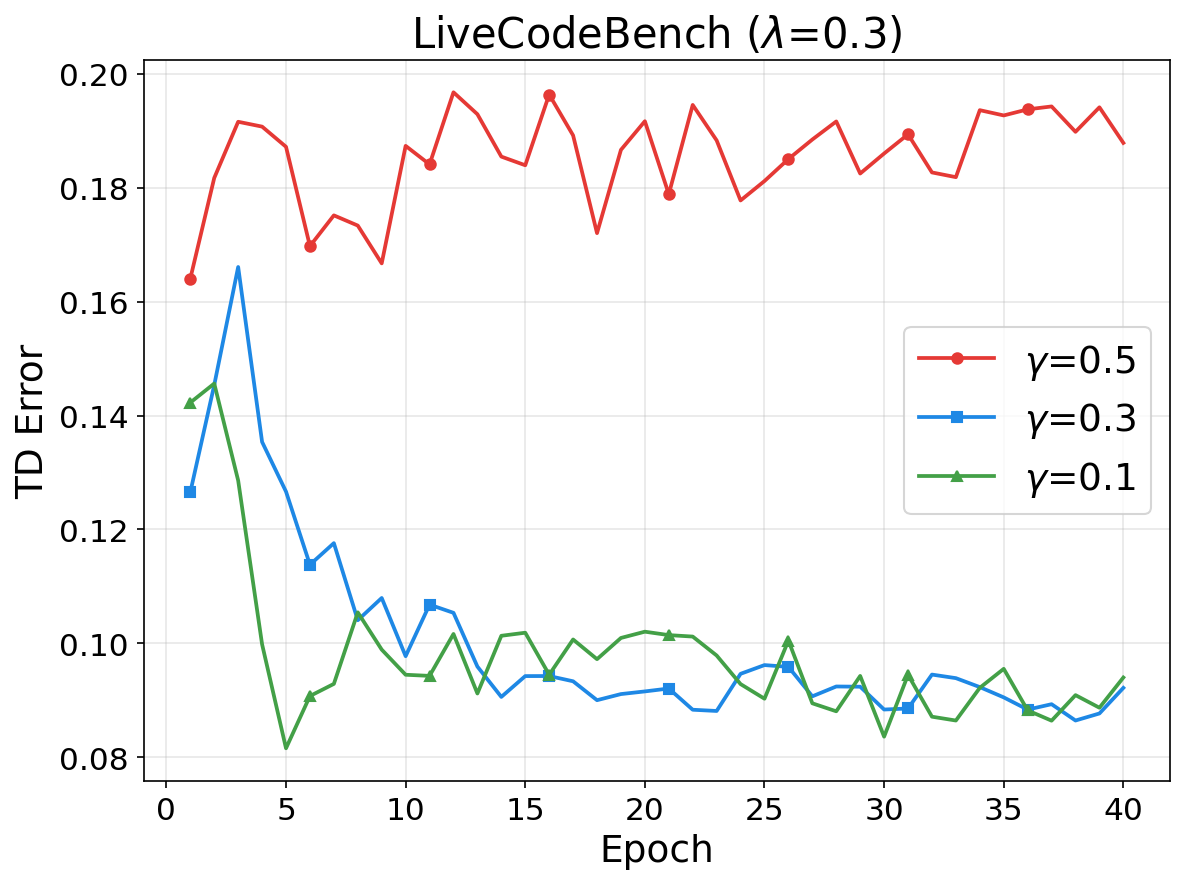}
\hfill
\includegraphics[width=0.24\linewidth]{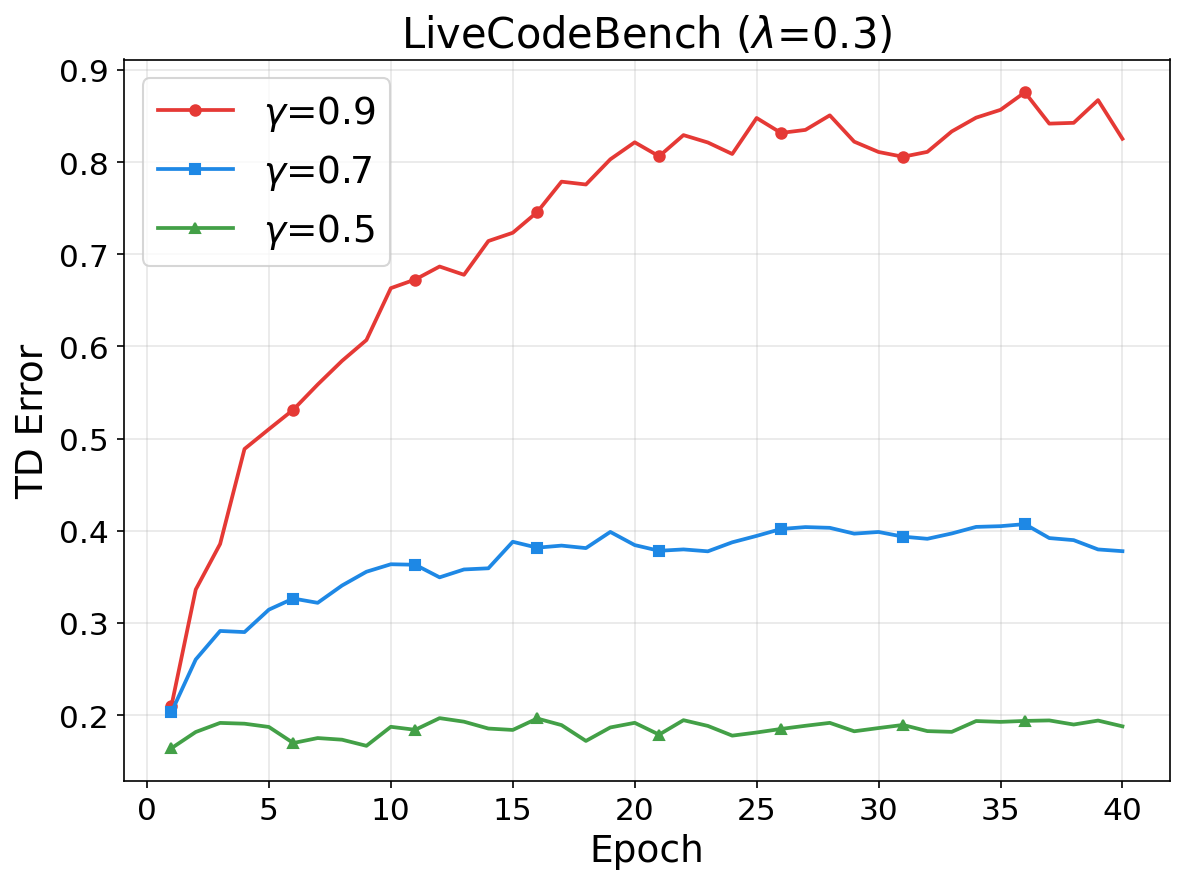}
\\[5pt]
\includegraphics[width=0.24\linewidth]{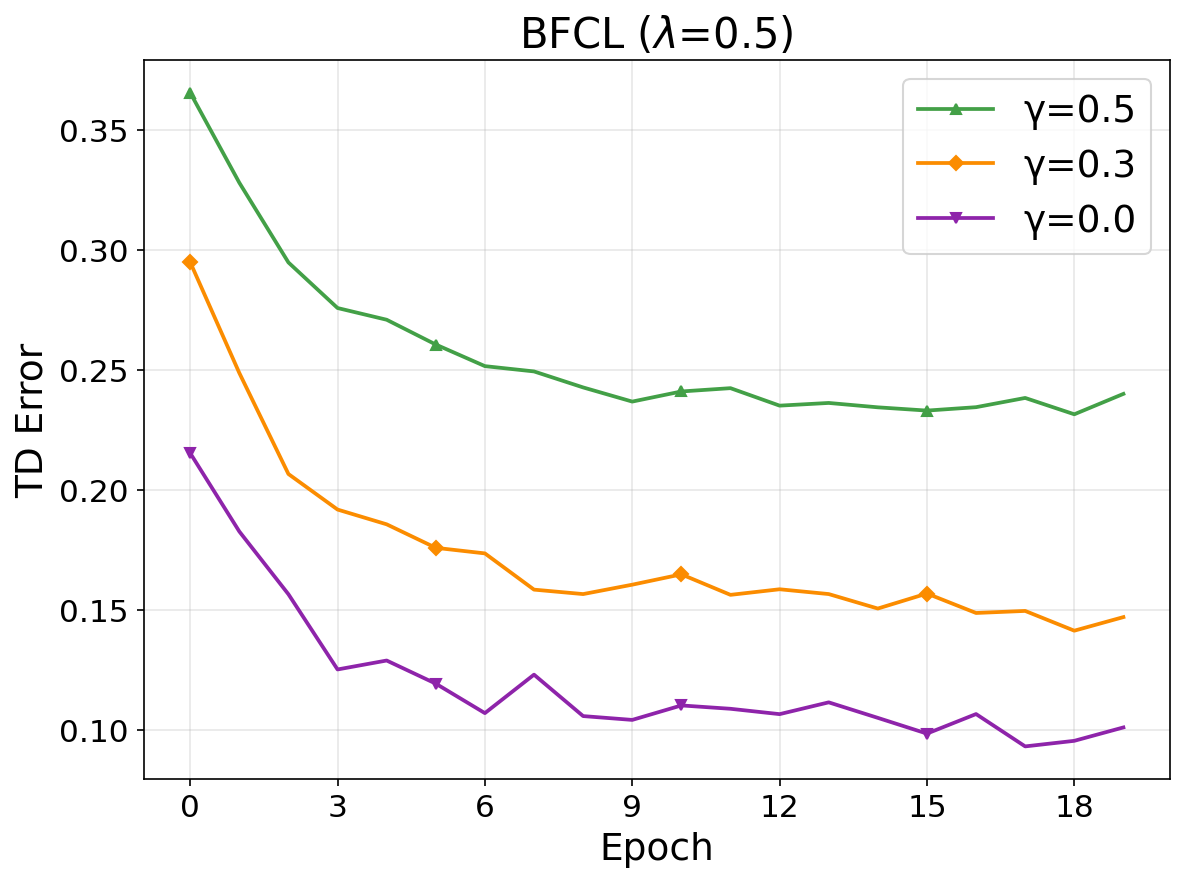}
\hfill
\includegraphics[width=0.24\linewidth]{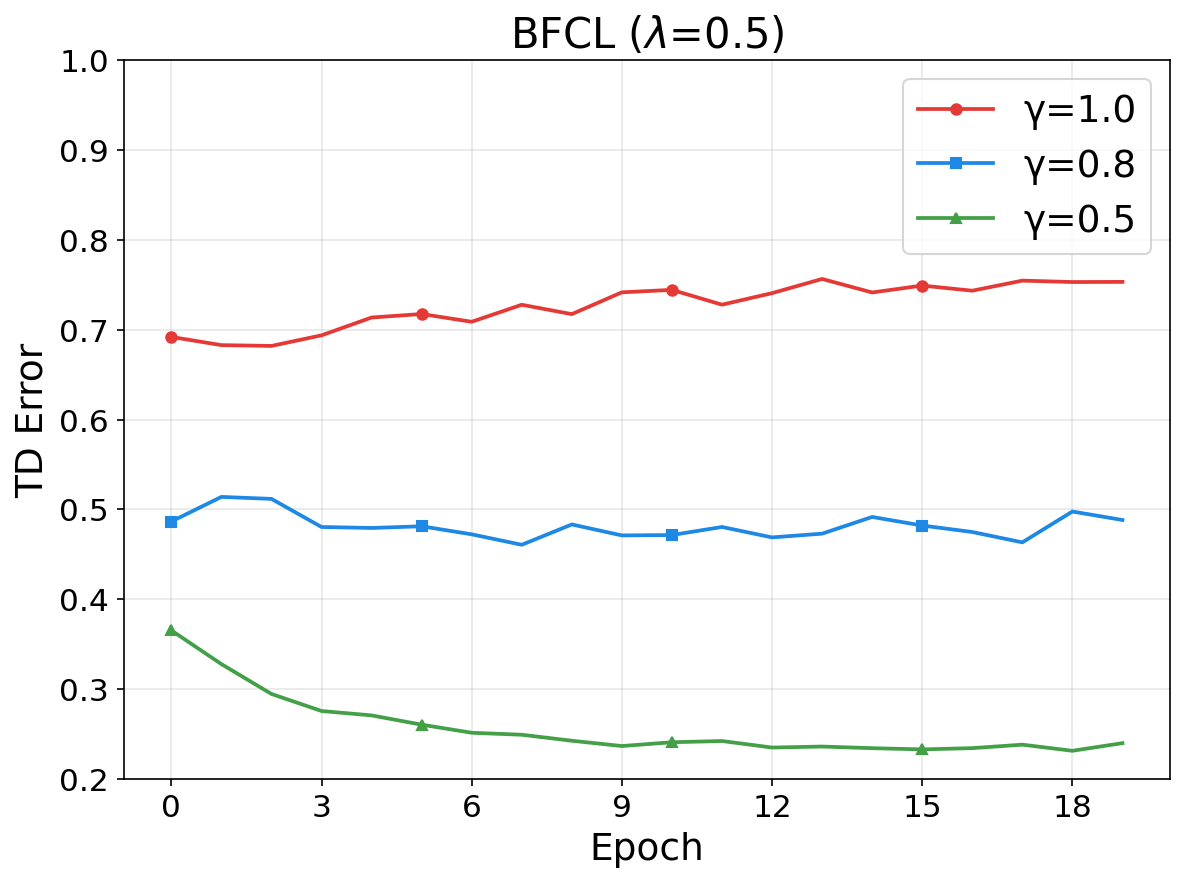}
\hfill
\includegraphics[width=0.24\linewidth]{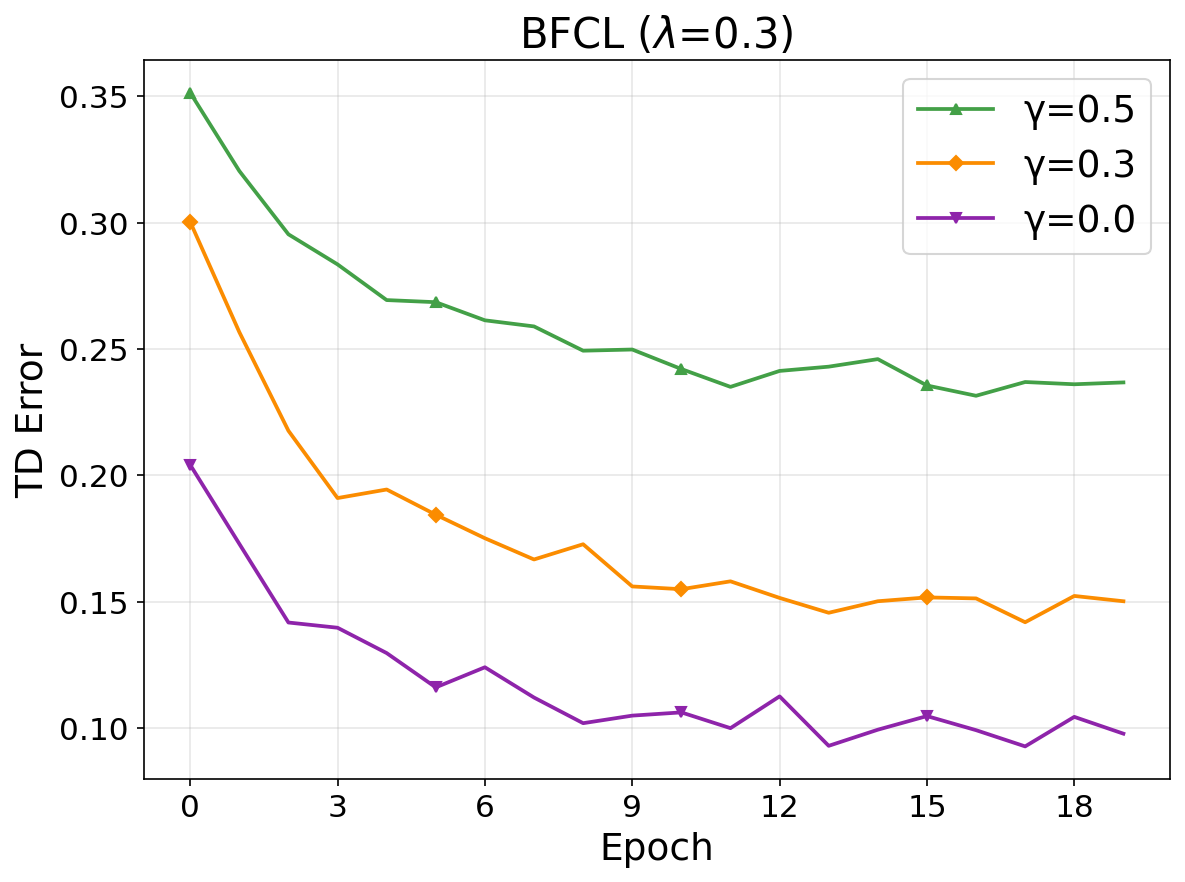}
\hfill
\includegraphics[width=0.24\linewidth]{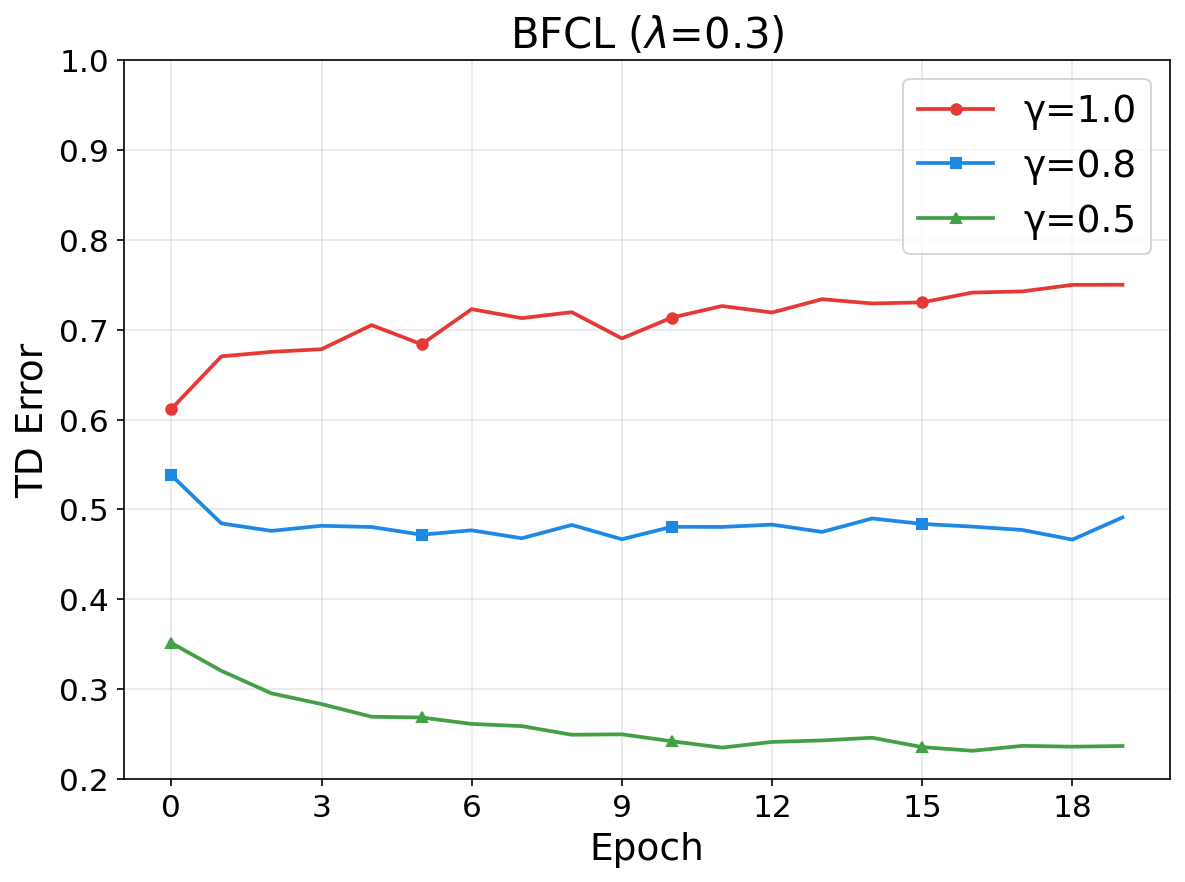}
\caption{{TD error under different $\gamma$ on LiveCodeBench (top) and BFCL (bottom).}}
\label{fig:gamma_td_lcb}
\label{fig:gamma_td_bfcl}
\end{figure}

\subsection{$\lambda$ Ablation on BFCL}
\label{app:lambda_sweep_bfcl}

\begin{figure}[H]
\centering
\includegraphics[width=0.32\linewidth]{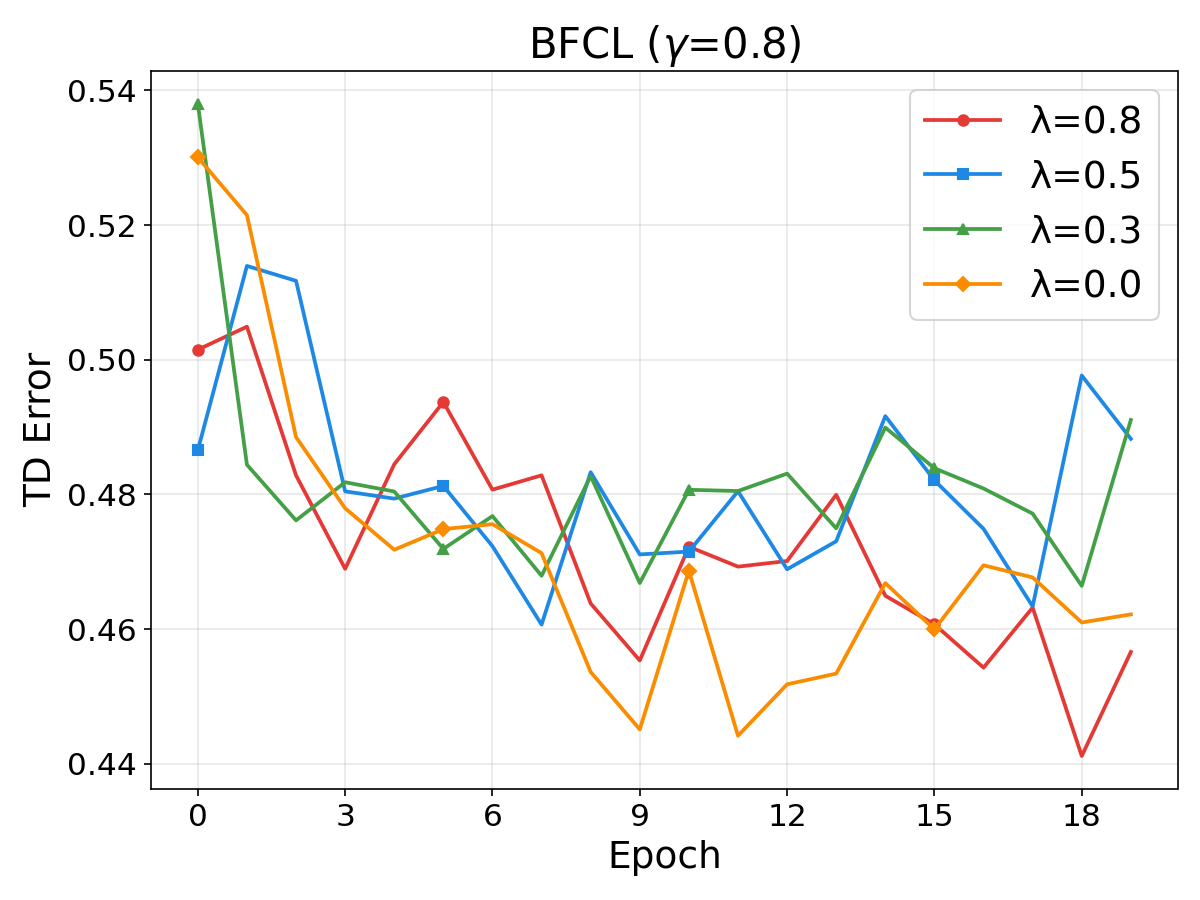}
\hfill
\includegraphics[width=0.32\linewidth]{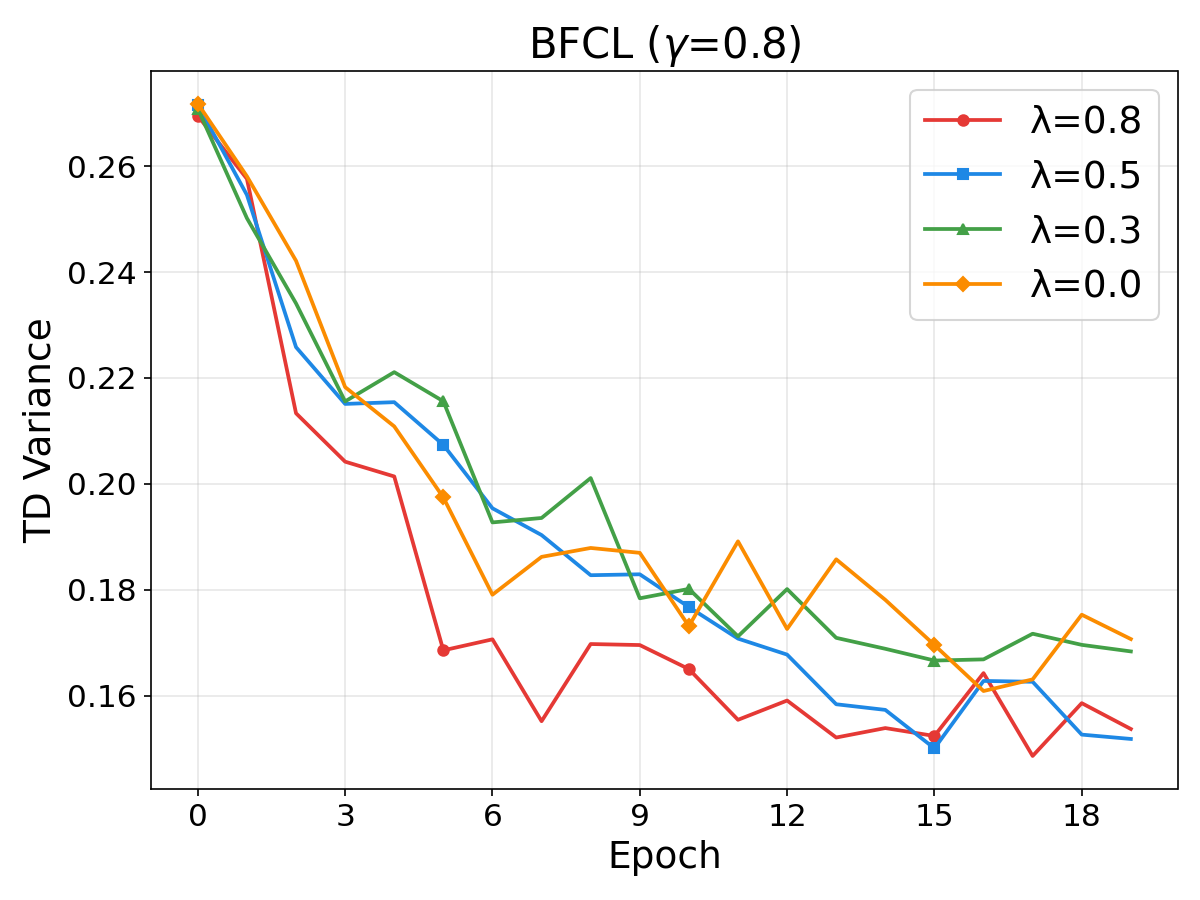}
\hfill
\includegraphics[width=0.32\linewidth]{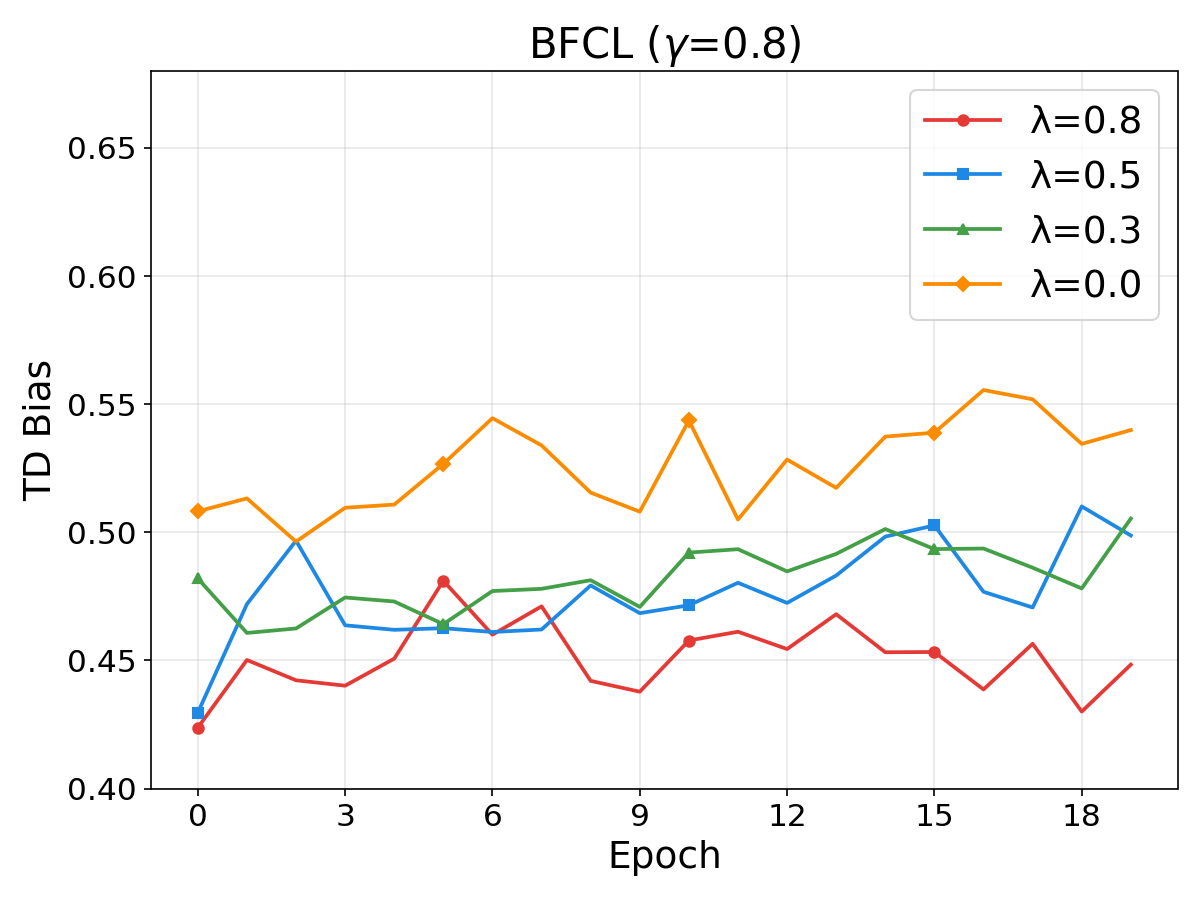}
\caption{{TD error, TD variance, and TD bias for the BFCL $\lambda$ sweep at $\gamma=0.8$. The corresponding Train SR curves appear in Figure~\ref{fig:gamma_lambda_sweeps} (bottom).}}
\label{fig:lambda_sweep_bfcl}
\end{figure}

\subsection{$D$ Ablation on BFCL}
\label{app:Depth_Ablations}

\begin{figure}[H]
\centering
\includegraphics[width=0.24\linewidth]{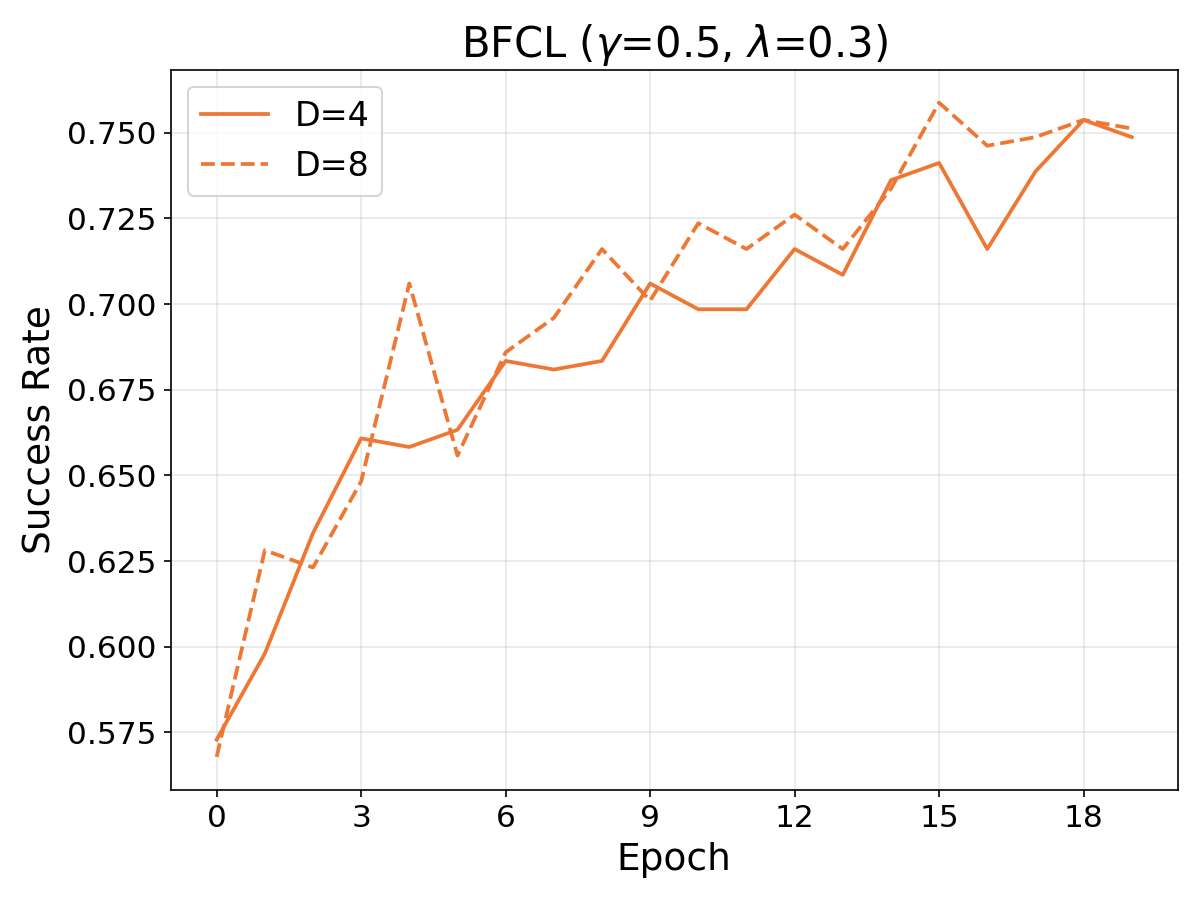}\hfill
\includegraphics[width=0.24\linewidth]{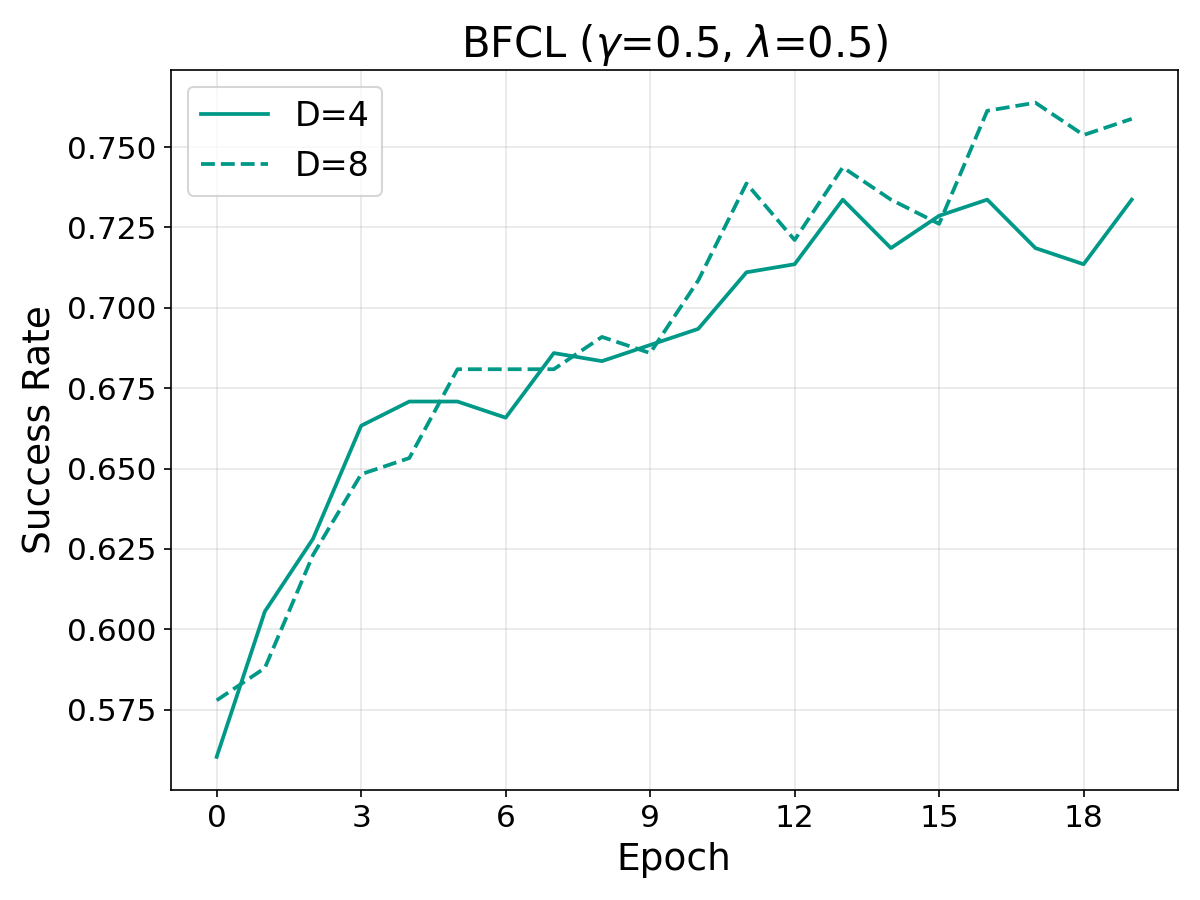}\hfill
\includegraphics[width=0.24\linewidth]{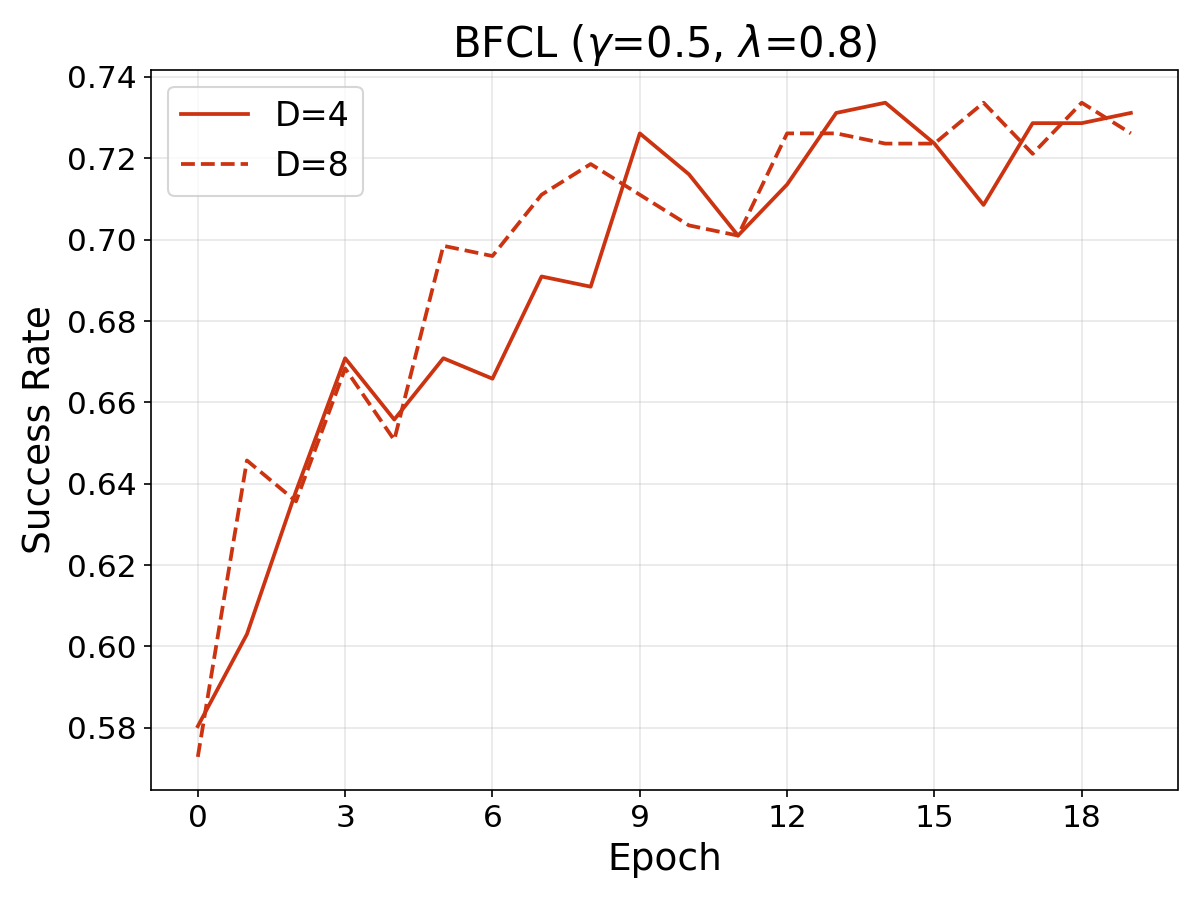}\hfill
\includegraphics[width=0.24\linewidth]{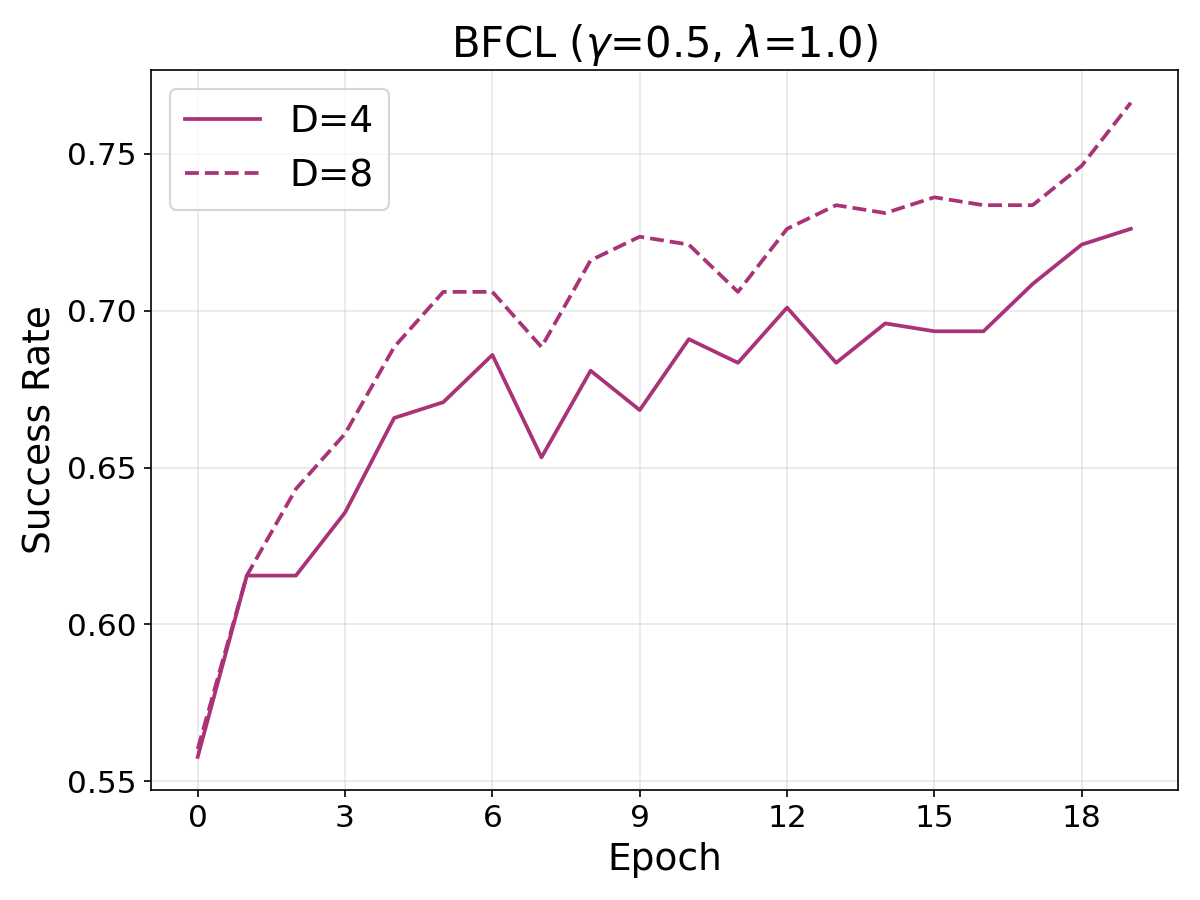}

\vspace{1em}

\includegraphics[width=0.24\linewidth]{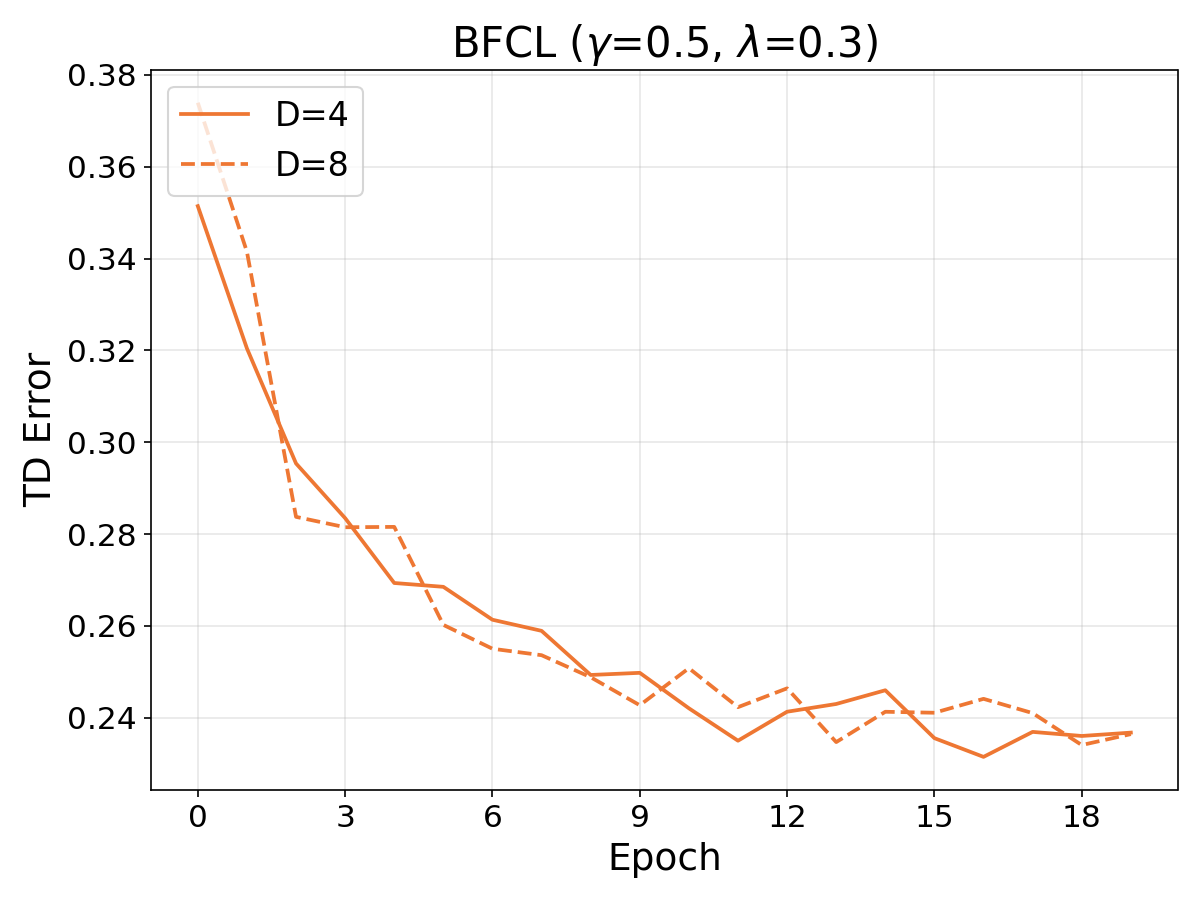}\hfill
\includegraphics[width=0.24\linewidth]{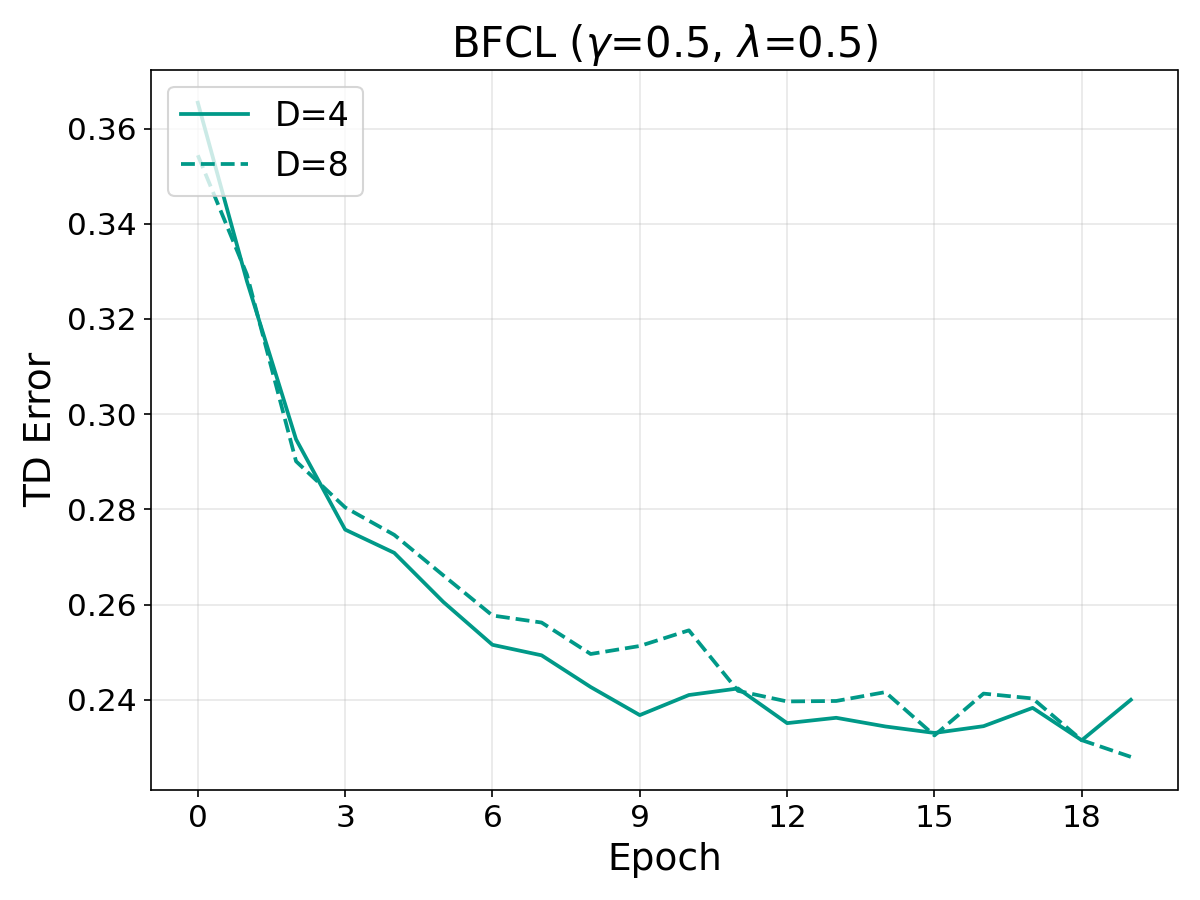}\hfill
\includegraphics[width=0.24\linewidth]{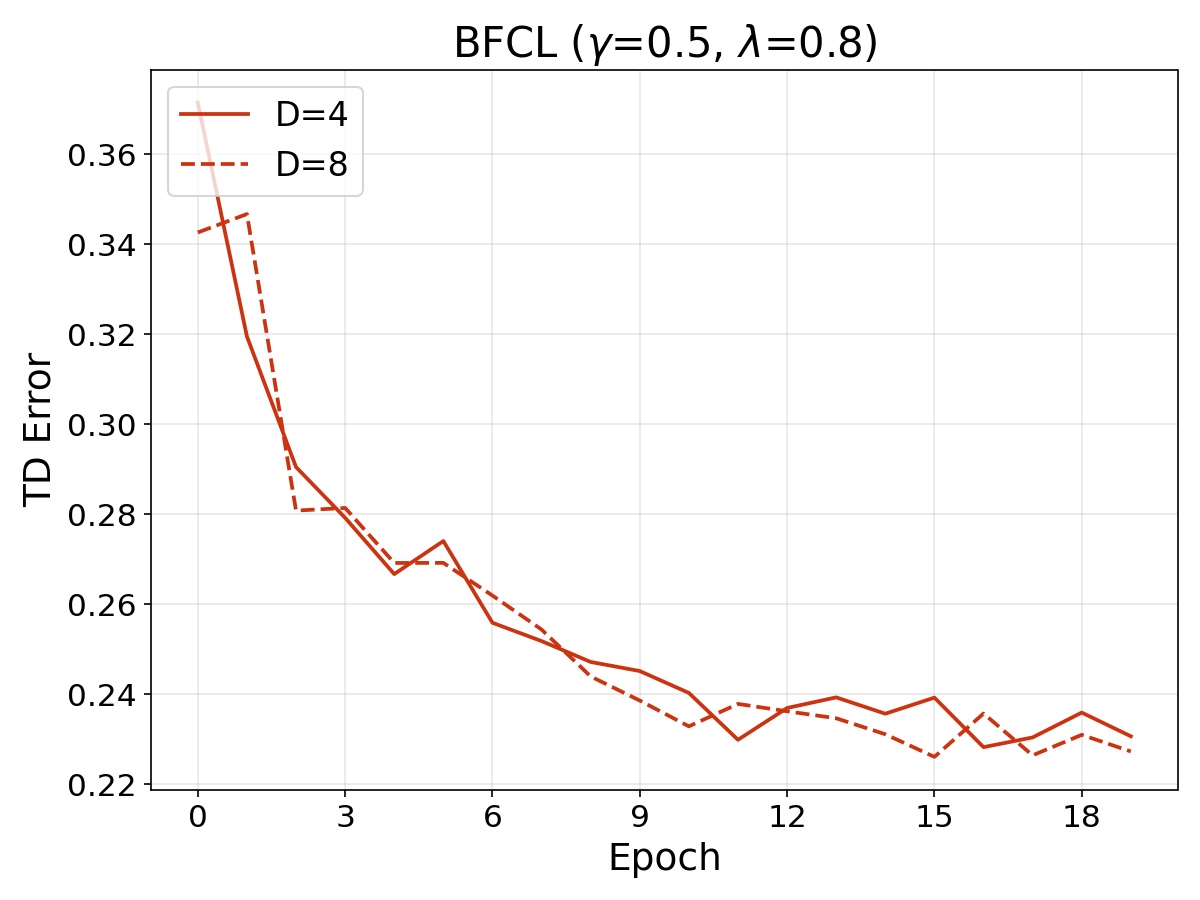}\hfill
\includegraphics[width=0.24\linewidth]{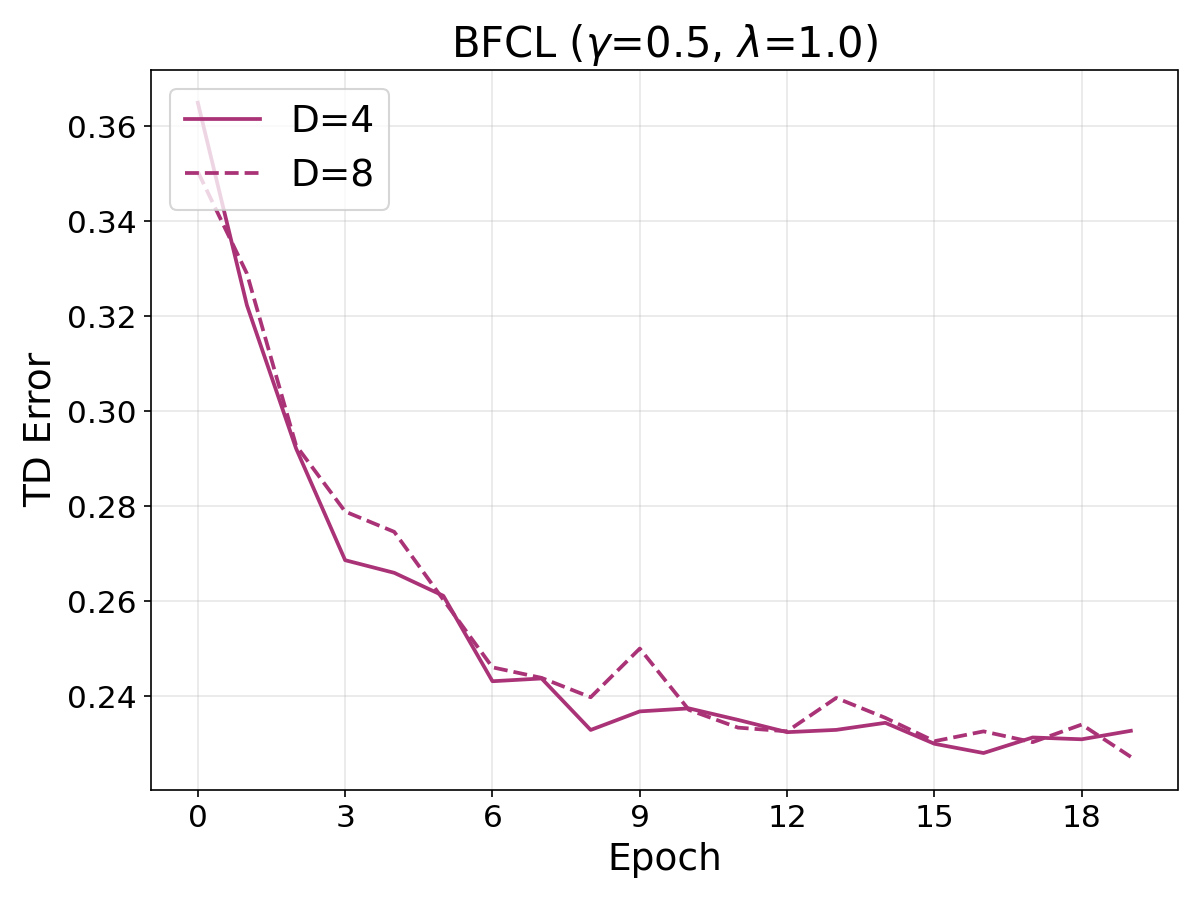}

\caption{SR (top row) and TD error (bottom row) under different $D$ on BFCL.}
\label{fig:depth_combined}
\end{figure}

\subsection{\texorpdfstring{{Similarity--Q Retrieval Weight Ablations}}{Similarity--Q Retrieval Weight Ablations}}
\label{app:retrieval_weight_ablation}

{Tables~\ref{tab:retrieval_weights_lcb} and \ref{tab:retrieval_weights_bfcl} report the complete sweep used to test whether semantic similarity or learned Q-values alone explain MemQ's retrieval behavior. The best weight depends on the benchmark and metric, but mixed policies consistently provide stronger overall tradeoffs than either endpoint. These ablations were conducted separately from the main experiments. The best-performing weights identified here were not adopted in the main experiments, which use the configurations reported in Appendix~\ref{app:hyperparameters}.}

\begingroup
\begin{table}[H]

\centering
\begin{minipage}[t]{0.49\linewidth}
\vspace{0pt}
\centering
\captionsetup{font=footnotesize,skip=5pt}
\caption{Similarity/Q retrieval-weight ablation on LiveCodeBench (\%).}
\label{tab:retrieval_weights_lcb}
\footnotesize
\setlength{\tabcolsep}{3pt}
\begin{tabular}{@{}ccccc@{}}
\toprule
$w_s$ & $w_q$ & CSR & Final SR & Evaluation SR \\
\midrule
0.0 & 1.0 & 64.29 & 52.86 & 45.71 \\
0.2 & 0.8 & 67.14 & 58.57 & 42.86 \\
0.4 & 0.6 & 65.71 & 58.57 & 48.57 \\
0.6 & 0.4 & \textbf{67.62} & \textbf{61.79} & \textbf{51.43} \\
0.8 & 0.2 & 62.14 & 60.00 & 48.57 \\
1.0 & 0.0 & 58.57 & 55.00 & 42.86 \\
\bottomrule
\end{tabular}
\end{minipage}
\hfill
\begin{minipage}[t]{0.49\linewidth}
\vspace{0pt}
\centering
\captionsetup{font=footnotesize,skip=5pt}
\caption{Similarity/Q retrieval-weight ablation on BFCL (\%). Metric-wise maxima are highlighted separately.}
\label{tab:retrieval_weights_bfcl}
\footnotesize
\setlength{\tabcolsep}{3pt}
\begin{tabular}{@{}ccccc@{}}
\toprule
$w_s$ & $w_q$ & CSR & Final SR & Evaluation SR \\
\midrule
0.0 & 1.0 & 85.43 & 61.56 & 59.41 \\
0.2 & 0.8 & 86.18 & 67.17 & \textbf{63.70} \\
0.4 & 0.6 & \textbf{86.93} & \textbf{73.60} & 61.39 \\
0.6 & 0.4 & 86.68 & 71.77 & 60.07 \\
0.8 & 0.2 & 83.17 & 68.34 & 63.04 \\
1.0 & 0.0 & 78.64 & 65.75 & 56.77 \\
\bottomrule
\end{tabular}
\end{minipage}
\end{table}
\endgroup

\section{Baseline Implementation Details}
\label{app:baseline_implementations}

RAG: We use the official MemRL codebase for the RAG experiments because it has already integrated RAG by disabling the value-related hyperparameters and doing raw trajectory saving.

{Self-RAG: We initially attempted to reproduce the original method using its released Llama-2 checkpoints, but they did not function reliably in our agentic evaluations. These checkpoints are fine-tuned on instruction-following and knowledge-intensive data augmented with retrieved passages and reflection tokens~\citep{asai2024selfrag}, a training setting that differs from the interactive tasks considered here. We therefore implement its sequence of on-demand retrieval, relevance, support, utility critique, and critique-based selection with the shared frozen backbone, eliciting these decisions through prompting.}

Mem0: We use the official mem0 repository for its experiments. All the hyperparameters shared with other baselines align with those baselines.

MemP: We use the official MemRL codebase for the MemP experiments because it has already integrated MemP.

MemRL: We use the official MemRL codebase for the MemRL experiments.

\section{Hyperparameter Configurations}
\label{app:hyperparameters}

We report hyperparameters for each benchmark separately, since key parameters ($\gamma$, $\lambda$, $w_s$, $w_q$, Batch size, $\theta_\mathrm{sim}$, Initial Q, epochs) vary across benchmarks. Table~\ref{tab:hp_shared} lists settings shared across all benchmarks. Tables~\ref{tab:hp_llab}--\ref{tab:hp_automationbench} give per-benchmark configurations for each method.

\begin{table}[H]
\centering
\caption{Settings shared across all benchmarks and methods.}
\label{tab:hp_shared}
\begin{tabular}{lll}
\toprule
\textbf{Parameter} & \textbf{Value} & \textbf{Description} \\
\midrule
normalize\_dq & true & Normalize Q-value deltas \\
success\_reward & 1.0 & Reward for correct answer \\
failure\_reward & 0.0 & Reward for wrong answer \\
normalize\_sim & false & Raw similarity scores (already in $[0,1]$) \\
normalize\_q & true (value-based methods) & Normalize Q for retrieval scoring \\
temperature & 0.0 & Deterministic LLM generation \\
$D$ & 4 & {Maximum backward-path depth for DAG traversal} \\
Discount floor & $(\gamma\lambda)^d < 10^{-12}$ & {Prune deep provenance paths} \\
\bottomrule
\end{tabular}
\end{table}

\paragraph{LifeLongAgentBench.}
{LLM backbone: GPT-4o-mini. Embedding model: \texttt{text-embedding-3-large} ($d=3072$). The artifact uses 350 training instances and 150 held-out evaluation instances; every method uses this split, and hyperparameters are not tuned on the held-out set. Max agent steps per task: 15.}

\begin{table}[H]
\centering
\caption{LifeLongAgentBench hyperparameters.}
\label{tab:hp_llab}
\setlength{\tabcolsep}{3.5pt}
\small
\begin{tabular}{l ccccccc}
\toprule
 & \textbf{MemQ} & \textbf{MemRL} & \textbf{MemP} & \textbf{Mem0} & {\textbf{Self-RAG}} & \textbf{RAG} \\
\midrule
$\alpha$ & 0.3 & 0.3 & N/A & N/A & N/A & N/A \\
$\gamma$ & 0.5 & N/A & N/A & N/A & N/A & N/A \\
$\lambda$ & 0.7 & N/A & N/A & N/A & N/A & N/A \\
$\varepsilon$ & 0.01 & 0.01 & 0.01 & 0.0 & 0.01 & 0.01 \\
$k_\mathrm{ret}$ & 10 & 10 & 10 & 5 & 10 & 10 \\
$k_\mathrm{top}$ & 5 & 5 & 5 & 5 & 5 & 5 \\
$\theta_\mathrm{sim}$ & 0.5 & 0.5 & 0.5 & 0.0 & 0.5 & 0.5 \\
$w_s$ & 0.5 & 0.5 & N/A & N/A & N/A & N/A \\
$w_q$ & 0.5 & 0.5 & N/A & N/A & N/A & N/A \\
Batch size & 64 & 64 & 64 & 64 & 64 & 64 \\
Max tokens & 10240 & 10240 & 10240 & 10240 & 10240 & 10240 \\
Initial Q & 0.66 & 0.66 & N/A & N/A & N/A & N/A \\
Build & proceduralization & proceduralization & proceduralization &trajectory  & trajectory & trajectory \\
Update & adjustment & adjustment & adjustment & vanilla& vanilla & vanilla \\
Epochs & 20 & 20 & 20 & 20 & 20 & 20 \\
\bottomrule
\end{tabular}
\end{table}

\paragraph{LiveCodeBench.}
LLM backbone: Gemma-4-E4B-it. Embedding model: Qwen3-Embedding-8B. 175 problems (140 train / 35 evaluation).

\begin{table}[H]
\centering
\caption{LiveCodeBench hyperparameters.}
\label{tab:hp_lcb}
\setlength{\tabcolsep}{3.5pt}
\small
\begin{tabular}{l ccccccc}
\toprule
 & \textbf{MemQ} & \textbf{MemRL} & \textbf{MemP} & \textbf{Mem0} & {\textbf{Self-RAG}} & \textbf{RAG} \\
\midrule
$\alpha$ & 0.3 & 0.3 & N/A & N/A & N/A & N/A \\
$\gamma$ & 0.3 & N/A & N/A & N/A & N/A & N/A \\
$\lambda$ & 0.95 & N/A & N/A & N/A & N/A & N/A \\
$\varepsilon$ & 0.0 & 0.0 & 0.0 & 0.0 & 0.0 & 0.0 \\
$k_\mathrm{ret}$ & 5 & 5 & 5 & 5 & 5 & 5 \\
$k_\mathrm{top}$ & 3 & 3 & 3 & 5 & 3 & 3 \\
$\theta_\mathrm{sim}$ & 0.6 & 0.6 & 0.6 & 0.0 & 0.6 & 0.6 \\
$w_s$ & 0.4 & 0.4 & N/A & N/A & N/A & N/A \\
$w_q$ & 0.6 & 0.6 & N/A & N/A & N/A & N/A \\
Batch size & 32 & 32 & 32 & 32 & 32 & 32 \\
Max tokens & 10240 & 10240 & 10240 & 10240 & 10240 & 10240 \\
Initial Q & 0.37 & 0.37 & N/A & N/A & N/A & N/A \\
Build & proceduralization & proceduralization & proceduralization &trajectory  & trajectory & trajectory \\
Update & adjustment & adjustment & adjustment & vanilla& vanilla & vanilla \\
Epochs & 40 & 40 & 40 & 40 & 40 & 40 \\
\bottomrule
\end{tabular}
\end{table}

\paragraph{MMMU Pro.}
LLM backbone: Gemma-4-E4B-it. Embedding model: Qwen3-Embedding-8B. 1730 problems (1384 train / 346 valid).

\begin{table}[H]
\centering
\caption{MMMU Pro hyperparameters.}
\label{tab:hp_mmmu}
\setlength{\tabcolsep}{3.5pt}
\small
\begin{tabular}{l ccccccc}
\toprule
 & \textbf{MemQ} & \textbf{MemRL} & \textbf{MemP} & \textbf{Mem0} & {\textbf{Self-RAG}} & \textbf{RAG} \\
\midrule
$\alpha$ & 0.3 & 0.3 & N/A & N/A & N/A & N/A \\
$\gamma$ & 0.5 & N/A & N/A & N/A & N/A & N/A \\
$\lambda$ & 0.7 & N/A & N/A & N/A & N/A & N/A \\
$\varepsilon$ & 0.0 & 0.0 & 0.0 & 0.0 & 0.0 & 0.0 \\
$k_\mathrm{ret}$ & 5 & 5 & 5 & 5 & 5 & 5 \\
$k_\mathrm{top}$ & 3 & 3 & 3 & 5 & 3 & 3 \\
$\theta_\mathrm{sim}$ & 0.55 & 0.55 & 0.55 & 0.0 & 0.50 & 0.55 \\
$w_s$ & 0.5 & 0.5 & N/A & N/A & N/A & N/A \\
$w_q$ & 0.5 & 0.5 & N/A & N/A & N/A & N/A \\
Batch size & 128 & 128 & 128 & 128 & 128 & 128 \\
Max tokens & 4096 & 4096 & 4096 & 4096 & 4096 & 4096 \\
Initial Q & 0.526 & 0.526 & N/A & N/A & N/A & N/A \\
Build & proceduralization & proceduralization & proceduralization &trajectory  & trajectory & trajectory \\
Update & adjustment & adjustment & adjustment & vanilla& vanilla & vanilla \\
Epochs & 50 & 50 & 50 & 50 & 50 & 50 \\
\bottomrule
\end{tabular}
\end{table}

\paragraph{ERQA.}
LLM backbone: Gemma-4-E4B-it. Embedding model: Qwen3-Embedding-8B. 400 problems (320 train / 80 valid).

\begin{table}[H]
\centering
\caption{ERQA hyperparameters.}
\label{tab:hp_erqa}
\setlength{\tabcolsep}{3.5pt}
\small
\begin{tabular}{l ccccccc}
\toprule
 & \textbf{MemQ} & \textbf{MemRL} & \textbf{MemP} & \textbf{Mem0} & {\textbf{Self-RAG}} & \textbf{RAG} \\
\midrule
$\alpha$ & 0.3 & 0.3 & N/A & N/A & N/A & N/A \\
$\gamma$ & 0.5 & N/A & N/A & N/A & N/A & N/A \\
$\lambda$ & 0.99 & N/A & N/A & N/A & N/A & N/A \\
$\varepsilon$ & 0.0 & 0.0 & 0.0 & 0.0 & 0.0 & 0.0 \\
$k_\mathrm{ret}$ & 5 & 5 & 5 & 5 & 5 & 5 \\
$k_\mathrm{top}$ & 3 & 3 & 3 & 5 & 3 & 3 \\
$\theta_\mathrm{sim}$ & 0.6 & 0.6 & 0.6 & 0.0 & 0.5 & 0.6 \\
$w_s$ & 0.4 & 0.5 & N/A & N/A & N/A & N/A \\
$w_q$ & 0.6 & 0.5 & N/A & N/A & N/A & N/A \\
Batch size & 64 & 64 & 64 & 64 & 64 & 64 \\
Max tokens & 4096 & 4096 & 4096 & 4096 & 4096 & 4096 \\
Initial Q & 0.37 & 0.37 & N/A & N/A & N/A & N/A \\
Build & proceduralization & proceduralization & proceduralization &trajectory  & trajectory & trajectory \\
Update & adjustment & adjustment & adjustment & vanilla& vanilla & vanilla \\
Epochs & 100 & 100 & 100 & 100 & 100 & 100 \\
\bottomrule
\end{tabular}
\end{table}

\paragraph{BFCL.}
LLM backbone: Qwen3.5-35B-A3B {with FP8 quantization}. Embedding model: Qwen3-Embedding-8B. 500 problems (400 train / 100 test).

\begin{table}[H]
\centering
\caption{BFCL hyperparameters.}
\label{tab:hp_bfcl}
\setlength{\tabcolsep}{3.5pt}
\small
\begin{tabular}{l ccccccc}
\toprule
 & \textbf{MemQ} & \textbf{MemRL} & \textbf{MemP} & \textbf{Mem0} & {\textbf{Self-RAG}} & \textbf{RAG} \\
\midrule
$\alpha$ & 0.3 & 0.3 & N/A & N/A & N/A & N/A \\
$\gamma$ & 0.5 & N/A & N/A & N/A & N/A & N/A \\
$\lambda$ & 0.8 & N/A & N/A & N/A & N/A & N/A \\
$\varepsilon$ & 0.01 & 0.01 & 0.01 & 0.0 & 0.01 & 0.01 \\
$k_\mathrm{ret}$ & 10 & 10 & 10 & 5 & 10 & 5 \\
$k_\mathrm{top}$ & 5 & 5 & 5 & 5 & 5 & 5 \\
$\theta_\mathrm{sim}$ & 0.5 & 0.5 & 0.5 & 0.0 & 0.5 & 0.5 \\
$w_s$ & 0.7 & 0.7 & N/A & N/A & N/A & N/A \\
$w_q$ & 0.3 & 0.3 & N/A & N/A & N/A & N/A \\
Batch size & 100 & 100 & 100 & 100 & 100 & 100 \\
Max tokens & 4096 & 4096 & 4096 & 4096 & 4096 & 4096 \\
Initial Q & 0.5 & 0.5 & N/A & N/A & N/A & N/A \\
Build & proceduralization & proceduralization & proceduralization &trajectory  & trajectory  & trajectory \\
Update & adjustment & adjustment & adjustment & vanilla & vanilla & vanilla \\
Epochs & 20 & 20 & 20 & 20 & 20 & 20 \\
\bottomrule
\end{tabular}
\end{table}

\paragraph{GPQA Diamond.}
LLM backbone: Gemma-4-E4B-it. Embedding model: Qwen3-Embedding-8B. 198 problems (158 train / 40 test).

\begin{table}[H]
\centering
\caption{GPQA Diamond hyperparameters.}
\label{tab:hp_gpqa}
\setlength{\tabcolsep}{3.5pt}
\small
\begin{tabular}{l ccccccc}
\toprule
 & \textbf{MemQ} & \textbf{MemRL} & \textbf{MemP} & \textbf{Mem0} & {\textbf{Self-RAG}} & \textbf{RAG} \\
\midrule
$\alpha$ & 0.3 & 0.3 & N/A & N/A & N/A & N/A \\
$\gamma$ & 0.5 & N/A & N/A & N/A & N/A & N/A \\
$\lambda$ & 0.7 & N/A & N/A & N/A & N/A & N/A \\
$\varepsilon$ & 0.01 & 0.01 & 0.01 & 0.0 & 0.01 & 0.01 \\
$k_\mathrm{ret}$ & 10 & 10 & 10 & 5 & 10 & 10 \\
$k_\mathrm{top}$ & 5 & 5 & 5 & 5 & 5 & 5 \\
$\theta_\mathrm{sim}$ & 0.5 & 0.5 & 0.5 & 0.0 & 0.5 & 0.5 \\
$w_s$ & 0.5 & 0.7 & N/A & N/A & N/A & N/A \\
$w_q$ & 0.5 & 0.3 & N/A & N/A & N/A & N/A \\
Batch size & 40 & 100 & 100 & 100 & 100 & 100 \\
Max tokens & 16384 & 16384 & 16384 & 16384 & 16384 & 16384 \\
Initial Q & 0.5 & 0.5 & N/A & N/A & N/A & N/A \\
Build & proceduralization & proceduralization & proceduralization &trajectory  & trajectory  & trajectory \\
Update & adjustment & adjustment & adjustment & vanilla& vanilla & vanilla \\
Epochs & 20 & 20 & 20 & 20 & 20 & 20 \\
\bottomrule
\end{tabular}
\end{table}

\paragraph{{DSBench.}}
\label{app:dsbench_curve_protocol}
{LLM backbone: DeepSeek-v4-flash-0731. Embedding model: Qwen3-Embedding-8B. Training: 350 tasks from 30 challenges; held-out evaluation: 116 tasks from 8 challenges.}

\begin{table}[H]
\centering

\caption{{DSBench hyperparameters.}}
\label{tab:hp_dsbench}
\setlength{\tabcolsep}{3.5pt}
\small
\begin{tabular}{l ccccccc}
\toprule
 & \textbf{MemQ} & \textbf{MemRL} & \textbf{MemP} & \textbf{Mem0} & {\textbf{Self-RAG}} & \textbf{RAG} \\
\midrule
$\alpha$ & 0.3 & 0.3 & N/A & N/A & N/A & N/A \\
$\gamma$ & 0.9 & N/A & N/A & N/A & N/A & N/A \\
$\lambda$ & 0.5 & N/A & N/A & N/A & N/A & N/A \\
$\varepsilon$ & 0.1 & 0.1 & 0.0 & 0.0 & 0.0 & 0.0 \\
$k_\mathrm{ret}$ & 3 & 3 & 3 & 3 & 3 & 3 \\
$k_\mathrm{top}$ & 5 & 5 & 5 & 5 & 5 & 5 \\
$\theta_\mathrm{sim}$ & 0.0 & 0.0 & 0.0 & 0.0 & 0.0 & 0.0 \\
$w_s$ & 0.5 & 0.5 & N/A & N/A & N/A & N/A \\
$w_q$ & 0.5 & 0.5 & N/A & N/A & N/A & N/A \\
Batch size & 30 & 30 & 30 & 30 & 30 & 30 \\
Max tokens & 32768 & 32768 & 32768 & 32768 & 32768 & 32768 \\
Initial Q & 0.66 & 0 & N/A & N/A & N/A & N/A \\
Build & proceduralization & proceduralization & proceduralization & trajectory & trajectory & trajectory \\
Update & adjustment & adjustment & adjustment & vanilla & vanilla & vanilla \\
Epochs & 10 & 10 & 10 & 10 & 10 & 10 \\
\bottomrule
\end{tabular}
\end{table}

\paragraph{{AutomationBench.}}
{LLM backbone: DeepSeek-v4-flash-0731 (284B). Embedding model and other
unspecified settings follow DSBench. The split contains 480 training tasks and
120 held-out/test tasks.}

\begin{table}[H]
\centering
\caption{{AutomationBench hyperparameters.}}
\label{tab:hp_automationbench}
\setlength{\tabcolsep}{3.5pt}
\small
\begin{tabular}{l ccccccc}
\toprule
 & \textbf{MemQ} & \textbf{MemRL} & \textbf{MemP} & \textbf{Mem0} & {\textbf{Self-RAG}} & \textbf{RAG} \\
\midrule
$\alpha$ & 0.3 & 0.3 & N/A & N/A & N/A & N/A \\
$\gamma$ & 0.5 & N/A & N/A & N/A & N/A & N/A \\
$\lambda$ & 0.8 & N/A & N/A & N/A & N/A & N/A \\
$\varepsilon$ & 0.1 & 0.1 & 0.0 & 0.0 & 0.0 & 0.0 \\
$k_\mathrm{ret}$ & 3 & 3 & 3 & 3 & 3 & 3 \\
$k_\mathrm{top}$ & 5 & 5 & 5 & 5 & 5 & 5 \\
$\theta_\mathrm{sim}$ & 0.0 & 0.0 & 0.0 & 0.0 & 0.0 & 0.0 \\
$w_s$ & 0.7 & 0.5 & N/A & N/A & N/A & N/A \\
$w_q$ & 0.3 & 0.5 & N/A & N/A & N/A & N/A \\
Batch size & 50 & 50 & 50 & 50 & 50 & 50 \\
Max tokens & 16384 & 16384 & 16384 & 16384 & 16384 & 16384 \\
Initial Q & 0.66 & 0 & N/A & N/A & N/A & N/A \\
Build & proceduralization & proceduralization & proceduralization & trajectory & trajectory & trajectory \\
Update & adjustment & adjustment & adjustment & vanilla & vanilla & vanilla \\
Epochs & 10 & 10 & 10 & 10 & 10 & 10 \\
\bottomrule
\end{tabular}
\end{table}

\end{document}